\documentclass{article}

\usepackage{PRIMEarxiv}
\usepackage[utf8]{inputenc} 
\usepackage[T1]{fontenc}    
\usepackage{hyperref}       
\usepackage{url}            
\usepackage{booktabs}       
\usepackage{amsfonts}       
\usepackage{nicefrac}       
\usepackage{microtype}      
\usepackage{lipsum}
\usepackage{fancyhdr}       
\usepackage{graphicx}       
\graphicspath{{media/}}     
\usepackage{rotating}       
\usepackage{subcaption}     
\usepackage{pdflscape}
\usepackage{afterpage}

\usepackage{pifont}
\usepackage{adjustbox}
\usepackage{array}
\usepackage{amsmath}
\usepackage{pgfplots}
\usepackage{mathtools} 
\usepackage{multirow}

\usepackage{tikz}
\usetikzlibrary{decorations.pathreplacing,calligraphy}

\tikzset{ 
diagonal fill/.style 2 args={fill=#2, path picture={
\fill[#1, sharp corners] (path picture bounding box.south west) -|
                         (path picture bounding box.north east) -- cycle;}},
reversed diagonal fill/.style 2 args={fill=#2, path picture={
\fill[#1, sharp corners] (path picture bounding box.north west) |- 
                         (path picture bounding box.south east) -- cycle;}}
}

\newcommand{\pwf}[1][\beta]{$\prescript{}{PW}{f}_#1$}
\newcommand{\paf}[1][\beta]{$\prescript{}{PA}{f}_#1$}
\newcommand{\dtpaf}[2][\beta]{$\prescript{}{dtPA}{f}_#1^{#2}$}
\newcommand{\pakf}[2][\beta]{$\prescript{#2\%}{PA}{f}_#1$}
\newcommand{\lsf}[2][\beta]{$\prescript{}{ls}{f}_#1^{#2}$}
\newcommand{\cf}[1][\beta]{$\prescript{}{C}{f}_#1$}
\newcommand{\segf}[1][\beta]{$\prescript{}{S}{f}_#1$}
\newcommand{\rf}[3][\beta]{$\prescript{#2}{R}{f}_{#1}^{#3}$}
\newcommand{\af}[1][\beta]{$\prescript{}{A}{f}_#1$}
\newcommand{\ttolf}[2][\beta]{$\prescript{}{t}{f}_#1^{#2}$}
\newcommand{\taf}[4][\beta]{$\prescript{}{Ta}{f}_#1^{#3}$} 
\newcommand{\etaf}[3][\beta]{$\prescript{}{eTa}{f}_#1$} 
\newcommand{\nab}{$NAB$}
\newcommand{\tempdist}{$TD$}
\newcommand{\aucroc}{${AUC}_{{ROC}}$}
\newcommand{\aucpr}{${AUC}_{{PR}}$}
\newcommand{\vusroc}[1]{${VUS}_{{ROC}}^#1$}
\newcommand{\vuspr}[1]{${VUS}_{{PR}}^#1$}
\newcommand{\patk}[1][K]{$P@#1$}
\newcommand{\bestpwf}[1][\beta]{$\prescript{best}{PW}{f}_#1$}

\usetikzlibrary{shapes.misc}
\tikzset{cross/.style={cross out, draw=white, minimum size=#1, inner sep=0pt, outer sep=0pt},
cross/.default={2pt}}

\newcommand{\anomaly}[3]{
  \fill[red] (#1 cm,#2 cm) circle (#3 cm);
  \draw (#1 cm,#2 cm) node[cross=#3 mm + #3 mm, line width=#3 cm + #3mm + #3mm + #3mm +#3mm +#3mm]{}; 
}
\newcommand{\nomaly}[3]{
  \fill[gray!20!darkgray] (#1,#2) circle (#3);
}

\DeclareRobustCommand{\legendanomaly}{
\begin{tikzpicture}
  \anomaly{0}{0}{0.1}
\end{tikzpicture}~Anomalous~point,~\begin{tikzpicture}
  \nomaly{0}{0}{0.1}
\end{tikzpicture}~Normal~point.
}

\pgfkeys{
  points/.is family,
  points,
  first x/.initial=0,
  second x/.initial=.2,
  last x/.initial=4,
  y/.initial=0,
  radius/.initial=.1,
}
\newcommand\pointsset[1]{\pgfkeys{points,#1}}
\newcommand\anomalies[1][]{
  \pointsset{#1,
    first x/.get=\minx,
    second x/.get=\nextx,
    last x/.get=\lastx,
    y/.get=\y,
    radius/.get=\r,
  }
  \foreach \x in {\minx, \nextx, ..., \lastx} {
    \anomaly{\x}{\y}{\r}
  };
}\newcommand\nomalies[1][]{
  \pointsset{#1,
    first x/.get=\minx,
    second x/.get=\nextx,
    last x/.get=\lastx,
    y/.get=\y,
    radius/.get=\r,
  }
  \foreach \x in {\minx, \nextx, ..., \lastx} {
    \nomaly{\x}{\y}{\r}
  };
}

\newcommand{\showLengthProblemI}[0]{
\begin{tabular}[t]{ccccccccccccccc}
\toprule
\begin{tikzpicture}[baseline=-\the\dimexpr\fontdimen22\textfont2\relax]
\draw (0,0) -- (4.6,0);
\nomalies[first x=0, second x=0.2, last x=4.79, y=0, radius=0.1]
\anomalies[first x=0.6, second x=0.8, last x=0.81, y=0, radius=0.1]
\anomalies[first x=2.2, second x=2.4, last x=2.41, y=0, radius=0.1]
\anomalies[first x=3.8, second x=4.0, last x=4.61, y=0, radius=0.1]
\end{tikzpicture}
&\pwf[1]&\paf[1]&\dtpaf[1]{2}&\pakf[1]{20}&\lsf[1]{2}&\segf[1]&\cf[1]&\ttolf[1]{2}&\rf[1]{flat}{0.2}&\taf[1]{0.5}{0}{0.5}&\etaf[1]{0.5}{0.1}&\af[1]&\nab&\tempdist\\
\midrule
\begin{tikzpicture}[baseline=-\the\dimexpr\fontdimen22\textfont2\relax]
\draw (0,0) -- (4.6,0);
\nomalies[first x=0, second x=0.2, last x=4.79, y=0, radius=0.1]
\anomalies[first x=3.8, second x=4.0, last x=4.61, y=0, radius=0.1]
\end{tikzpicture}
&\textbf{0.71}&\textbf{0.71}&\textbf{0.71}&\textbf{0.71}&0.6&0.5&0.5&\textbf{0.71}&0.5&0.5&0.5&0.5&33.33&46\\
\begin{tikzpicture}[baseline=-\the\dimexpr\fontdimen22\textfont2\relax]
\draw (0,0) -- (4.6,0);
\nomalies[first x=0, second x=0.2, last x=4.79, y=0, radius=0.1]
\anomalies[first x=0.6, second x=0.8, last x=0.81, y=0, radius=0.1]
\anomalies[first x=2.2, second x=2.4, last x=2.41, y=0, radius=0.1]
\end{tikzpicture}
&0.62&0.62&0.62&0.62&\textbf{0.73}&\textbf{0.8}&\textbf{0.8}&0.62&\textbf{0.8}&\textbf{0.8}&\textbf{0.8}&\textbf{0.8}&\textbf{66.67}&\textbf{45}\\
\bottomrule
\end{tabular}

}
\newcommand{\showLengthProblemII}[0]{
\begin{tabular}[t]{cccccccccccccc}
\toprule
\begin{tikzpicture}[baseline=-\the\dimexpr\fontdimen22\textfont2\relax]
\draw (0,0) -- (4.2,0);
\nomalies[first x=0, second x=0.2, last x=4.39, y=0, radius=0.1]
\anomaly{0.4}{0}{0.1}
\anomaly{0.8}{0}{0.1}
\anomaly{1.2}{0}{0.1}
\anomalies[first x=3.0, second x=3.2, last x=3.81, y=0, radius=0.1]
\end{tikzpicture}
&\pwf[1]&\paf[1]&\dtpaf[1]{2}&\pakf[1]{20}&\lsf[1]{2}&\segf[1]&\cf[1]&\ttolf[1]{2}&\rf[1]{flat}{0.2}&\taf[1]{0.5}{0}{0.5}&\etaf[1]{0.5}{0.1}&\af[1]&\tempdist\\
\midrule
\begin{tikzpicture}[baseline=-\the\dimexpr\fontdimen22\textfont2\relax]
\draw (0,0) -- (4.2,0);
\nomalies[first x=0, second x=0.2, last x=4.39, y=0, radius=0.1]
\anomalies[first x=0.4, second x=0.6, last x=1.21, y=0, radius=0.1]
\end{tikzpicture}
&0.46&0.46&0.46&0.46&0.67&\textbf{0.86}&0.67&0.55&\textbf{0.71}&\textbf{0.77}&\textbf{0.77}&\textbf{0.77}&57\\
\begin{tikzpicture}[baseline=-\the\dimexpr\fontdimen22\textfont2\relax]
\draw (0,0) -- (4.2,0);
\nomalies[first x=0, second x=0.2, last x=4.39, y=0, radius=0.1]
\anomaly{0.4}{0}{0.1}
\anomalies[first x=3.0, second x=3.2, last x=3.81, y=0, radius=0.1]
\end{tikzpicture}
&\textbf{0.86}&\textbf{0.86}&\textbf{0.86}&\textbf{0.86}&\textbf{1.0}&0.67&\textbf{0.67}&\textbf{0.93}&0.67&0.67&0.67&0.67&\textbf{6}\\
\bottomrule
\end{tabular}

}
\newcommand{\showShortPrecitions}[0]{
\begin{tabular}[t]{ccccccccccccccc}
\toprule
\begin{tikzpicture}[baseline=-\the\dimexpr\fontdimen22\textfont2\relax]
\draw (0,0) -- (4.2,0);
\nomalies[first x=0, second x=0.2, last x=4.39, y=0, radius=0.1]
\anomalies[first x=2.8, second x=3.0, last x=4.01, y=0, radius=0.1]
\end{tikzpicture}
&\pwf[1]&\paf[1]&\dtpaf[1]{2}&\pakf[1]{20}&\lsf[1]{2}&\segf[1]&\cf[1]&\ttolf[1]{2}&\rf[1]{flat}{0.2}&\taf[1]{0.5}{0}{0.5}&\etaf[1]{0.5}{0.1}&\af[1]&\nab&\tempdist\\
\midrule
\begin{tikzpicture}[baseline=-\the\dimexpr\fontdimen22\textfont2\relax]
\draw (0,0) -- (4.2,0);
\nomalies[first x=0, second x=0.2, last x=4.39, y=0, radius=0.1]
\anomaly{0.4}{0}{0.1}
\anomaly{3.0}{0}{0.1}
\end{tikzpicture}
&0.22&\textbf{0.93}&\textbf{0.93}&0.22&\textbf{0.89}&\textbf{0.67}&\textbf{0.67}&0.53&0.39&0.12&0.53&0.68&\textbf{93.8}&\textbf{28}\\
\begin{tikzpicture}[baseline=-\the\dimexpr\fontdimen22\textfont2\relax]
\draw (0,0) -- (4.2,0);
\nomalies[first x=0, second x=0.2, last x=4.39, y=0, radius=0.1]
\anomalies[first x=0.4, second x=0.6, last x=1.41, y=0, radius=0.1]
\anomalies[first x=3.0, second x=3.2, last x=4.01, y=0, radius=0.1]
\end{tikzpicture}
&\textbf{0.63}&0.7&0.7&\textbf{0.7}&0.73&\textbf{0.67}&\textbf{0.67}&\textbf{0.67}&\textbf{0.64}&\textbf{0.65}&\textbf{0.65}&\textbf{0.76}&66.3&58\\
\bottomrule
\end{tabular}

}
\newcommand{\showDetectionOverCovering}[0]{
\begin{tabular}[t]{ccccccccccccccc}
\toprule
\begin{tikzpicture}[baseline=-\the\dimexpr\fontdimen22\textfont2\relax]
\draw (0,0) -- (5.4,0);
\nomalies[first x=0, second x=0.2, last x=5.59, y=0, radius=0.1]
\anomalies[first x=1.0, second x=1.2, last x=2.41, y=0, radius=0.1]
\anomalies[first x=4.0, second x=4.2, last x=5.41, y=0, radius=0.1]
\end{tikzpicture}
&\pwf[1]&\paf[1]&\dtpaf[1]{2}&\pakf[1]{20}&\lsf[1]{2}&\segf[1]&\cf[1]&\ttolf[1]{2}&\rf[1]{flat}{0.2}&\taf[1]{0.5}{0}{0.5}&\etaf[1]{0.5}{0.1}&\af[1]&\nab&\tempdist\\
\midrule
\begin{tikzpicture}[baseline=-\the\dimexpr\fontdimen22\textfont2\relax]
\draw (0,0) -- (5.4,0);
\nomalies[first x=0, second x=0.2, last x=5.59, y=0, radius=0.1]
\anomalies[first x=1.0, second x=1.2, last x=2.41, y=0, radius=0.1]
\end{tikzpicture}
&\textbf{0.67}&0.67&0.67&\textbf{0.67}&0.71&0.67&0.67&\textbf{0.67}&\textbf{0.67}&\textbf{0.67}&0.67&0.67&50.0&92\\
\begin{tikzpicture}[baseline=-\the\dimexpr\fontdimen22\textfont2\relax]
\draw (0,0) -- (5.4,0);
\nomalies[first x=0, second x=0.2, last x=5.59, y=0, radius=0.1]
\anomaly{1.0}{0}{0.1}
\anomaly{4.0}{0}{0.1}
\end{tikzpicture}
&0.22&\textbf{1.0}&\textbf{1.0}&0.22&\textbf{1.0}&\textbf{1.0}&\textbf{1.0}&0.55&0.46&0.12&\textbf{0.72}&\textbf{0.77}&\textbf{100.0}&\textbf{56}\\
\bottomrule
\end{tabular}

}
\newcommand{\showLabellingProblem}[0]{
\begin{tabular}[t]{cccccccccccccc}
\toprule
&\pwf[1]&\paf[1]&\dtpaf[1]{2}&\pakf[1]{20}&\lsf[1]{2}&\segf[1]&\cf[1]&\ttolf[1]{2}&\rf[1]{flat}{0.2}&\taf[1]{0.5}{0}{0.5}&\etaf[1]{0.5}{0.1}&\af[1]&\tempdist\\
\midrule
\begin{tikzpicture}[baseline=-\the\dimexpr\fontdimen22\textfont2\relax]
\draw (0,0) -- (5.8,0);
\nomalies[first x=0, second x=0.2, last x=5.99, y=0, radius=0.1]
\anomaly{3.6}{0}{0.1}
\anomaly{4.8}{0}{0.1}
\end{tikzpicture}
&{0.25}&{0.92}&{0.92}&{0.25}&{0.86}&{0.67}&{0.67}&{0.91}&{0.4}&{0.14}&{0.54}&{0.91}&{10}\\
\midrule
\begin{tikzpicture}[baseline=-\the\dimexpr\fontdimen22\textfont2\relax]
\draw (0,0) -- (5.8,0);
\nomalies[first x=0, second x=0.2, last x=5.99, y=0, radius=0.1]
\anomalies[first x=3.6, second x=3.8, last x=4.61, y=0, radius=0.1]
\end{tikzpicture}
&{0.25}&{0.25}&{0.25}&{0.25}&{0.4}&{0.67}&{0.25}&{0.91}&{0.4}&{0.14}&{0}&{0.83}&{10}\\
\bottomrule
\end{tabular}

}
\newcommand{\showNonbinaryDetectionOverCovering}[0]{
\begin{tabular}[t]{ccccccc}
\toprule
\begin{tikzpicture}[baseline=-\the\dimexpr\fontdimen22\textfont2\relax]
\draw (0,0) -- (3.24,0);
\nomalies[first x=0, second x=0.12, last x=3.35, y=0, radius=0.06]
\anomalies[first x=0.6, second x=0.72, last x=1.45, y=0, radius=0.06]
\anomalies[first x=2.4, second x=2.52, last x=3.25, y=0, radius=0.06]
\end{tikzpicture}
&\patk[16]&\bestpwf&\aucroc&\aucpr&\vusroc{4}&\vuspr{4}\\
\midrule
\begin{tikzpicture}[scale=1.2, baseline=-\the\dimexpr\fontdimen22\textfont2\relax]
\foreach \i/\a in
{0.0/ -0.083, 0.1/ -0.083, 0.2/ -0.083, 0.3/ -0.083, 0.4/ -0.083, 0.5/ 0.25, 0.6/ 0.25, 0.7/ 0.25, 0.8/ 0.25, 0.9/ 0.25, 1.0/ 0.25, 1.1/ 0.25, 1.2/ 0.25, 1.3/ -0.083, 1.4/ -0.083, 1.5/ -0.083, 1.6/ -0.083, 1.7/ -0.083, 1.8/ -0.083, 1.9/ -0.083, 2.0/ -0.083, 2.1/ -0.083, 2.2/ -0.083, 2.3/ -0.083, 2.4/ -0.083, 2.5/ -0.083, 2.6/ -0.083, 2.7/ -0.083}{
\coordinate (now) at (\i,\a) {};
  \ifthenelse{\equal{\i}{0.0}}{}{
  \draw[-, blue] (prev) -- (now);
  }
  \coordinate (prev) at (\i,\a) {};
}
\end{tikzpicture}
&\textbf{0.57}&\textbf{0.73}&\textbf{0.75}&\textbf{0.79}&\textbf{0.65}&\textbf{0.86}\\
\begin{tikzpicture}[scale=1.2, baseline=-\the\dimexpr\fontdimen22\textfont2\relax]
\foreach \i/\a in
{0.0/ -0.083, 0.1/ -0.083, 0.2/ -0.083, 0.3/ -0.083, 0.4/ -0.083, 0.5/ 0.25, 0.6/ -0.083, 0.7/ -0.083, 0.8/ -0.083, 0.9/ -0.083, 1.0/ -0.083, 1.1/ -0.083, 1.2/ -0.083, 1.3/ -0.083, 1.4/ -0.083, 1.5/ -0.083, 1.6/ -0.083, 1.7/ -0.083, 1.8/ -0.083, 1.9/ -0.083, 2.0/ 0.25, 2.1/ -0.083, 2.2/ -0.083, 2.3/ -0.083, 2.4/ -0.083, 2.5/ -0.083, 2.6/ -0.083, 2.7/ -0.083}{
\coordinate (now) at (\i,\a) {};
  \ifthenelse{\equal{\i}{0.0}}{}{
  \draw[-, blue] (prev) -- (now);
  }
  \coordinate (prev) at (\i,\a) {};
}
\end{tikzpicture}
&\textbf{0.57}&\textbf{0.73}&0.56&0.62&0.59&0.83\\
\bottomrule
\end{tabular}

}
\newcommand{\showNonbinaryLengthI}[0]{
\begin{tabular}[t]{ccccccc}
\toprule
\begin{tikzpicture}[baseline=-\the\dimexpr\fontdimen22\textfont2\relax]
\draw (0,0) -- (3.12,0);
\nomalies[first x=0, second x=0.12, last x=3.23, y=0, radius=0.06]
\anomalies[first x=0.36, second x=0.48, last x=0.49, y=0, radius=0.06]
\anomalies[first x=1.32, second x=1.44, last x=1.45, y=0, radius=0.06]
\anomalies[first x=2.28, second x=2.4, last x=3.01, y=0, radius=0.06]
\end{tikzpicture}
&\patk[11]&\bestpwf&\aucroc&\aucpr&\vusroc{4}&\vuspr{4}\\
\midrule
\begin{tikzpicture}[scale=1.2, baseline=-\the\dimexpr\fontdimen22\textfont2\relax]
\foreach \i/\a in
{0.0/ -0.083, 0.1/ -0.083, 0.2/ -0.083, 0.3/ -0.083, 0.4/ -0.083, 0.5/ -0.083, 0.6/ -0.083, 0.7/ -0.083, 0.8/ -0.083, 0.9/ -0.083, 1.0/ -0.083, 1.1/ -0.083, 1.2/ -0.083, 1.3/ -0.083, 1.4/ -0.083, 1.5/ -0.083, 1.6/ -0.083, 1.7/ -0.083, 1.8/ 0.0, 1.9/ 0.167, 2.0/ 0.25, 2.1/ 0.25, 2.2/ 0.25, 2.3/ 0.25, 2.4/ 0.25, 2.5/ 0.167, 2.6/ 0.0}{
\coordinate (now) at (\i,\a) {};
  \ifthenelse{\equal{\i}{0.0}}{}{
  \draw[-, blue] (prev) -- (now);
  }
  \coordinate (prev) at (\i,\a) {};
}
\end{tikzpicture}
&\textbf{0.41}&\textbf{0.78}&\textbf{0.8}&\textbf{0.78}&0.61&\textbf{0.74}\\
\begin{tikzpicture}[scale=1.2, baseline=-\the\dimexpr\fontdimen22\textfont2\relax]
\foreach \i/\a in
{0.0/ -0.083, 0.1/ -0.083, 0.2/ 0.0, 0.3/ 0.167, 0.4/ 0.167, 0.5/ 0.0, 0.6/ -0.083, 0.7/ -0.083, 0.8/ -0.083, 0.9/ -0.083, 1.0/ 0.0, 1.1/ 0.167, 1.2/ 0.167, 1.3/ 0.0, 1.4/ -0.083, 1.5/ -0.083, 1.6/ -0.083, 1.7/ -0.083, 1.8/ -0.083, 1.9/ -0.083, 2.0/ -0.083, 2.1/ -0.083, 2.2/ -0.083, 2.3/ -0.083, 2.4/ -0.083, 2.5/ -0.083, 2.6/ -0.083}{
\coordinate (now) at (\i,\a) {};
  \ifthenelse{\equal{\i}{0.0}}{}{
  \draw[-, blue] (prev) -- (now);
  }
  \coordinate (prev) at (\i,\a) {};
}
\end{tikzpicture}
&\textbf{0.41}&0.58&0.6&0.62&\textbf{0.61}&0.69\\
\bottomrule
\end{tabular}

}
\newcommand{\showScoreValueProblem}[0]{
\begin{tabular}[t]{ccccccc}
\toprule
\begin{tikzpicture}[baseline=-\the\dimexpr\fontdimen22\textfont2\relax]
\draw (0,0) -- (3.0,0);
\nomalies[first x=0, second x=0.15, last x=3.14, y=0, radius=0.075]
\anomalies[first x=1.2, second x=1.35, last x=1.51, y=0, radius=0.075]
\end{tikzpicture}
&\patk[3]&\bestpwf&\aucroc&\aucpr&\vusroc{4}&\vuspr{4}\\
\midrule
\begin{tikzpicture}[scale=1.5, baseline=-\the\dimexpr\fontdimen22\textfont2\relax]
\foreach \i/\a in
{0.0/ -0.042, 0.1/ -0.04, 0.2/ -0.037, 0.3/ -0.033, 0.4/ -0.029, 0.5/ -0.022, 0.6/ -0.013, 0.7/ 0.0, 0.8/ 0.022, 0.9/ 0.067, 1.0/ 0.2, 1.1/ 0.067, 1.2/ 0.022, 1.3/ 0.0, 1.4/ -0.013, 1.5/ -0.022, 1.6/ -0.029, 1.7/ -0.033, 1.8/ -0.037, 1.9/ -0.04, 2.0/ -0.042}{
\coordinate (now) at (\i,\a) {};
  \ifthenelse{\equal{\i}{0.0}}{}{
  \draw[-, blue] (prev) -- (now);
  }
  \coordinate (prev) at (\i,\a) {};
}
\end{tikzpicture}
&\textbf{0.67}&\textbf{0.75}&\textbf{0.96}&\textbf{0.76}&\textbf{0.96}&\textbf{0.83}\\
\begin{tikzpicture}[scale=1.5, baseline=-\the\dimexpr\fontdimen22\textfont2\relax]
\foreach \i/\a in
{0.0/ -0.039, 0.1/ 0.027, 0.2/ 0.062, 0.3/ 0.09, 0.4/ 0.113, 0.5/ 0.134, 0.6/ 0.153, 0.7/ 0.17, 0.8/ 0.186, 0.9/ 0.201, 1.0/ 0.216, 1.1/ 0.201, 1.2/ 0.186, 1.3/ 0.17, 1.4/ 0.153, 1.5/ 0.134, 1.6/ 0.113, 1.7/ 0.09, 1.8/ 0.062, 1.9/ 0.027, 2.0/ -0.039}{
\coordinate (now) at (\i,\a) {};
  \ifthenelse{\equal{\i}{0.0}}{}{
  \draw[-, blue] (prev) -- (now);
  }
  \coordinate (prev) at (\i,\a) {};
}
\end{tikzpicture}
&\textbf{0.67}&\textbf{0.75}&\textbf{0.96}&\textbf{0.76}&\textbf{0.96}&\textbf{0.83}\\
\bottomrule
\end{tabular}

}
\newcommand{\showNonbinaryCloseFp}[0]{
\begin{tabular}[t]{ccccccc}
\toprule
\begin{tikzpicture}[baseline=-\the\dimexpr\fontdimen22\textfont2\relax]
\draw (0,0) -- (3.2,0);
\nomalies[first x=0, second x=0.2, last x=3.39, y=0, radius=0.1]
\anomalies[first x=2.4, second x=2.6, last x=2.81, y=0, radius=0.1]
\end{tikzpicture}
&\patk[3]&\bestpwf&\aucroc&\aucpr&\vusroc{4}&\vuspr{4}\\
\midrule
\begin{tikzpicture}[scale=2, baseline=-\the\dimexpr\fontdimen22\textfont2\relax]
\foreach \i/\a in
{0.0/ -0.05, 0.1/ -0.05, 0.2/ -0.05, 0.3/ -0.05, 0.4/ -0.046, 0.5/ -0.029, 0.6/ 0.024, 0.7/ 0.106, 0.8/ 0.15, 0.9/ 0.106, 1.0/ 0.024, 1.1/ -0.029, 1.2/ -0.046, 1.3/ -0.05, 1.4/ -0.05, 1.5/ -0.05, 1.6/ -0.05}{
\coordinate (now) at (\i,\a) {};
  \ifthenelse{\equal{\i}{0.0}}{}{
  \draw[-, blue] (prev) -- (now);
  }
  \coordinate (prev) at (\i,\a) {};
}
\end{tikzpicture}
&0.0&0.38&0.39&0.17&0.46&0.17\\
\begin{tikzpicture}[scale=2, baseline=-\the\dimexpr\fontdimen22\textfont2\relax]
\foreach \i/\a in
{0.0/ -0.05, 0.1/ -0.05, 0.2/ -0.05, 0.3/ -0.05, 0.4/ -0.05, 0.5/ -0.046, 0.6/ -0.029, 0.7/ 0.024, 0.8/ 0.106, 0.9/ 0.15, 1.0/ 0.106, 1.1/ 0.024, 1.2/ -0.029, 1.3/ -0.046, 1.4/ -0.05, 1.5/ -0.05, 1.6/ -0.05}{
\coordinate (now) at (\i,\a) {};
  \ifthenelse{\equal{\i}{0.0}}{}{
  \draw[-, blue] (prev) -- (now);
  }
  \coordinate (prev) at (\i,\a) {};
}
\end{tikzpicture}
&0.0&0.43&0.54&0.21&0.6&0.22\\
\begin{tikzpicture}[scale=2, baseline=-\the\dimexpr\fontdimen22\textfont2\relax]
\foreach \i/\a in
{0.0/ -0.05, 0.1/ -0.05, 0.2/ -0.05, 0.3/ -0.05, 0.4/ -0.05, 0.5/ -0.05, 0.6/ -0.046, 0.7/ -0.029, 0.8/ 0.024, 0.9/ 0.106, 1.0/ 0.15, 1.1/ 0.106, 1.2/ 0.024, 1.3/ -0.029, 1.4/ -0.046, 1.5/ -0.05, 1.6/ -0.05}{
\coordinate (now) at (\i,\a) {};
  \ifthenelse{\equal{\i}{0.0}}{}{
  \draw[-, blue] (prev) -- (now);
  }
  \coordinate (prev) at (\i,\a) {};
}
\end{tikzpicture}
&0.0&0.5&0.68&0.27&0.75&0.33\\
\begin{tikzpicture}[scale=2, baseline=-\the\dimexpr\fontdimen22\textfont2\relax]
\foreach \i/\a in
{0.0/ -0.05, 0.1/ -0.05, 0.2/ -0.05, 0.3/ -0.05, 0.4/ -0.05, 0.5/ -0.05, 0.6/ -0.05, 0.7/ -0.046, 0.8/ -0.029, 0.9/ 0.024, 1.0/ 0.106, 1.1/ 0.15, 1.2/ 0.106, 1.3/ 0.024, 1.4/ -0.029, 1.5/ -0.046, 1.6/ -0.05}{
\coordinate (now) at (\i,\a) {};
  \ifthenelse{\equal{\i}{0.0}}{}{
  \draw[-, blue] (prev) -- (now);
  }
  \coordinate (prev) at (\i,\a) {};
}
\end{tikzpicture}
&\textbf{0.33}&\textbf{0.6}&\textbf{0.82}&\textbf{0.39}&\textbf{0.88}&\textbf{0.56}\\
\bottomrule
\end{tabular}

}
\newcommand{\showNonbinaryNonsmoothCloseFp}[0]{
\begin{tabular}[t]{ccccccc}
\toprule
\begin{tikzpicture}[baseline=-\the\dimexpr\fontdimen22\textfont2\relax]
\draw (0,0) -- (3.2,0);
\nomalies[first x=0, second x=0.2, last x=3.39, y=0, radius=0.1]
\anomalies[first x=2.4, second x=2.6, last x=2.81, y=0, radius=0.1]
\end{tikzpicture}
&\patk[3]&\bestpwf&\aucroc&\aucpr&\vusroc{4}&\vuspr{4}\\
\midrule
\begin{tikzpicture}[scale=2, baseline=-\the\dimexpr\fontdimen22\textfont2\relax]
\foreach \i/\a in
{0.0/ -0.05, 0.1/ -0.05, 0.2/ -0.05, 0.3/ -0.05, 0.4/ -0.05, 0.5/ -0.05, 0.6/ -0.05, 0.7/ -0.05, 0.8/ 0.15, 0.9/ -0.05, 1.0/ -0.05, 1.1/ -0.05, 1.2/ -0.05, 1.3/ -0.05, 1.4/ -0.05, 1.5/ -0.05, 1.6/ -0.05}{
\coordinate (now) at (\i,\a) {};
  \ifthenelse{\equal{\i}{0.0}}{}{
  \draw[-, blue] (prev) -- (now);
  }
  \coordinate (prev) at (\i,\a) {};
}
\end{tikzpicture}
&\textbf{0.18}&\textbf{0.3}&\textbf{0.46}&\textbf{0.18}&0.48&0.11\\
\begin{tikzpicture}[scale=2, baseline=-\the\dimexpr\fontdimen22\textfont2\relax]
\foreach \i/\a in
{0.0/ -0.05, 0.1/ -0.05, 0.2/ -0.05, 0.3/ -0.05, 0.4/ -0.05, 0.5/ -0.05, 0.6/ -0.05, 0.7/ -0.05, 0.8/ -0.05, 0.9/ 0.15, 1.0/ -0.05, 1.1/ -0.05, 1.2/ -0.05, 1.3/ -0.05, 1.4/ -0.05, 1.5/ -0.05, 1.6/ -0.05}{
\coordinate (now) at (\i,\a) {};
  \ifthenelse{\equal{\i}{0.0}}{}{
  \draw[-, blue] (prev) -- (now);
  }
  \coordinate (prev) at (\i,\a) {};
}
\end{tikzpicture}
&\textbf{0.18}&\textbf{0.3}&\textbf{0.46}&\textbf{0.18}&0.48&0.11\\
\begin{tikzpicture}[scale=2, baseline=-\the\dimexpr\fontdimen22\textfont2\relax]
\foreach \i/\a in
{0.0/ -0.05, 0.1/ -0.05, 0.2/ -0.05, 0.3/ -0.05, 0.4/ -0.05, 0.5/ -0.05, 0.6/ -0.05, 0.7/ -0.05, 0.8/ -0.05, 0.9/ -0.05, 1.0/ 0.15, 1.1/ -0.05, 1.2/ -0.05, 1.3/ -0.05, 1.4/ -0.05, 1.5/ -0.05, 1.6/ -0.05}{
\coordinate (now) at (\i,\a) {};
  \ifthenelse{\equal{\i}{0.0}}{}{
  \draw[-, blue] (prev) -- (now);
  }
  \coordinate (prev) at (\i,\a) {};
}
\end{tikzpicture}
&\textbf{0.18}&\textbf{0.3}&\textbf{0.46}&\textbf{0.18}&0.5&0.19\\
\begin{tikzpicture}[scale=2, baseline=-\the\dimexpr\fontdimen22\textfont2\relax]
\foreach \i/\a in
{0.0/ -0.05, 0.1/ -0.05, 0.2/ -0.05, 0.3/ -0.05, 0.4/ -0.05, 0.5/ -0.05, 0.6/ -0.05, 0.7/ -0.05, 0.8/ -0.05, 0.9/ -0.05, 1.0/ -0.05, 1.1/ 0.15, 1.2/ -0.05, 1.3/ -0.05, 1.4/ -0.05, 1.5/ -0.05, 1.6/ -0.05}{
\coordinate (now) at (\i,\a) {};
  \ifthenelse{\equal{\i}{0.0}}{}{
  \draw[-, blue] (prev) -- (now);
  }
  \coordinate (prev) at (\i,\a) {};
}
\end{tikzpicture}
&\textbf{0.18}&\textbf{0.3}&\textbf{0.46}&\textbf{0.18}&\textbf{0.56}&\textbf{0.4}\\
\bottomrule
\end{tabular}

}
\newcommand{\showAucRocProblemII}[0]{
\begin{tabular}[t]{lcc}
\toprule
\begin{tikzpicture}[baseline=-\the\dimexpr\fontdimen22\textfont2\relax]
\draw (0,0) -- (7.56,0);
\nomalies[first x=0, second x=0.12, last x=7.67, y=0, radius=0.06]
\anomalies[first x=1.2, second x=1.32, last x=1.69, y=0, radius=0.06]
\end{tikzpicture}
&\aucroc&\vusroc{4}\\
\midrule
\begin{tikzpicture}[scale=1.2, baseline=-\the\dimexpr\fontdimen22\textfont2\relax]
\draw[-, white] (-0.05,0.2) -- (0.0,0.2); 
\foreach \i/\a in
{0.0/ -0.083, 0.1/ -0.083, 0.2/ -0.083, 0.3/ -0.083, 0.4/ -0.083, 0.5/ -0.083, 0.6/ -0.083, 0.7/ -0.083, 0.8/ -0.083, 0.9/ -0.083, 1.0/ 0.25, 1.1/ 0.25, 1.2/ 0.25, 1.3/ 0.25, 1.4/ -0.083, 1.5/ -0.083, 1.6/ -0.083, 1.7/ -0.083, 1.8/ -0.083, 1.9/ -0.083, 2.0/ -0.083, 2.1/ -0.083, 2.2/ -0.083, 2.3/ -0.083, 2.4/ -0.083, 2.5/ -0.083, 2.6/ -0.083, 2.7/ -0.083, 2.8/ -0.083, 2.9/ -0.083, 3.0/ -0.083, 3.1/ -0.083}{
\coordinate (now) at (\i,\a) {};
  \ifthenelse{\equal{\i}{0.0}}{}{
  \draw[-, blue] (prev) -- (now);
  }
  \coordinate (prev) at (\i,\a) {};
}
\end{tikzpicture}
&\textbf{0.9}&\textbf{0.88}\\
\begin{tikzpicture}[scale=1.2, baseline=-\the\dimexpr\fontdimen22\textfont2\relax]
\draw[-, white] (-0.05,0.2) -- (0.0,0.2); 
\foreach \i/\a in
{0.0/ -0.083, 0.1/ -0.083, 0.2/ -0.083, 0.3/ 0.25, 0.4/ 0.25, 0.5/ 0.25, 0.6/ 0.25, 0.7/ 0.25, 0.8/ 0.25, 0.9/ 0.25, 1.0/ 0.25, 1.1/ 0.25, 1.2/ 0.25, 1.3/ 0.25, 1.4/ 0.25, 1.5/ 0.25, 1.6/ -0.083, 1.7/ -0.083, 1.8/ -0.083, 1.9/ -0.083, 2.0/ -0.083, 2.1/ -0.083, 2.2/ -0.083, 2.3/ -0.083, 2.4/ -0.083, 2.5/ -0.083, 2.6/ -0.083, 2.7/ -0.083, 2.8/ -0.083, 2.9/ -0.083, 3.0/ -0.083, 3.1/ -0.083}{
\coordinate (now) at (\i,\a) {};
  \ifthenelse{\equal{\i}{0.0}}{}{
  \draw[-, blue] (prev) -- (now);
  }
  \coordinate (prev) at (\i,\a) {};
}
\end{tikzpicture}
&0.85&0.86\\
\midrule
\begin{tikzpicture}[scale=1.2, baseline=-\the\dimexpr\fontdimen22\textfont2\relax]
\draw[-, white] (-0.05,0.2) -- (0.0,0.2); 
\foreach \i/\a in
{0.0/ -0.083, 0.1/ -0.083, 0.2/ -0.083, 0.3/ -0.083, 0.4/ -0.083, 0.5/ -0.083, 0.6/ -0.083, 0.7/ -0.083, 0.8/ -0.083, 0.9/ -0.083, 1.0/ 0.25, 1.1/ 0.25, 1.2/ 0.25, 1.3/ 0.25, 1.4/ -0.083, 1.5/ -0.083, 1.6/ -0.083, 1.7/ -0.083, 1.8/ -0.083, 1.9/ -0.083, 2.0/ -0.083, 2.1/ -0.083, 2.2/ -0.083, 2.3/ -0.083, 2.4/ -0.083, 2.5/ -0.083, 2.6/ -0.083, 2.7/ -0.083, 2.8/ -0.083, 2.9/ -0.083, 3.0/ -0.083, 3.1/ -0.083, 3.2/ -0.083, 3.3/ -0.083, 3.4/ -0.083, 3.5/ -0.083, 3.6/ -0.083, 3.7/ -0.083, 3.8/ -0.083, 3.9/ -0.083, 4.0/ -0.083, 4.1/ -0.083, 4.2/ -0.083, 4.3/ -0.083, 4.4/ -0.083, 4.5/ -0.083, 4.6/ -0.083, 4.7/ -0.083, 4.8/ -0.083, 4.9/ -0.083, 5.0/ -0.083, 5.1/ -0.083, 5.2/ -0.083, 5.3/ -0.083, 5.4/ -0.083, 5.5/ -0.083, 5.6/ -0.083, 5.7/ -0.083, 5.8/ -0.083, 5.9/ -0.083, 6.0/ -0.083, 6.1/ -0.083, 6.2/ -0.083, 6.3/ -0.083}{
\coordinate (now) at (\i,\a) {};
  \ifthenelse{\equal{\i}{0.0}}{}{
  \draw[-, blue] (prev) -- (now);
  }
  \coordinate (prev) at (\i,\a) {};
}
\end{tikzpicture}
&0.9&0.87\\
\begin{tikzpicture}[scale=1.2, baseline=-\the\dimexpr\fontdimen22\textfont2\relax]
\draw[-, white] (-0.05,0.2) -- (0.0,0.2); 
\foreach \i/\a in
{0.0/ -0.083, 0.1/ -0.083, 0.2/ -0.083, 0.3/ 0.25, 0.4/ 0.25, 0.5/ 0.25, 0.6/ 0.25, 0.7/ 0.25, 0.8/ 0.25, 0.9/ 0.25, 1.0/ 0.25, 1.1/ 0.25, 1.2/ 0.25, 1.3/ 0.25, 1.4/ 0.25, 1.5/ 0.25, 1.6/ -0.083, 1.7/ -0.083, 1.8/ -0.083, 1.9/ -0.083, 2.0/ -0.083, 2.1/ -0.083, 2.2/ -0.083, 2.3/ -0.083, 2.4/ -0.083, 2.5/ -0.083, 2.6/ -0.083, 2.7/ -0.083, 2.8/ -0.083, 2.9/ -0.083, 3.0/ -0.083, 3.1/ -0.083, 3.2/ -0.083, 3.3/ -0.083, 3.4/ -0.083, 3.5/ -0.083, 3.6/ -0.083, 3.7/ -0.083, 3.8/ -0.083, 3.9/ -0.083, 4.0/ -0.083, 4.1/ -0.083, 4.2/ -0.083, 4.3/ -0.083, 4.4/ -0.083, 4.5/ -0.083, 4.6/ -0.083, 4.7/ -0.083, 4.8/ -0.083, 4.9/ -0.083, 5.0/ -0.083, 5.1/ -0.083, 5.2/ -0.083, 5.3/ -0.083, 5.4/ -0.083, 5.5/ -0.083, 5.6/ -0.083, 5.7/ -0.083, 5.8/ -0.083, 5.9/ -0.083, 6.0/ -0.083, 6.1/ -0.083, 6.2/ -0.083, 6.3/ -0.083}{
\coordinate (now) at (\i,\a) {};
  \ifthenelse{\equal{\i}{0.0}}{}{
  \draw[-, blue] (prev) -- (now);
  }
  \coordinate (prev) at (\i,\a) {};
}
\end{tikzpicture}
&\textbf{0.93}&\textbf{0.94}\\
\bottomrule
\end{tabular}

}\newcommand{\showAucRocProblemIII}[0]{
\begin{tabular}[t]{lcc}
\toprule
\begin{tikzpicture}[baseline=-\the\dimexpr\fontdimen22\textfont2\relax]
\draw (0,0) -- (7.56,0);
\nomalies[first x=0, second x=0.12, last x=7.67, y=0, radius=0.06]
\anomalies[first x=0.48, second x=0.6, last x=0.85, y=0, radius=0.06]
\end{tikzpicture}
&\aucroc&\vusroc{4}\\
\midrule
\begin{tikzpicture}[scale=1.2, baseline=-\the\dimexpr\fontdimen22\textfont2\relax]
\draw[-, white] (-0.05,0.2) -- (0.0,0.2); 
\foreach \i/\a in
{0.0/ 0.25, 0.1/ 0.234, 0.2/ 0.19, 0.3/ 0.129, 0.4/ 0.066, 0.5/ 0.012, 0.6/ -0.028, 0.7/ -0.055}{
\coordinate (now) at (\i,\a) {};
  \ifthenelse{\equal{\i}{0.0}}{}{
  \draw[-, blue] (prev) -- (now);
  }
  \coordinate (prev) at (\i,\a) {};
}
\end{tikzpicture}
&0.0&0.1\\
\begin{tikzpicture}[scale=1.2, baseline=-\the\dimexpr\fontdimen22\textfont2\relax]
\draw[-, white] (-0.05,0.2) -- (0.0,0.2); 
\foreach \i/\a in
{0.0/ 0.25, 0.1/ 0.234, 0.2/ 0.19, 0.3/ 0.129, 0.4/ 0.066, 0.5/ 0.012, 0.6/ -0.028, 0.7/ -0.055, 0.8/ -0.07, 0.9/ -0.078, 1.0/ -0.081, 1.1/ -0.083, 1.2/ -0.083, 1.3/ -0.083, 1.4/ -0.083, 1.5/ -0.083}{
\coordinate (now) at (\i,\a) {};
  \ifthenelse{\equal{\i}{0.0}}{}{
  \draw[-, blue] (prev) -- (now);
  }
  \coordinate (prev) at (\i,\a) {};
}
\end{tikzpicture}
&0.67&0.71\\
\begin{tikzpicture}[scale=1.2, baseline=-\the\dimexpr\fontdimen22\textfont2\relax]
\draw[-, white] (-0.05,0.2) -- (0.0,0.2); 
\foreach \i/\a in
{0.0/ 0.25, 0.1/ 0.234, 0.2/ 0.19, 0.3/ 0.129, 0.4/ 0.066, 0.5/ 0.012, 0.6/ -0.028, 0.7/ -0.055, 0.8/ -0.07, 0.9/ -0.078, 1.0/ -0.081, 1.1/ -0.083, 1.2/ -0.083, 1.3/ -0.083, 1.4/ -0.083, 1.5/ -0.083, 1.6/ -0.083, 1.7/ -0.083, 1.8/ -0.083, 1.9/ -0.083, 2.0/ -0.083, 2.1/ -0.083, 2.2/ -0.083, 2.3/ -0.083, 2.4/ -0.083, 2.5/ -0.083, 2.6/ -0.083, 2.7/ -0.083, 2.8/ -0.083, 2.9/ -0.083, 3.0/ -0.083, 3.1/ -0.083}{
\coordinate (now) at (\i,\a) {};
  \ifthenelse{\equal{\i}{0.0}}{}{
  \draw[-, blue] (prev) -- (now);
  }
  \coordinate (prev) at (\i,\a) {};
}
\end{tikzpicture}
&0.86&0.88\\
\begin{tikzpicture}[scale=1.2, baseline=-\the\dimexpr\fontdimen22\textfont2\relax]
\draw[-, white] (-0.05,0.2) -- (0.0,0.2); 
\foreach \i/\a in
{0.0/ 0.25, 0.1/ 0.234, 0.2/ 0.19, 0.3/ 0.129, 0.4/ 0.066, 0.5/ 0.012, 0.6/ -0.028, 0.7/ -0.055, 0.8/ -0.07, 0.9/ -0.078, 1.0/ -0.081, 1.1/ -0.083, 1.2/ -0.083, 1.3/ -0.083, 1.4/ -0.083, 1.5/ -0.083, 1.6/ -0.083, 1.7/ -0.083, 1.8/ -0.083, 1.9/ -0.083, 2.0/ -0.083, 2.1/ -0.083, 2.2/ -0.083, 2.3/ -0.083, 2.4/ -0.083, 2.5/ -0.083, 2.6/ -0.083, 2.7/ -0.083, 2.8/ -0.083, 2.9/ -0.083, 3.0/ -0.083, 3.1/ -0.083, 3.2/ -0.083, 3.3/ -0.083, 3.4/ -0.083, 3.5/ -0.083, 3.6/ -0.083, 3.7/ -0.083, 3.8/ -0.083, 3.9/ -0.083, 4.0/ -0.083, 4.1/ -0.083, 4.2/ -0.083, 4.3/ -0.083, 4.4/ -0.083, 4.5/ -0.083, 4.6/ -0.083, 4.7/ -0.083, 4.8/ -0.083, 4.9/ -0.083, 5.0/ -0.083, 5.1/ -0.083, 5.2/ -0.083, 5.3/ -0.083, 5.4/ -0.083, 5.5/ -0.083, 5.6/ -0.083, 5.7/ -0.083, 5.8/ -0.083, 5.9/ -0.083, 6.0/ -0.083, 6.1/ -0.083, 6.2/ -0.083, 6.3/ -0.083}{
\coordinate (now) at (\i,\a) {};
  \ifthenelse{\equal{\i}{0.0}}{}{
  \draw[-, blue] (prev) -- (now);
  }
  \coordinate (prev) at (\i,\a) {};
}
\end{tikzpicture}
&\textbf{0.93}&\textbf{0.94}\\
\bottomrule
\end{tabular}

}
\newcommand{\showNonbinaryShortPredictions}[0]{
\begin{tabular}[t]{ccccccc}
\toprule
\begin{tikzpicture}[baseline=-\the\dimexpr\fontdimen22\textfont2\relax]
\draw (0,0) -- (3.15,0);
\nomalies[first x=0, second x=0.15, last x=3.29, y=0, radius=0.075]
\anomalies[first x=2.1, second x=2.25, last x=3.01, y=0, radius=0.075]
\end{tikzpicture}
&\patk[7]&\bestpwf&\aucroc&\aucpr&\vusroc{4}&\vuspr{4}\\
\midrule
\begin{tikzpicture}[scale=1.5, baseline=-\the\dimexpr\fontdimen22\textfont2\relax]
\foreach \i/\a in
{0.0/ -0.067, 0.1/ 0.0, 0.2/ 0.067, 0.3/ 0.0, 0.4/ -0.067, 0.5/ -0.067, 0.6/ -0.067, 0.7/ -0.067, 0.8/ -0.067, 0.9/ -0.067, 1.0/ -0.067, 1.1/ -0.067, 1.2/ -0.067, 1.3/ -0.067, 1.4/ 0.0, 1.5/ 0.067, 1.6/ 0.0, 1.7/ -0.067, 1.8/ -0.067, 1.9/ -0.067, 2.0/ -0.067, 2.1/ -0.067}{
\coordinate (now) at (\i,\a) {};
  \ifthenelse{\equal{\i}{0.0}}{}{
  \draw[-, blue] (prev) -- (now);
  }
  \coordinate (prev) at (\i,\a) {};
}
\end{tikzpicture}
&0.32&0.48&0.61&0.4&0.61&0.49\\
\begin{tikzpicture}[scale=1.5, baseline=-\the\dimexpr\fontdimen22\textfont2\relax]
\foreach \i/\a in
{0.0/ -0.067, 0.1/ 0.0, 0.2/ 0.133, 0.3/ 0.2, 0.4/ 0.2, 0.5/ 0.2, 0.6/ 0.2, 0.7/ 0.133, 0.8/ 0.0, 0.9/ -0.067, 1.0/ -0.067, 1.1/ -0.067, 1.2/ -0.067, 1.3/ -0.067, 1.4/ 0.0, 1.5/ 0.133, 1.6/ 0.2, 1.7/ 0.2, 1.8/ 0.2, 1.9/ 0.2, 2.0/ 0.133, 2.1/ 0.0}{
\coordinate (now) at (\i,\a) {};
  \ifthenelse{\equal{\i}{0.0}}{}{
  \draw[-, blue] (prev) -- (now);
  }
  \coordinate (prev) at (\i,\a) {};
}
\end{tikzpicture}
&\textbf{0.5}&\textbf{0.63}&\textbf{0.76}&\textbf{0.49}&\textbf{0.73}&\textbf{0.63}\\
\bottomrule
\end{tabular}

}

\title{Navigating the Metric Maze: A Taxonomy of Evaluation Metrics for Anomaly Detection in Time Series
}

\author{
  Sondre S{\o}rb{\o}$^1$ \\
  \texttt{sondre.sorbo@sintef.no} \\
   \And
  Massimiliano Ruocco$^{1,2}$ \\
  \texttt{massimiliano.ruocco@sintef.no} \\
  \And
  $^1$\textmd{SINTEF Digital, Trondheim, Norway} \\
  $^2$Norwegian University of Science and Technology, Trondheim, Norway \\
}

\begin{document}
\maketitle

\begin{abstract}

The field of time series anomaly detection is constantly advancing, with several methods available, making it a challenge to determine the most appropriate method for a specific domain. The evaluation of these methods is facilitated by the use of metrics, which vary widely in their properties. Despite the existence of new evaluation metrics, there is limited agreement on which metrics are best suited for specific scenarios and domain, and the most commonly used metrics have faced criticism in the literature. This paper provides a comprehensive overview of the metrics used for the evaluation of time series anomaly detection methods, and also defines a taxonomy of these based on how they are calculated. By defining a set of properties for evaluation metrics and a set of specific case studies and experiments, twenty metrics are analyzed and discussed in detail, highlighting the unique suitability of each for specific tasks. Through extensive experimentation and analysis, this paper argues that the choice of evaluation metric must be made with care, taking into account the specific requirements of the task at hand. 

\end{abstract}


\section{Introduction}
With the growing trend of Industry 4.0, the amount of generated time series data increases, resulting in a huge demand for better time series analysis tools.
The study of Time Series Anomaly Detection (TSAD) has become increasingly popular in recent years due to its widespread application in various fields such as cyber-physical systems \cite{Feng2021TimeSA}, rail transit \cite{Wang2022ImprovedLT}, online service systems \cite{Ma2021JumpStartingMT}, smart grids \cite{Zhang2021TimeSA}, spacecraft telemetry \cite{Baireddy2021SpacecraftTA}, Internet of Things \cite{Chen2021LearningGS} and healthcare \cite{Keogh2006FindingUM}. The rapid advancement of machine learning technology has also opened up new opportunities for developing and improving TSAD methods. With the vast number of different machine learning architectures and techniques available, researchers are constantly exploring new ways to create more accurate anomaly detectors. Whether it be through trying out new algorithms, combining different approaches, or incorporating new data sources, the possibilities for improving TSAD are endless.

This highlights the importance of careful evaluation of TSAD algorithms, and the need for proper selection of evaluation metrics.
The choice of evaluation metric should be guided by the nature of the time series data and the specific requirements of the task at hand. 
Using the wrong metrics can lead to incorrect conclusions about the performance of an algorithm, potentially leading to incorrect decisions about its use in real-world applications. 
For example, Figure \ref{tab:PA_problem} shows a prediction evaluated by two of the most used metrics in the literature. They vastly disagree on the quality of the prediction. Despite this, most papers give very little attention to the choice of metric. It is important to understand the limitations and trade-offs of different evaluation metrics, and to make an informed choice when evaluating TSAD algorithms. Additionally, the development of new and improved evaluation metrics should continue to be a priority in the field of TSAD, to ensure that the best algorithms are selected and used in real-world applications.

\begin{figure*}[ht]
    \centering
\begin{tabular}[]{cccc}
\toprule
\multicolumn{2}{c}{Time series}&\multicolumn{2}{c}{Metrics}\\
\midrule
Labels:&
\begin{tikzpicture}[baseline=-\the\dimexpr\fontdimen22\textfont2\relax]
\nomalies[first x=0, second x=0.2, last x=6.39, y=0, radius=0.1]
\anomalies[first x=3.4, second x=3.6, last x=5.21, y=0, radius=0.1]
\end{tikzpicture}
&\pwf[1]&\paf[1]\\
\midrule
Prediction:&
\begin{tikzpicture}[baseline=-\the\dimexpr\fontdimen22\textfont2\relax]
\nomalies[first x=0, second x=0.2, last x=6.39, y=0, radius=0.1]
\anomaly{1.6}{0}{0.1}
\anomaly{4.8}{0}{0.1}
\end{tikzpicture}
&0.17&0.95\\
\bottomrule
\end{tabular}
    \caption{A hypothetical scenario where there is one long anomalous event in the labels, but the detector predicts two short events, only one of which is within the labelled event.
    Two of the most used evaluation metrics, the point-wise $f_1$ score (\pwf[1]) and point-adjusted $f_1$ score (\paf[1]), score the same prediction very differently. Both metrics output values between 0 and 1, where 1 is optimal.
    \newline\legendanomaly}
    \label{tab:PA_problem}
\end{figure*}

TSAD has recently been the subject of criticism in regards to its conventional evaluation metrics. A number of studies have pointed out shortcomings in the commonly used metrics, and proposed alternative metrics that address these issues \cite{Tatbul2018PrecisionAR,Hwang2019TimeSeriesAP,Abdulaal2021PracticalAT, Hwang2022DoYK,Kim2022TowardsAR, Doshi2022RewardOP,Garg2022AnEO, Paparrizos2022VolumeUT,Huet2022LocalEO}. 

For example, the work of \cite{Kim2022TowardsAR} criticize the point-adjust metric, and show that a detection algorithm outputting random noise is expected to produce very good scores, and capable of outperforming state of the art methods on most of the common benchmark datasets. The same conclusion is reached experimentally by \cite{Doshi2022RewardOP}. The work of \cite{Kim2022ASO}, include a review of several TSAD evaluation metrics from the perspective of industrial control systems, and discuss several properties required for the metrics. The work of \cite{Wu2022CurrentTS} analysed the most commonly used TSAD datasets and found that the majority suffered from flaws such as trivial anomalies, unrealistic anomaly density, mislabelled ground truth, and a high ratio of anomalies at the end of the time series. To address these issues, they introduced a new benchmark dataset, the UCR time series anomaly archive, and also discussed potential issues with the evaluation metrics. Finally, the work of \cite{Paparrizos2022TSBUADAE} point out the lack of consensus regarding the appropriate datasets for benchmarking TSAD algorithms and present a benchmark suite derived from a combination of previous TSAD datasets and transformed classification datasets, which have been subjected to various transformations to increase the complexity and difficulty of the benchmark. They include several evaluation metrics in their work to provide a comprehensive evaluation of the TSAD algorithms.

In this paper, we aim to fill the gap in the literature by providing a comprehensive review of the evaluation metrics used and proposed in the field of time series anomaly detection. To the best of our knowledge, no prior works have offered a thorough overview of all the metrics used in the field. The main contributions of this paper are:

\begin{itemize}
    \item A comprehensive description of the existing evaluation metrics, highlighting their key properties, both desirable and undesirable.
    \item A novel and structured taxonomy of the metrics, based on their calculation methods, to facilitate understanding and comparison. To the best of our knowledge, this is the first time a systematic taxonomy for TSAD evaluation metrics is defined.
    \item An in-depth analysis of the impact of the choice of evaluation metric through a set of hypothetical case studies.
    \item A clear summary of each metric in terms of a set of defined properties.
\end{itemize}

In Section~\ref{sec:background} we define and introduce terms and concepts central to the topic of evaluating TSAD algorithms.
We state the scope and limitations of this work in Section~\ref{sec:method}.
In Section~\ref{sec:properties} we define 10 different properties distinguishing the metrics, all of which are presented and described briefly in Section~\ref{sec:metrics}. In Section~\ref{sec:metrics} we also present the taxonomy of these metrics.
Section~\ref{sec:experiments} presents a series of case studies for testing the properties of the metrics, resulting in a categorization of the metrics in Section~\ref{sec:categorization}, based on the properties from Section~\ref{sec:properties}. Finally, we summarize our findings and draw some conclusions in Section~\ref{sec:conclusion}.

\section{Background}
\label{sec:background}

In this section, we provide an overview of the fundamental concepts necessary to understand the subsequent discussion in this work.

\paragraph{Time series}
A time series is a sequence of numbers or vectors, indexed by the time. We will refer to each time step as a point. Although not apparent in the definition, the underlying assumption when working with time series, is that the value of the points are dependent on the time variable.

\paragraph{Time series anomaly}
An anomaly in a time series is defined in various ways \cite{Schmidl2022AnomalyDI}, but is in general a point or a subsequence of contiguous points with unexpected or abnormal values. We refer to the subsequence as an anomalous event, and each point in it as an anomalous point - not to be confused with a point anomaly, a term often used for events of length 1. 
Contrasting anomaly detection in independent data, the abnormality may stem from unsatisfied expectations of the time dependency. That is, a point can have a normal value for the time series in general, but anomalous in the context of its preceding values\footnote{Several works operate with different classes of time series anomalies \cite{Kovcs2019EvaluationMF,Goswami2022UnsupervisedMS,Lai2021RevisitingTS,Choi2021DeepLF}, some of which considers if an anomaly is outside the normal values for all points, or just its temporal context.}. Furthermore, what is considered as anomalies depends on the domain and origin of the time series. Finally, it is often unclear just how anomalous an event should be in order to be considered an anomaly. 
This lack of an exact definition of time series anomalies is some of the reason it is difficult to come up with reliable evaluation metrics.

\paragraph{Time series anomaly detection (TSAD)}
The goal of TSAD is to identify anomalies in a time series.
While a variety of techniques exist for detecting anomalies in time series data, a detailed review of which can be found in the work of  \cite{Schmidl2022AnomalyDI}, ranging from simple to complex and encompassing both machine learning and other approaches, it is not in the scope of this paper to discuss these techniques. Rather, our aim is to provide a comprehensive overview of the metrics used to evaluate these methods and offer a taxonomy of metrics based on their properties. 
In TSAD, the input data is typically a the time series of data points and the output is a prediction indicating which instances are anomalous. In our work we will refer to the output of the detection algorithm as \textit{prediction}.

\paragraph{Evaluation}
Evaluation is the task of assigning a score to each prediction, such that a higher (or lower) score means that the prediction is better.
Since anomalies are rare events and can have different characteristics, detectors are usually evaluated on different datasets in order to have a wider spectrum of possible anomalies. 
In order to easily and objectively sort anomaly detectors in terms of performance, the score must be a single scalar.
While is often useful to use several evaluation metrics, to get insights about which detector performs well in certain scenarios, we consider this another task, which we refer to as performance analysis, as opposed to performance evaluation.

\paragraph{Labels}
Evaluation is done by comparing the prediction to a time series of binary labels, that represents the ground truth (GT) of which points are anomalous or not.
Note that the use of binary labels is a source for several kinds of errors and inaccuracies - when an anomaly starts, ends, and what even should be considered anomalous is a question that rarely has a definite answer, except for synthetical data. Therefore, there are several different labelling strategies, that will lead to quite different labels on the same dataset - e.g. the Numenta labelling strategy discussed in Section~\ref{subsec:metrics:other}.
Furthermore, when labels are made manually by humans, they will often have inconsistencies.

Changes in labels will necessarily affect the evaluation scores, especially if an event is included or excluded, as there are usually very few anomalies.
The impact of slight changes in length and position of events however, highly depend on the metric, and will be discussed and tested later in this article.

Due to high variability in both what is considered as anomalies, and how they are labelled, the relevance of results on data from across domains is not obvious. When selecting a detector for use on a specific TSAD task, one should evaluate detectors on a dataset with both similar time series, anomalies, and a labelling strategy in line with the desired output of the detection algorithm\footnote{An alternative approach is unsupervised model selection, as described in \cite{Goswami2022UnsupervisedMS}. They present three ways to select the best model based on datasets without labels - by considering prediction/reconstruction error, model centrality and performance on synthetically injected anomalies. The two former methods skips the need for the kind of evaluation metrics presented in this paper altogether.}.

\subsection{Thresholding}
\label{subsec:thresholding}

\begin{figure}
    \centering
    \includegraphics[width=0.7\textwidth]{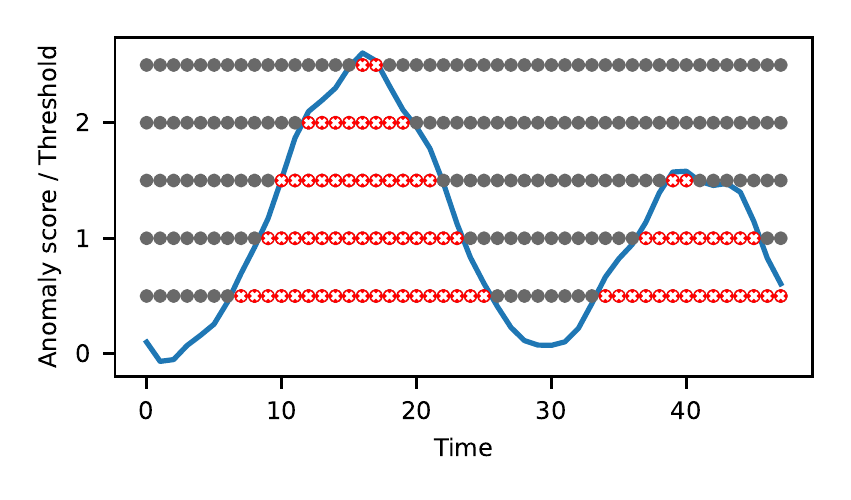}
    \caption{Visualization of thresholds. Each line of dots represent a possible binary prediction from the anomaly score, if the threshold is set to its respective value on the y-axis. Lowering the threshold increases the number of anomalous points predicted. \legendanomaly
    }
    \label{fig:thresholding}
\end{figure}

An anomaly detector outputs an anomaly score, a time series with scalar values indicating how anomalous each time point is. In order to get a binary prediction, only time steps with anomaly score higher than some threshold are considered anomalous.
This is visualized in Figure~\ref{fig:thresholding}.

There are several ways of choosing a threshold, some fully automatic, like the non-parametric dynamic thresholding introduced in the work of \cite{Hundman2018DetectingSA}, others as simple as just choosing $$\text{mean}+n\cdot
\text{std}$$ for some $n$ \cite{Geiger2020TadGANTS, Liu2022TimeSA}\footnote{As different methods have anomaly scores with different statistics, this may not be fair when comparing different methods. As an example, a method based on reconstruction error will have different outcomes depending on whether it uses MSE or RMSE error.}.

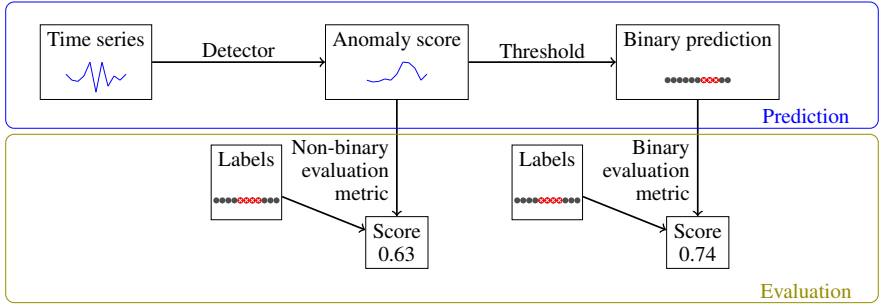
\begin{figure*}
\centering
\scalebox{0.8}{
\begin{tikzpicture}[align=center]
\node (a1)[draw, shape=rectangle, color=black] at (-5,0) {\textcolor{black}{Time series}\\ \\};
\foreach \i/\a in
{-5.5/-0.2, -5.4/-0.3, -5.3/-0.32, -5.2/-0.23, -5.1/0, -5/-.5, -4.9/0, -4.8/-.4, -4.7/-0.23, -4.6/-0.3, -4.5/-0.2}{
  \coordinate (now) at (\i,\a) {};
  \ifthenelse{\equal{\i}{-5.5}}{}{
    \draw[-, blue] (prev) -- (now);
  }
  \coordinate (prev) at (\i,\a) {};
}
\node (a3)[draw, shape=rectangle, color=black] at (0,0) {\textcolor{black}{Anomaly score}\\ \\};
\foreach \i/\a in
{-0.5/-0.3, -0.4/-0.33, -0.3/-0.32, -0.2/-0.28, -0.1/-0.3, 0/-.2, 0.1/0, 0.2/-0.01, 0.3/-0.1, 0.4/-0.3, 0.5/-0.2}{
  \coordinate (now) at (\i,\a) {};
  \ifthenelse{\equal{\i}{-0.5}}{}{
    \draw[-, blue] (prev) -- (now);
  }
  \coordinate (prev) at (\i,\a) {};
}
\node (a5)[draw, shape=rectangle, color=black] at (5,0) {\textcolor{black}{Binary prediction}\\ \\};
\nomalies[first x=4.5, second x=4.6, last x=5.59, y=-0.3, radius=0.05]
\anomalies[first x=5.1, second x=5.2, last x=5.35, y=-0.3, radius=0.05]

\draw[->, thick] (a1) edge node {Detector\\} (a3);
\draw[->, thick] (a3) edge node {Threshold\\} (a5);

\node (l1)[draw, shape=rectangle, color=black] at (-2.5,-2) {\textcolor{black}{Labels}\\ \\};

\nomalies[first x=-3, second x=-2.9, last x=-1.95, y=-2.3, radius=0.05]
\anomalies[first x=-2.6, second x=-2.5, last x=-2.25, y=-2.3, radius=0.05]

\node (l2)[draw, shape=rectangle, color=black] at (2.5,-2) {\textcolor{black}{Labels}\\ \\};
\nomalies[first x=3, second x=2.9, last x=1.95, y=-2.3, radius=0.05]
\anomalies[first x=2.4, second x=2.5, last x=2.75, y=-2.3, radius=0.05]

\node (b3)[draw, shape=rectangle, color=black] at (0,-3) {\textcolor{black}{Score}\\0.63};
\node (b5)[draw, shape=rectangle, color=black] at (5,-3) {\textcolor{black}{Score}\\0.74};
\draw[->, thick] (a3) -- (b3) node[left, pos=0.6, align=right] {Non-binary\\evaluation\\metric};
\draw[->, thick] (a5) -- (b5) node[left, pos=0.6, align=right] {Binary\\evaluation\\metric};

\draw[->, thick] (l1) -- (b3);
\draw[->, thick] (l2) -- (b5);

\draw[rounded corners, blue] (-6.5,-1.1) rectangle (8,1);
\node (q1) [blue] at (6.8, -0.9) {Prediction};
\draw[rounded corners, olive] (-6.5,-1.2) rectangle (8,-4);
\node (q1) [olive] at (6.8, -3.8) {Evaluation};
\end{tikzpicture}
}
    \caption{Time series anomaly detection pipeline. The binary predictions are found by applying a threshold to the anomaly score outputted by the detector. The evaluation is done at one of two points, either using the anomaly score, or the binary predictions. In both cases, binary labels are used for comparizon.}
    \label{fig:pipeline}
\end{figure*}

Anomaly detections can be evaluated either before or after the thresholding, as shown in Figure~\ref{fig:pipeline}. We define \textit{binary evaluation metrics} as metrics evaluating the binary prediction, and \textit{non-binary evaluation metrics} as those evaluating the anomaly score. While the latter class uses the anomaly score as input, thresholding is still done, but as part of the metric. This usually involves calculating a score at several or all thresholds\footnote{By all thresholds we mean all thresholds that yield unique sets of anomaly points - at most one more than the number of time points in the time series.}, and either choosing the optimal score or combining the scores.

The difference between the classes may seem subtle, but involves a foundational difference in what is evaluated. Binary metrics evaluate the combination of the detector and the thresholding strategy, while non-binary metrics aim at only evaluating the detector.
The argument for the latter class is that thresholding is a seperate issue, and since any detector can be used with any thresholding strategy, detectors should be compared indepentently of this choice. Using non-binary metrics ensure that thresholding is done equally for all detectors, which might be more fair.
However, as thresholding is indeed a part of the non-binary metrics as well, this class of metrics is not independent of thresholds, but rather a compromize between them - and the metric might focus overly on irrelevant thresholds.
Finally, as thresholding is done in practice, it may make more sense to evaluate the whole pipeline in unison, using a binary metric. This also allows for using the thresholding strategies that work well with specific detectors.

\subsection{Traditional evaluation metrics}

\begin{figure}[]
    \centering
    \tabcolsep=0.2cm
    \renewcommand{\arraystretch}{1.5}
    \begin{tabular}{ccc|c|}
        &&\multicolumn{2}{c}{Prediction}\\
        \cline{3-4}
&
\multicolumn{1}{c|}{~}

&
    \begin{tikzpicture}
        \anomaly{0}{0}{0.1}
    \end{tikzpicture}
&
\begin{tikzpicture}
  \nomaly{0}{0}{0.1}
\end{tikzpicture}\\
        \cline{2-4}
\multirow{2}{*}{\rotatebox{90}{Label}}

&
\multicolumn{1}{|c|}{
    \begin{tikzpicture}
        \anomaly{0}{0}{0.1}
    \end{tikzpicture}
}
&TP&FN\\
        \cline{2-4}
         & 
\multicolumn{1}{|c|}{
    \begin{tikzpicture}
        \nomaly{0}{0}{0.1}
    \end{tikzpicture} 
}
&FP&TN\\
        \cline{2-4}
    \end{tabular}
    \caption{The confusion matrix for anomaly detection. Each point can have one of two label values, and one of two prediction values, resulting in four different classes of points. \legendanomaly}
    \label{tab:confusion_matrix}
\end{figure}

\label{subsec:traditional_metrics}
Before embarking on the time series specific metrics, it is beneficial to understand some of the evaluation metrics used for anomaly detection and classification in general.
Common for most of the evaluation metrics is the use of the confusion matrix.
The confusion matrix considers the possible combinations of binary prediction and labels, and includes the number of 
\begin{itemize}
    \item true positives (TP): points that are labelled and predicted as anomalies, 
    \item false positives (FP): points that are labelled normal but predicted as anomalous, 
    \item false negatives (FN): points that are labelled anomalies but predicted normal, 
    \item true negatives (TN): points that are labelled and predicted as normal,
\end{itemize}
as seen in Figure~\ref{tab:confusion_matrix}.
We refer to these four numbers as counting metrics. They are not used for evaluation directly, but are needed for calculating the following metrics:

\textbf{Accuracy} is the fraction of correctly predicted points, i.e. $\frac{TP + TN}{TP+TN+FN+FP}$. Although simple, and to the uncritical eye informative, this metric should not be used for classifications with imbalanced classes, which anomaly detection is by definition. Since most points are normal, a prediction of only normal points will get a high accuracy despite not being useful at all.

\textbf{Recall}, also known as sensitivity and true positive rate, is the fraction of true anomalies that are correctly classified, i.e. $\frac{TP}{TP+FN}$. False positives are not penalized, thus predicting all points as anomalous will get a perfect recall of 1. For this reason, recall is usually not used on its own.

\textbf{Precision} is the fraction of anomalous predictions that are actual anomalies, i.e. $\frac{TP}{TP+FP}$. Like recall, this is not used on its own, since false negatives are not penalized, and only marking the most obvious anomaly will be the best strategy.

\textbf{$f_1$-score} is the harmonic mean of precision and recall, $\frac{2PR}{P+R}$. The priorization of recision and recall is a trade-off - strict threshold yield few predicted anomalies, thus high precision but low recall, and vice versa. Depending on the situation, it might be (very) preferable to have a false positive than a false negative, or opposite. Thus, a more general definition is $f_{\beta}$-score, defined by $\frac{(1+\beta)PR}{R+\beta^2P}$. The value of $\beta$ is the chosen so that the score reflects the relative importance of precision and recall.
We will use $\beta=1$ in the examples of this paper, as is also common in the literature when comparing methods, but we highlight that an informed choice should be made for this parameter when using this metric for real world problems.

\textbf{False positive rate} is the fraction of normally predicted points that are actually anomalies, $\frac{FP}{FP+TN}$. Contrary to recall, optimal score is obtained by predicting all the points as normal. This is used for calculating the \aucroc~score described in Section~\ref{subsec:auc}.

\textbf{Precision@k} is the precision of the $k$ points with highest anomaly score. Although this is just the precision with a specific thresholding strategy, it deserves some extra attention. This is because, since the denominatior $TP+FP=k$ is predetermined, false positives are indeed penalized. Thus this becomes a valid metric in itself, not needing to be combined with recall. In fact, recall@k is the \textit{same value} as precision@k, except for a predetermined constant $\frac{k}{TP+FN}$\footnote{Note that this is not true for all the redefined versions of precision and recall presented later in this paper.}.
Compared to the above metrics, this strategy requires a number of anomalies $k$ instead of a threshold. This may be a simpler and more intuitive choice - a common practice is to use the number/fraction of anomalies in the dataset. It may also be more fair when comparing methods with differently distributed anomaly score, than many other threshold selection strategies.

The metrics above are often used for time series without adaptation, by regarding every time stamp individually. 
A large number of the evaluation metrics designed specifically for time series are versions of precision and recall that are redefined to handle events in a different way, either by a redefined confusion matrix, or by redefining precision and recall to not use the counting metrics at all. These are then usually used either to calculate f-score, or an AUC score, which we will discuss in Section~\ref{subsec:auc}.

\section{Method}
\label{sec:method}

Several choices were made for the purpose of limiting the scope of this paper, and keep it concise.
We did not include metrics from similar domains like time series classification, anomaly detection for non-time series, or change point detection. The latter, although similar to TSAD, only contain point anomalies.

Furthermore, we only consider single scalar metrics aimed at performance evaluation for detector selection, and not supplementary statistics for performance analysis.
This means we will not consider the numerous variants of precision and recall as their own metrics, only as part of the $f_1$-score or the \aucpr~score described in section \ref{subsec:auc}.
Precision and recall are occasionally used for detector selection in situations where false positives and negatives have very different costs. However, due to the simple optimal strategies described in Section \ref{subsec:traditional_metrics}, $f_{\beta}$ with a large/small $\beta$ is a much better alternative.
Other interesting statistics excluded by this choice are \textit{early detection} \cite{Buda2017ADEAE}, \textit{before/after true positives} \cite{Nalepa2022EvaluatingAF} and \textit{alert delay} \cite{Xu2018UnsupervisedAD}. Combinations of these statistics with other statistics could result in evaluation metrics with valuable properties.
ROC- and PR-curves (see section \ref{subsec:auc}) are often used for visualising properties of the anomaly score. We will only consider these for the purpose of calculating the much used single scalar AUC metrics.

There are several ways to vary each metric, by using techniques from one metric on one of the others. Indeed, some of the metrics are indeed modified versions of another metrics, in such a way that all the other metrics could be modified in that same way. Studying all these combinations is not feasible without expanding the work substantially, so we will only study such modifications in their originally proposed, or most used, form. 
This should give an idea of the effect of the modification. Readers that are interested in a specific metric, either one included here, or that could be made by combining ideas from the ones included, are encouraged to conduct their own experiments.

Finally, for obvious reasons, we only consider metrics that \textit{either} are rigorously defined in their original paper, \textit{or} have open source implementations available.

\section{Properties}
\label{sec:properties}
In order to systematically evaluate the various metrics used in TSAD, we have defined several properties that differentiate the metrics. It is important to note that these properties are in general not inherently positive or negative, but rather the desirability of each property depends on the specific context and scenario. To achieve this, we defined a set of properties for these metrics and analysed how these properties affect the results of the metrics. We have organized the properties into three categories: (1) \textbf{Preferences}: properties related to the predictions generated by the metrics, (2) \textbf{Requirements}: requirements for utilising the metrics, and (3) \textbf{Suitability}: properties regarding the general suitability of the metrics in TSAD applications.

\subsection{Preferences}
As time series anomaly detection methods rarely produce perfect prediction, a good metric needs to be able to prefer the best imperfect detection available, for the situation for which the detector will be used.
We listed five properties regarding what kind of prediction are preferred by the metrics. 

\textbf{Early detection}. In the literature and in practical scenarios, two distinct contexts can be identified. In the first context, detection of a possible anomaly should occur as soon as possible \cite{Lavin2015EvaluatingRA}, such as when anomaly detection is used in real-time systems where an anomaly indicates there is an issue requiring immediate attention. In these cases, detecting the anomaly at a late stage is of no value since it is too late to rectify the problem. In the second context, data is analysed offline, or on a much larger time scale, where detection and reaction time is far greater than anomaly length, e.g. for diagnosis based on ECG monitoring \cite{Chuah2007ECGAD, Sivaraks2015RobustAA}. In these cases, the differences between early and late detection are of no practical relevance. 

\textbf{Long anomalies}.
Longer anomalies could indicate more serious problems which are also more important to detect, or they might just indicate more subtle anomalies which are harder to locate \cite{Kim2022ASO}. The shortest anomalies might also be the most important ones, e.g. if they indicate serious problems that were fixed quickly, while the less serious ones were ignored and therefore lasted much longer.
In most metrics, the contribution of an anomaly to the final score is either proportional to its length, or independent of its length. As many commonly used TSAD datasets have both long anomalies and single point anomalies, this difference has a great impact. 

\textbf{Short predicted anomalies}. Some detectors, e.g. window-based ones\footnote{Window-based detection methods evaluate the abnormality of windows (contiguous subsequences of a predefined length) of the time series instead of each point separately, and then aggregate the results from all the windows into an anomaly score.}, produce anomaly scores with a short peaks, while other methods produce wider areas of high anomaly score. The latter will generally result in longer predicted events. This might not have a big impact on the value of the prediction, but some metrics have a strong preference for short predicted events, independent of the length of the labelled anomaly.

\textbf{Partial detection}. The ability to detect a subset of the anomaly (referred to as "partial detection") can be more important than correctly detecting its exact span time (referred to as "covering"). According to the work of \cite{Xu2018UnsupervisedAD}, an operator receiving an alert of an anomaly will investigate the data manually, and the manual inspection will be the determining factor going forward, rendering the exact location and duration of the detection less relevant. However, \cite{Hwang2019TimeSeriesAP} notes that the operator may not necessarily find the anomaly if it is subtle and of a much longer duration than the detection, which would make the location and duration of the detection significant.

\textbf{Proximity}. The start and end of an anomaly is often unclear \cite{Kim2022ASO}, and when manually labelled, the labels might not be very reliable \cite{Wu2022CurrentTS}. Furthermore, a predicted event being off by a few time steps might still be very useful. Indeed, window-based detection methods might report the anomaly at either end of the window \cite{Wu2022CurrentTS}. In offline anomaly detection, this should not overly effect the score.
For these reasons, detecting an anomaly close to a labelled anomaly should be valued by the metric.

\subsection{Requirements}

Different metrics use different input, and require different degree of parameter specifications.

\textbf{Require no thresholding} \cite{Paparrizos2022VolumeUT}, as discussed in Section~\ref{subsec:thresholding}.

\textbf{Require few parameters} \cite{Paparrizos2022VolumeUT}. Correctly specifying numerous of parameters to reflect specific needs can be resource demanding. Furthermore, it is easier to compare results across research papers when they do not use different parameters. Nevertheless, TSAD tasks vary greatly, and parameters offer flexibility needed for a metric to be useful for most specific cases.

\subsection{Suitability}
The different metrics are also meant for different kinds of use, and might not always be suitable. We list up three properties related to the suitability of the metrics in different use cases. 

\textbf{Time aware}. Metrics not made for time series or sequential data do not use the chronology when calculating the score. Awareness of the labels and predictions of surrounding points is necessary for capturing the underlying time dependency specific for time series.

\textbf{Insensitivity to True Negatives}. Given that anomalies are by definition rare events, a low score should be given when no anomalies are detected, even though the prediction is correct most of the time. Furthermore, it is useful not to be affected by how large the portion of true negative time points is, as this is a rather uninformative part of the data. 

\textbf{Generality}. A metric that is appropriate for many real scenarios, is also useful for research that is not domain/problem-specific, as the results would be relevant for more situations. However, since TSAD is used for such a large span of different problems, no metric can suit all situations.

Finally, we highlight that there are several possible desirable properties not included here due to our scope limitations. Such properties can be valuable insights about the performance of the method, e.g. where it performs well or not \cite{Huet2022LocalEO}, or how early the detections are \cite{Nalepa2022EvaluatingAF},
or, for multivariate time series, which signals are the most involved in the anomaly. The latter property is often measured using distinct explainability measures \cite{Su2019RobustAD,Chen2021DAEMONUA,Garg2022AnEO,Li2021MultivariateTS}.

\section{TSAD Evaluation Metrics: a Taxonomy}

\label{sec:metrics}
In this section, a comprehensive examination of the evaluation metrics found through our research is presented. The metrics are divided into two categories, binary metrics in Section~\ref{subsec:binary_metrics} and non-binary metrics in Section~\ref{subsec:Nonbinary}. For each category, a taxonomy based on their definitions is introduced, followed by a description of each metric including their capabilities and potential limitations in utilization.

A rigorous definition of each metric is not included in this study, as some of them are quite complex, with details not necessary for this work.
Readers are referred to the cited literature for further information. However, an effort has been made to provide a concise and intuitive understanding of the metrics. In addition, the most noteworthy, distinctive, or potentially problematic characteristics of the metrics are also discussed.

\subsection{Binary evaluation metrics}
\label{subsec:binary_metrics}
We define \textit{binary evaluation metrics} as metrics evaluating binary predictions, where each data point is classified as either normal or anomalous, aligning with the binary labelling.

\begin{figure*}[ht!]
    \centering
\begin{tikzpicture}[align=center]

\draw[rounded corners, black] (2.8,0.8) rectangle (5.2,4.9);
    \node [draw, shape=rectangle, fill=red!40] at (4,4.2) {Point-based\\metric(s)};
    \node [draw, shape=rectangle, fill=blue!40] at (4,3) {Event-based\\metric(s)};
    \node [draw, shape=rectangle, diagonal fill={red!40}{blue!40}] at (4,1.6) {Point- and\\event-based\\metric(s)}; 

\node (b2) [draw, shape=rectangle] at (-0.5,4) {Binary\\evaluation metrics};
\node (c3) [draw, shape=rectangle] at (-2,2) {f-scores};
\node (c4) [draw, shape=rectangle] at (1,2) {Other\\metrics};
\node (e3) [draw, shape=rectangle, diagonal fill={red!40}{blue!40}] at (2.5,0) {\nab};
\node (e4) [draw, shape=rectangle, fill=red!40] at (1.5,0) {\tempdist};

    \node (e1) [draw, shape=rectangle] at (-3.5,0) {Precision/recall\\based on\\confusion matrix};
    \node (e2) [draw, shape=rectangle] at (-0.5,0) {Redefined\\precision/recall};
    \node (h1) [draw, shape=rectangle] at (-5,-3) {Point-wise\\counting metrics};
    \node (h2) [draw, shape=rectangle] at (-2,-3) {Adjusted\\point-wise\\counting metrics};
    \node (h3) [draw, shape=rectangle] at (1,-3) {Redefined\\counting metrics};
    \node (h4) [draw, shape=rectangle, fill=blue!40] at (4,-3) {\rf{bias}{\alpha}, \af,\\ \taf{\alpha}{\delta}{\theta}, \etaf{\theta_p}{\theta_r}};
    \node (j5) [draw, shape=rectangle, fill=blue!40] at (3,-5) {\segf};
    \node (j1) [draw, shape=rectangle, fill=red!40] at (-5,-5) {\pwf};
    \node (j2) [draw, shape=rectangle, fill=red!40] at (-2,-5) {\paf, \dtpaf[\beta]{k}, \pakf[\beta]{K}, \lsf[1]{n}};
    \node (j3) [draw, shape=rectangle, fill=red!40] at (1,-5) {\ttolf{\tau}}; 
    \node (j4) [draw, shape=rectangle, diagonal fill={red!40}{blue!40}] at (2,-5) {\cf}; 

\draw[->, thick] (b2) -- (c3);
\draw[->, thick] (b2) -- (c4);

\draw[->, thick] (c3) -- (e1);
\draw[->, thick] (c3) -- (e2);
\draw[->, thick] (c4) -- (e3);
\draw[->, thick] (c4) -- (e4);

\draw[->, thick] (e1) -- (h1);
\draw[->, thick] (e1) -- (h2);
\draw[->, thick] (e1) -- (h3);
\draw[->, thick] (e2) -- (h4);

\draw[->, thick] (h1) -- (j1);
\draw[->, thick] (h2) -- (j2);
\draw[->, thick] (h3) -- (j3);
\draw[->, thick] (h3) -- (j4);
\draw[->, thick] (h3) -- (j5);

\end{tikzpicture}
    \caption{
    A taxonomy of binary evaluation metrics. A large number of these are f-scores based on various definitions of precision and recall.
    Precision and recall can be defined in many ways. Compared to the original point-wise definition, the difference can be present in the point-wise predictions, the counting metrics (TP, FP, TN, FN) or the formulas for precision and recall. The metrics can also be divided into point-based and event-based metrics, that count respectively individual points or contiguous events when aggregating to the total score. The point- and event-based metrics use both of these methods for parts of the total score.
    }
    \label{fig:binary}
\end{figure*}
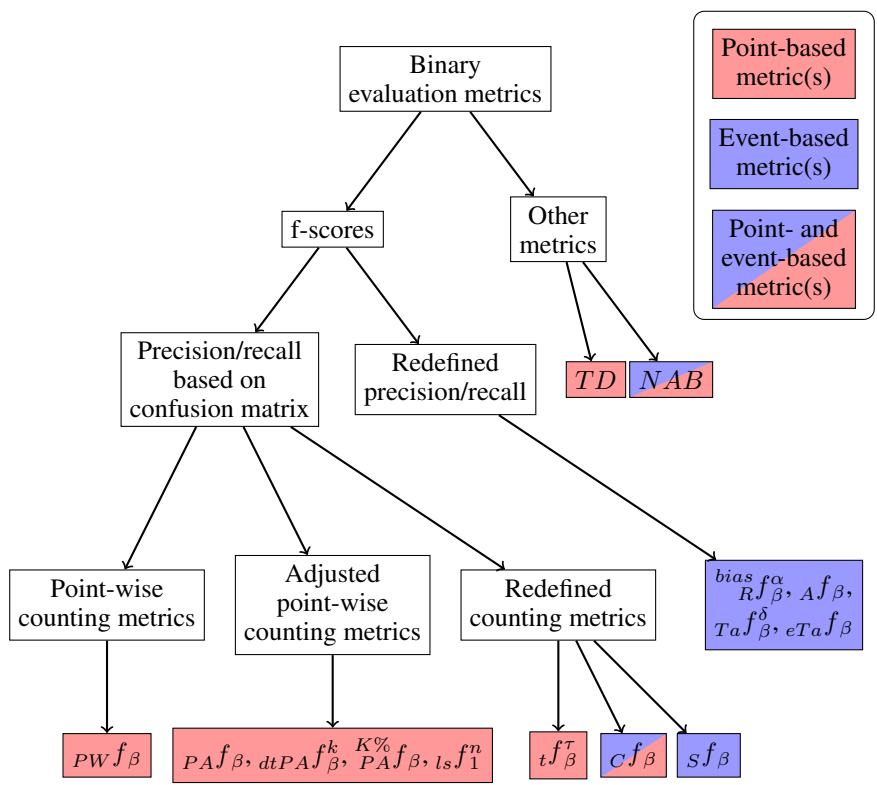

Figure~\ref{fig:binary} shows the proposed taxonomy of binary evaluation metrics, based on how their definitions use counting metrics (TP, TN, FP, FN), precision, recall or f-score. This information is relevant when combining techniques from different metrics, as such techniques may only work on one type of metrics. 

Most of the metrics are based on the f-score, with some modification of the definitions. 
The \textit{point-wise counting metrics} (Section~\ref{subsec:metrics:pointwise}) is the f-score based on counting metrics calculated in each time point. The \textit{adjusted point-wise counting metrics} (Section~\ref{subsec:metrics:pointadjusted}) also use counting metrics in each point, but an adjustment is done to the prediction \textit{before} the counting, in order to be more suited for anomalous events. For the \textit{redefined counting metrics} (Section~\ref{subsec:metrics:redefined_counting}) the counting itself is done in some other way. \textit{Redefined precision/recall} (Section~\ref{subsec:metrics:redefined_PR}) are not based on counting metrics at all, but calculated from some different formulas. They still use the terms precision and recall because the base concepts are the same.
Finally, the \textit{other metrics} (Section~\ref{subsec:metrics:other}) are not based on f-score at all. 

The metrics are aslo categorized based on their calculation approach, as either point-based or event-based. All the metrics are computed by aggregating the contributing parts of the time series, but in different ways. The \textit{point-based metrics} evaluate each time point individually, whereas the \textit{event-based metrics} evaluate entire events as a single subscore, regardless of the number of time points it comprises. This distinction has significant implications for what is considered a good prediction, as will be demonstrated in Section~\ref{sec:experiments}.
Some metrics calculate part of the score in a point-based way and part event-based. We name these metrics \textit{point- and event-based}.

\subsubsection{Point-wise metrics}
\label{subsec:metrics:pointwise}

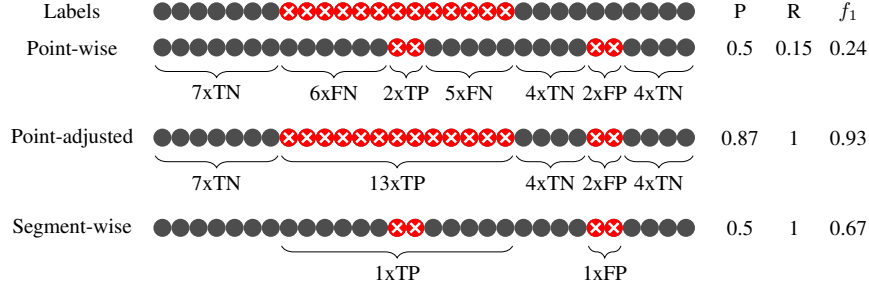
\begin{figure*}[ht!]
    \centering
\scalebox{0.8}{
  \begin{tikzpicture}[]
    \node (GT) at (-1.5,0) {Labels};
    \nomalies[first x=0, second x = 0.3, last x = 9, y = 0, radius=0.15]
    \anomalies[first x=2.1, second x = 2.4, last x = 5.9, y = 0, radius=0.15]

    \node (GT) at (-1.5,-0.6) {Point-wise};

    \nomalies[first x=0, second x = 0.3, last x = 9, y = -0.6, radius=0.15]
    \anomalies[first x=3.9, second x = 4.2, last x = 4.2, y = -0.6, radius=0.15]
    \anomalies[first x=7.2, second x = 7.5, last x = 7.5, y = -0.6, radius=0.15]

\draw [black, decorate,decoration={brace,amplitude=.2cm, mirror}, xshift=0cm, yshift=0cm] (-0.12,-0.9) -- +(0:2.04) node [black, midway, xshift=0cm, yshift=-.45cm] {7xTN};
\draw [black, decorate,decoration={brace,amplitude=.2cm, mirror}, xshift=0cm, yshift=0cm] (1.98,-0.9) -- +(0:1.74) node [black, midway, xshift=0cm, yshift=-.45cm] {6xFN};
\draw [black, decorate,decoration={brace,amplitude=.2cm, mirror}, xshift=0cm, yshift=0cm] (3.78,-0.9) -- +(0:0.54) node [black, midway, xshift=0cm, yshift=-.45cm] {2xTP};
\draw [black, decorate,decoration={brace,amplitude=.2cm, mirror}, xshift=0cm, yshift=0cm] (4.38,-0.9) -- +(0:1.44) node [black, midway, xshift=0cm, yshift=-.45cm] {5xFN};
\draw [black, decorate,decoration={brace,amplitude=.2cm, mirror}, xshift=0cm, yshift=0cm] (5.88,-0.9) -- +(0:1.14) node [black, midway, xshift=0cm, yshift=-.45cm] {4xTN};
\draw [black, decorate,decoration={brace,amplitude=.2cm, mirror}, xshift=0cm, yshift=0cm] (7.08,-0.9) -- +(0:0.54) node [black, midway, xshift=0cm, yshift=-.45cm] {2xFP};
\draw [black, decorate,decoration={brace,amplitude=.2cm, mirror}, xshift=0cm, yshift=0cm] (7.68,-0.9) -- +(0:1.14) node [black, midway, xshift=0cm, yshift=-.45cm] {4xTN};
      
    \node (GT) at (-1.5,-2.1) {Point-adjusted};
    \nomalies[first x=0, second x = 0.3, last x = 9, y = -2.1, radius=0.15]
    \anomalies[first x=2.1, second x = 2.4, last x = 5.9, y = -2.1, radius=0.15]
    \anomalies[first x=7.2, second x = 7.5, last x = 7.5, y = -2.1, radius=0.15]

\draw [black, decorate,decoration={brace,amplitude=.2cm, mirror}, xshift=0cm, yshift=0cm] (-0.12,-2.4) -- +(0:2.04) node [black, midway, xshift=0cm, yshift=-.45cm] {7xTN};
\draw [black, decorate,decoration={brace,amplitude=.2cm, mirror}, xshift=0cm, yshift=0cm] (1.98,-2.4) -- +(0:3.84) node [black, midway, xshift=0cm, yshift=-.45cm] {13xTP};
\draw [black, decorate,decoration={brace,amplitude=.2cm, mirror}, xshift=0cm, yshift=0cm] (5.88,-2.4) -- +(0:1.14) node [black, midway, xshift=0cm, yshift=-.45cm] {4xTN};
\draw [black, decorate,decoration={brace,amplitude=.2cm, mirror}, xshift=0cm, yshift=0cm] (7.08,-2.4) -- +(0:0.54) node [black, midway, xshift=0cm, yshift=-.45cm] {2xFP};
\draw [black, decorate,decoration={brace,amplitude=.2cm, mirror}, xshift=0cm, yshift=0cm] (7.68,-2.4) -- +(0:1.14) node [black, midway, xshift=0cm, yshift=-.45cm] {4xTN};

    \node (GT) at (-1.5,-3.6) {Segment-wise};

    \nomalies[first x=0, second x = 0.3, last x = 9, y = -3.6, radius=0.15]
    \anomalies[first x=3.9, second x = 4.2, last x = 4.2, y = -3.6, radius=0.15]
    \anomalies[first x=7.2, second x = 7.5, last x = 7.5, y = -3.6, radius=0.15]
\draw [black, decorate,decoration={brace,amplitude=.2cm, mirror}, xshift=0cm, yshift=0cm] (1.98,-3.9) -- +(0:3.84) node [black, midway, xshift=0cm, yshift=-.45cm] {1xTP};
\draw [black, decorate,decoration={brace,amplitude=.2cm, mirror}, xshift=0cm, yshift=0cm] (7.08,-3.9) -- +(0:0.54) node [black, midway, xshift=0cm, yshift=-.45cm] {1xFP};

    \node at (9.6,0) {P};
    \node at (10.5,0) {R};
    \node at (11.4,0) {$f_1$};

    \node at (9.6,-0.6) {0.5};
    \node at (10.5,-0.6) {0.15};
    \node at (11.4,-0.6) {0.24};

    \node at (9.6,-2.1) {0.87};
    \node at (10.5,-2.1) {1};
    \node at (11.4,-2.1) {0.93};

    \node at (9.6,-3.6) {0.5};
    \node at (10.5,-3.6) {1};
    \node at (11.4,-3.6) {0.67};

  \end{tikzpicture}
}
    \caption{Counting metrics are found in various ways by the different metrics. \pwf~considers each time point individually. As does \paf, but only after making an adjustment by expanding partially predicted anomalies. The segment-wise counting metrics used in \segf, on the other hand, consider events. 
    \legendanomaly
    }
    \label{fig:PA}
    
\end{figure*}

\textbf{Point-wise f-score (\pwf)}. 
One of the most straightforward evaluation metrics involves treating each time point as a single observation and calculating the f-score as outlined in Section \ref{subsec:traditional_metrics}. This approach is exemplified in Figure~\ref{fig:PA}.
Although not made for time series, the use of point-wise f-score is widely used in TSAD \cite{Ahmed2022RCADRC,Han2022LearningSL,Huang2022ASV,Feng2022UnsupervisedMA,Wang2022VariationalTA, Campos2021UnsupervisedTS,Deng2021GraphNN,
Bashar2020TAnoGANTS,Niu2020LSTMBasedVF,Chen2020UnsupervisedAD,Mamandipoor2020MonitoringAD, Hsieh2019UnsupervisedOA,
Li2019MADGANMA, 
Zhang2018ADN}.
It is a simple metric, making it easy to implement and the results simple to understand.
Also, methods are rewarded for predicting all the points that are actually labelled as anomalies, and none of the other - exactly what an anomaly detector should do - as opposed to some of the metrics we will describe below.
Nevertheless, as we will see in the experiments of Section~\ref{sec:experiments}, the uneven event weighting and lack of tolerance can be highly problematic.

\DeclareRobustCommand{\disc}[1]{
\begin{tikzpicture}
\fill[#1] (0,0) circle (1.2 mm);
\end{tikzpicture}
}

\subsubsection{Point Adjusted metrics}
\label{subsec:metrics:pointadjusted}

\textbf{Point Adjusted f-score (\paf)}. The point adjusted metrics were first introduced by \cite{Xu2018UnsupervisedAD}, and has been widely used in TSAD \cite{Xu2022CalibratedOC, Goswami2022UnsupervisedMS, Challu2022DeepGM, Huang2022ASV,Tuli2022TranADDT,Chen2022LearningGS, Li2021MultivariateTS,Feng2021TimeSA,Dai2021SDFVAESA, Chen2021DAEMONUA,Du2021GANBasedAD,Choi2021DeepLF, Zhao2020MultivariateTA, Audibert2020USADUA,Shen2020TimeseriesAD,Su2019RobustAD}.
They propose that if a single point within a true anomalous segment is accurately detected, a human operator can examine the segment and identify the entire anomaly. As a result, the entire contiguous segment is marked as anomalous in the prediction prior to calculating point-wise precision, recall, and f-score.

Previous works \cite{Audibert2020USADUA, Garg2022AnEO, Doshi2022RewardOP, Kim2022TowardsAR} have shown that this metric can provide overly optimistic scores even if multiple anomalies are missed. In fact, the work of \cite{Doshi2022RewardOP, Kim2022TowardsAR} demonstrated that random guessing outperforms state-of-the-art methods using this metric. The cause of this is a seemingly unintended flaw of the metric, which is illustrated in Figure 6. Despite the argument that the whole anomaly is detected if an operator receives an alert within the anomaly, which legitimizes a recall of 1, only half of the alerts were correct, so the precision of the prediction should be 0.5. However, after adjustment, it is close to perfect. The greater the discrepancy between the duration of labelled and predicted anomalies, the more severe the problem becomes\footnote{Interestingly, the paper of \cite{Xu2018UnsupervisedAD} first using this metric have very short anomalies, compared to some of the datasets used in the papers that adopted this metric.}. Calculating precision prior to adjustment would avoid this issue and produce a precision-recall pair that aligns with the reason for the adjustment and the meaning of precision and recall. Nevertheless, we instead suggest using the composite f-score (Section~\ref{subsec:metrics:redefined_counting}), a more appropriate metric in cases when a warning during an anomaly is sufficient.

\textbf{Delay thresholded Point Adjusted f-score (\dtpaf{k})}.
The works of \cite{Ren2019TimeSeriesAD} and \cite{Chen2021AJM} use an adaptation of the point-adjusted metrics, where a GT anomaly is only considered detected if an anomaly is predicted within the first $k$ time steps of the anomaly. If not, all the points in the anomaly are marked as false negatives, even the ones predicted as anomalous.
With this metric, precision can still be unreasonably high, but it is much more difficult to achieve this, and the random guessing strategy that prevail for \paf~will have a much harder time getting high scores with this metric.

\textbf{Point adjusted metrics at K\% (\pakf{K})}.
The work of \cite{Kim2022TowardsAR} suggests altering the point adjusted metric by requiring a portion $K\%$ of the anomaly to be detected in order to make the adjustment. As with \dtpaf, this effectively reduce the effectiveness of random guessing, and short detections in general.
Furthermore, as argued by \cite{Hwang2019TimeSeriesAP} and \cite{Hwang2022DoYK}, an expert receiving a short alert within a much longer anomaly might not be able to see the anomaly, but by requiring a substantial part of the anomaly to be detected, the chance that an expert would actually notice it is much larger.

\textbf{Latency and sparsity-aware f-score (\lsf{n})}.
The work of \cite{Abdulaal2021PracticalAT} note that the point adjustment metrics do not value early detections, and changes the algorithm to only adjust the values of a contiguous anomaly segment after the first TP. They also note that false positive points require more resources if they are spread out, than in some close proximity (so that it only requires attention once). The prediction is therefore down-sampled by a used-specified factor $n$.

This way of awarding earliness reflects situations where the negative effects of an anomaly, which is proportional to its length, is avoided after the point that it is detected.

\subsubsection{Redefined counting metrics}
\label{subsec:metrics:redefined_counting}

\textbf{Segment-wise f-score (\segf)}.
The work of \cite{Hundman2018DetectingSA} introduced a segment-wise precision, recall and f-score, where each contiguous segment of anomalous points is considered one event. Here one true positive is recorded for each true anomalous segment with at least one predicted anomalous point, one false negative for each of the rest of the true anomalous segments, and one false positive for any predicted anomalous segment without any true anomalous points. Figure~\ref{fig:PA} shows an example of this.
This metric is used by \cite{Geiger2020TadGANTS, Nalepa2022EvaluatingAF, Meng2020SpacecraftAD, Flaborea2022AreWC}.

A problematic property of this metric is that extending the length of a predicted anomaly will never give worse score, and often better.
Thus it favours detectors with long contiguous events, all the way to the extreme case:
Predicting every point in the time series as anomalous will give perfect precision and recall for any time series with at least one anomaly.

\textbf{Composite f-score (\cf)}.
The work of \cite{Garg2022AnEO} suggested using a combination of point-wise and segment-wise metrics, and proposes the composite f-score, defined as the harmonic mean of point-wise precision and segment-wise recall.
The point-wise precision ensures that false positive points are discouraged, whereas extra true positive points in an already partially detected anomaly is only awarded through the increased precision.

\textbf{Time tolerant f-score (\ttolf{\tau})}\footnote{A similar metric is used in the work of \cite{Thill2020TimeSE}. Their data only contain point anomalies, and it is not obvious how to generalize to events, so we do not include it in this study.}.
The work of \cite{Scharwachter2020} defines (point-wise) precision and recall with temporal tolerance $\tau$, essentially by counting it as a true positive when a predicted anomaly point is closer than $\tau$ to a labelled anomaly point.
They then show that while the recall and precision of their example prediction increase drastically with the tolerance, the scores of a random prediction increases more, and the statistical significance decreases substantially. Hence reporting results with temporal tolerance may be less significant than without, despite the scores looking more impressive. It should be noted, however, that their data contain many short anomalies. A tolerance of a few time steps will have a much larger impact on the random prediction score in with such a dataset, than with fewer or larger anomalies.
Although these evaluation metrics are not widely used, similar tolerance techniques are - either in the metric (as here), in the labelling of the data (as in \nab, explained in Section \ref{subsec:metrics:other} ) or in detectors padding their predicted events before outputting them. Such significance tests can be useful when determining how much temporal tolerance to use.

\subsubsection{Redefined Precision and Recall}
\label{subsec:metrics:redefined_PR}

\textbf{Range-based f-score (\rf{bias}{\alpha})}.
The work of \cite{Tatbul2018PrecisionAR} argues that point-wise precision and recall fail to address many aspects present in time series for anomaly detection, and introduce range-based precision and recall. This metrics have been used in \cite{Jacob2021ExathlonAB} and \cite{Meng2020SpacecraftAD}.
These are rather complex and highly customizable metrics, with a tunable weight and up to 6 tunable functions to enable aligning the score with the goal of the detection task. Thorough guidelines, defaults and examples are provided in \cite{Tatbul2018PrecisionAR}.
The score is based on using up to 4 concepts to calculate the score: \textbf{Detecting} the anomaly range with at least one anomaly point, while also \textbf{covering} as large a portion of the anomaly range as possible. High \textbf{cardinality}, i.e. number of predicted segments within one labelled anomaly, can be punished, and a function rewarding the \textbf{position} of a detected anomaly within a labelled one can be specified.
Although evaluation metrics that consider the relative positions of detection and label are mostly useful for rewarding early detection, in these metric they can also be set to e.g. rewarding detections at the middle or at the end of the labelled anomalies, which authors argue can be useful in certain cases, e.g. as a way of preventing false positive alarms.
We have not found the cardinality concept in any other TSAD evaluation metric, and thus we have not considered as a desirable property. This may be more relevant for change point detection \cite{Gensler2014NovelCT}. 

\textbf{Time series aware f-score (\taf{\alpha}{\delta}{\theta})}.
The work of \cite{Hwang2019TimeSeriesAP} propose time-series aware precision and recall metrics. These metrics are similar to range-based precision and recall, but they do not consider the concepts of cardinality and position. The metrics also require that a certain portion $\theta$ of the labelled anomaly must be correctly predicted for it to be counted as a correct detection. The authors note that determining the end of a labelled anomaly can be challenging, and therefore include a region of length $\delta$ following the labelled event, with a positive but decreasing score, to account for this. This reduces the reliance on correct labelling and prediction at the end of and shortly after the. A slightly altered version of this metric can be found in \cite{Kim2022ASO}, where the method for determining the length of ambiguous sections was changed.

\textbf{Enhanced time series aware f-score (\etaf{\theta_p}{\theta_r})}. The author of \cite{Hwang2022DoYK} highlights that previous evaluation metrics may reward detections that overlap with actual anomalies, even if they are either too long or too short to be useful. To address this issue, they propose a metric that considers both a detection score and an overlap score. The metric requires that a certain part of the actual anomaly be detected and a certain part of the detected anomaly be true. Two parameters can be adjusted to control these portions. The precision calculation includes a weighting function that weights each event by the square root, as a compromize between typical point-based and event-based weighting.

\textbf{Affiliation-based f-score (\af)}. 
The work of \cite{Huet2022LocalEO}, tackles problems commonly seen in existing metrics and introduces a distance-based metric as a solution. They calculate the average of the local precision and recall for each anomaly event. Local precision is calculated by averaging the distance between each predicted anomaly point and its closest labelled anomaly point, and expressing it as the probability of outperforming a random prediction. Recall is calculated similarly, using the average distance from each labelled anomaly point to its closest predicted anomaly. By using distance, this metric evaluates the proximity of predicted and labelled anomalies, even if they don't overlap. It also values detection over coverage in a natural way.
Finally, by scoring locally, the results are more interpretable, since each anomaly and its impact on the score can be evaluated separately.

\subsubsection{Other metrics}
\label{subsec:metrics:other}

\textbf{NAB score (\nab)}. The Numenta Anomaly Benchmark (NAB), presented by \cite{Lavin2015EvaluatingRA}, includes a dataset for time series anomaly detection and a novel evaluation metric. The metric penalizes false positive points with a negative value, and rewards true anomalous segments with a positive value based on how early the first anomalous point was predicted. The score is normalized by comparing it to a scenario where no anomalies are detected.

Since only one point of the true positive points in an anomalous segment contribute to the score, while every false positive point contribute negatively, the score favours detectors predicting short events - it is almost never beneficial to predict two contiguous points as anomalous.

NAB also introduced a different approach to labelling anomalies. This approach allows for rewarding detectors predicting anomalies before they occur\footnote{That is, before they are visible to the human labeller.}, and makes the score less dependent on the individuals who label the anomalies. A simplified explanation of the approach is provided here (see the work of \cite{nab_whitepaper} for the full details). The process involves a group of labellers deciding the first anomalous point for each anomalous event. Then, the points on both sides are marked anomalous, such that the original starting point is in the center of the event, each event has the same duration, and 10\% of the dataset is labelled as anomalous.
 
This strategy is similar to the temporal tolerance technique in \ttolf{\tau}. However, in this case it is part of the labelling strategy, instead of the metric.
Thus is it not a part of the implementation used in this paper, and we will not see the effects of this in the experiments in Section~\ref{sec:experiments}.

The \nab~score is not widely used \footnote{Despite very many metrics papers referring and comparing to this metric, we only found one paper using it for evaluation, by the same authors \cite{Ahmad2017UnsupervisedRA}.}, 
but their datasets are commonly used for benchmarking, using other metrics \cite{Schmidl2022AnomalyDI, Paparrizos2022TSBUADAE}. The labelling strategy of this dataset highlight the importance of not blindly combining arbitrary metrics and datasets. Due to the labelling strategy, at least 50\% of the points labelled anomalous were considered normal by the labellers, invalidating metrics counting each point individually, like \pwf.

\textbf{Temporal distance (\tempdist)}. 
Temporal distance, presented by \cite{Kovcs2019EvaluationMF}, is a very simple metric - summing the distances from each labelled anomaly point to the closest predicted anomaly point, and from each predicted anomaly point to the closest labelled anomaly point.
The lower score the better. 
This metric prioritizes roughly finding all the correct anomalies over getting the detection exact, since any false positive/negative raises the score by the distance to the closest anomaly. As long FPs and FN are punished roughly proportionally to their length, the metric prioritizes long labelled anomalies, and a method predicting short events has an advantage when predicting FPs.
The work of \cite{Kovcs2019EvaluationMF} presents two version of this metric\footnote{They also present several other metrics, although they do not pass the limitations presented in Section~\ref{sec:method}.}, by summing either absolute or squared distances. Generalizing this, one could use any positive power of the absolute distance. We will consider this exponent a parameter, and use 1 in all the experiments. High values of this parameter punish great distances more than low values.

\begin{figure*}[!]
    \centering

\begin{tabular}[t]{cccc}
\toprule
Labels:&
\begin{tikzpicture}[baseline=-\the\dimexpr\fontdimen22\textfont2\relax]
\draw (0,0) -- (7.4,0);
\nomalies[first x=0, second x=0.2, last x=7.59, y=0, radius=0.1]
\anomalies[first x=5.8, second x=6.0, last x=6.01, y=0, radius=0.1]
\anomalies[first x=7.0, second x=7.2, last x=7.21, y=0, radius=0.1]
\end{tikzpicture}
&\af[1]&\tempdist\\
\midrule
Prediction 1:&
\begin{tikzpicture}[baseline=-\the\dimexpr\fontdimen22\textfont2\relax]
\draw (0,0) -- (7.4,0);
\nomalies[first x=0, second x=0.2, last x=7.59, y=0, radius=0.1]
\anomalies[first x=5.0, second x=5.2, last x=5.21, y=0, radius=0.1]
\anomalies[first x=7.0, second x=7.2, last x=7.21, y=0, radius=0.1]
\end{tikzpicture}
&\textbf{0.91}&14\\
Prediction 2:&
\begin{tikzpicture}[baseline=-\the\dimexpr\fontdimen22\textfont2\relax]
\draw (0,0) -- (7.4,0);
\nomalies[first x=0, second x=0.2, last x=7.59, y=0, radius=0.1]
\anomalies[first x=5.8, second x=6.0, last x=6.01, y=0, radius=0.1]
\anomalies[first x=6.8, second x=7.0, last x=7.01, y=0, radius=0.1]
\end{tikzpicture}
&0.9&\textbf{2}\\
\bottomrule
\end{tabular}
    \caption{We test the affiliation and temporal distance metrics on two predictions of the same label time series. The best score for each metric is shown in bold. The labels include two events, and each prediction is a bit early on one of them. The affiliation metric split the time series into periods with one event each, and calculate the relative distance of the closest predicted event. In this example, the first anomaly in prediction 1 is seen as closer to a true anomaly than the second anomaly in prediction 2. \tempdist, on the other hand, uses absolute distance, and prefers prediction 2.
    }
    \label{tab:af_problem}
\end{figure*}
Temporal distance might seem very similar to the affiliation f-score. However, there are some important differences. Since \af~is calculated locally for every event, it is an event-based score, while \tempdist~is point-based, the effects of which will be clear from the experiments in Section~\ref{sec:experiments}. It may also lead to some odd situations when two or more anomalies are relatively close, as seen in Figure~\ref{tab:af_problem}. While \tempdist consider the absolute distances, and therefore consider the first event in prediction 1 to be further from the labels than the second event in prediction 2, \af consider relative distances within the local surroundings of each event, and therefore consider the distance in the last anomaly in prediction 2 as bigger than the first anomaly in prediction 1.

\subsection{Non-binary evaluation metrics}
\label{subsec:Nonbinary}
The \textit{non-binary evaluation metrics} are those evaluation the anomaly score, as opposed to a binary prediction obtained by using a threshold on the anomaly score. For these metric, the thresholding step is part of the evaluation.

A taxonomy of non-binary evaluation metrics is proposed in Figure~\ref{fig:taxonomy_diagram}. The primary difference between these metrics lies in the way they handle the threshold. Some metrics, such as \patk[K]~and \textit{binary metrics with optimal threshold}, choose a single threshold, resulting in a single binary prediction. These metrics are still considered non-binary as the threshold selection is part of the metric. The other non-binary metrics evaluate all possible thresholds (\textit{metrics based on all thresholds}) and combine them into a single number score. This is done either by calculating the area under a curve (\textit{AUC metrics}) or the volume under a surface (\textit{VUS metrics}). The choice of non-binary metric will depend on the specific requirements and goals of the evaluation, and the suitability of each metric for the task at hand.

\begin{figure*}[ht!]
    \centering
\begin{tikzpicture}[align=center]

\draw[rounded corners, black] (4,2.9) rectangle (7,6.5);
    \node [draw, shape=rectangle, fill=red!40] at (5.5,5.5) {Point-based\\metric(s)};
    \node [draw, shape=rectangle, diagonal fill={red!40}{blue!40}] at (5.5,4) {Point- and/or\\event-based\\metric(s)};

\node (a1) [draw, shape=rectangle] at (-0.5,6) {Non-binary\\evaluation metrics};
\node (b1) [draw, shape=rectangle] at (-3, 4) {Metrics based on\\a single threshold};
\node (b2) [draw, shape=rectangle] at (2, 4) {Metrics based on\\all thresholds};

\node (c1) [draw, shape=rectangle, fill=red!40!white] at (-4, 2) {\patk};
\node (c2) [draw, shape=rectangle, diagonal fill={red!40}{blue!40}] at (-1,2) {Binary metrics\\with optimal threshold\\(See Figure~\ref{fig:binary})};
\node (c3) [draw, shape=rectangle] at (2,2) {AUC\\metrics};
\node (c4) [draw, shape=rectangle] at (4,2) {VUS\\metrics};
\node (d1) [draw, shape=rectangle, fill=red!40!white] at (0,0) {\aucroc};
\node (d2) [draw, shape=rectangle, fill=red!40!white] at (2,0) {\aucpr};
\node (d3) [draw, shape=rectangle, fill=red!40!white] at (4,0) {\vusroc{l}};
\node (d4) [draw, shape=rectangle, fill=red!40!white] at (6,0) {\vuspr{l}};

\draw[->, thick] (a1) -- (b1);
\draw[->, thick] (a1) -- (b2);

\draw[->, thick] (b1) -- (c1);
\draw[->, thick] (b1) -- (c2);
\draw[->, thick] (b2) -- (c3);
\draw[->, thick] (b2) -- (c4);

\draw[->, thick] (c3) -- (d1);
\draw[->, thick] (c3) -- (d2);
\draw[->, thick] (c4) -- (d3);
\draw[->, thick] (c4) -- (d4);

\end{tikzpicture}
    \caption{A taxonomy of non-binary evaluation metrics. Although the input is different from the binary metrics, they are quite similar, and indeed any binary metric can be made non-binary by using the optimal threshold strategy (See section \ref{subsec:metrics:optimal_threshold}).
    }
    \label{fig:taxonomy_diagram}
\end{figure*}

\subsubsection{Precision at K (\patk[K])}
\label{subsec:metrics:patk}
The point-wise \patk[K]~metric defined in section \ref{subsec:traditional_metrics} is occasionally used for TSAD evaluation \cite{Paparrizos2022VolumeUT, Paparrizos2022TSBUADAE}. Other definitions of precision than point-wise could in principle be used, e.g. the works of \cite{Deng2022GraphCA, Zhang2019ADA} uses an event-based variant of recall at K for spatiotemporal anomaly detection, although for precision it would require defining how the number $K$ of anomalies included in the prediction is counted.

A variant of \patk[1]~is the UCR score used by \cite{Rewicki2022IsIW}, defined by 
\cite{ucrArchivePP}. The duration of the GT anomaly is increased in both ends to include some time tolerance, before \patk[1]~is calculated.

\subsubsection{Binary metrics with optimal threshold}
\label{subsec:metrics:optimal_threshold}
Binary metrics are typically used with the threshold that yields the best score \cite{Liu2022TimeSA,Huang2022ASV,Campos2021UnsupervisedTS, Deng2021GraphNN, He2019TemporalCN, Lavin2015EvaluatingRA}. This can be achieved with any binary evaluation metric. The use of a metric combined with this thresholding strategy requires the input of an anomaly score, resulting in non-binary evaluation. The optimal threshold is determined by using labels, and can only be determined during the evaluation phase, thus providing an upper limit to the score that can be achieved using the binary metric. The relevance of this upper limit depends on the situation and the chosen binary metric\footnote{E.g. optimal threshold \segf~score is always 1, independent of the anomaly score.}.
For the sake of brevity, we will only consider the point-wise f-score with the optimal threshold strategy (\bestpwf) in the remainder of this work.

\subsubsection{Area under the curve (\aucroc, \aucpr)}
\label{subsec:auc}
The receiver operator characteristic (ROC) is an evaluation metric commonly used for TSAD, as well as in binary classification in general.
For each choice of threshold, the prediction has a specific value of recall and false positive rate.
Plotting these against each other result in the ROC-curve. This is often inspected directly, as it visualizes the trade-off between recall and false positives, e.g. how large false positive rate must be allowed for certain levels of recall. 
In order to get a single scalar evaluation metric from this curve, it is common to integrate the area under the curve (AUC), to get the \aucroc.
This value summarizes the detection performance across all thresholds, and is widely used in TSAD \cite{Feng2022UnsupervisedMA,Dai2022GraphAugmentedNF, Schmidl2022AnomalyDI,Campos2021UnsupervisedTS,Bhatia2021MStreamFA,Li2021DCTGANDC,Huang2020CrowdQuakeAN,Goodge2020RobustnessOA,Braei2020AnomalyDI,Zhang2020VELCAN,Ergen2020UnsupervisedAD,Wang2019StudyOW,Kieu2019OutlierDF,Zhou2019BeatGANAR,Pang2019DeepAD, Park2017AMA}.
An alternative method to comparing recall and false positive rate is to apply an area under curve approach to precision and recall, resulting in the calculation of the area under the precision-recall curve (\aucpr), also known as average precision. This approach too is commonly utilized in TSAD \cite{Li2022LearningRD, Campos2021UnsupervisedTS,Li2021AnomalyDO,He2020ASD,Huang2020CrowdQuakeAN,Chen2020UnsupervisedAD,Kieu2019OutlierDF,Zhou2019BeatGANAR,Pang2019DeepAD}.
In our experiments, we only consider the point-wise precision and recall for the PR curve, as is by far most used, although any other pairs can be used, like the point-adjusted \aucpr~by \cite{Dai2021SDFVAESA} or the range based \aucpr~by \cite{Schmidl2022AnomalyDI}. Variations of the ROC curves can be used as well, but the false positive rate is not defined for the event-based metrics.

The use of \aucroc~has been criticized for its integration over all thresholds, which can result in a large portion of the score coming from thresholds that may not be relevant for a specific use case \cite{Baker2001APD, Lobo2008AUCAM, Berrar2012CaveatsAP}. A possible solution can be to only consider parts of the curve, as suggested by \cite{Baker2001APD}, although it can be hard to determine how much of it to use.
Another possibility is to use \aucpr~instead. While \aucpr~also integrates over all thresholds, it has been argued that it is more informative than the ROC for imbalanced datasets \cite{Davis2006TheRB,Saito2015ThePP}, which by definition is the case for anomaly detection\footnote{As pointed out by \cite{Wu2022CurrentTS}, not all commonly used datasets for TSAD are particularly imbalanced. Finding the labels in these datasets cannot really be considered anomaly detection, but should rather be regarded as classification or segmentation.}. The reason is that precision and false positive rate respond differently to changes in false positives (FPs). In anomaly detection, the number of true negatives will typically be very large compared to FPs, making the false positive rate low for all relevant choices of threshold. As a result, only a small part of the ROC curve is relevant in such cases.

\begin{figure}[tb!]\centering
        \begin{subfigure}{0.6\linewidth}
            \centering
            \includegraphics[width=0.9\linewidth]{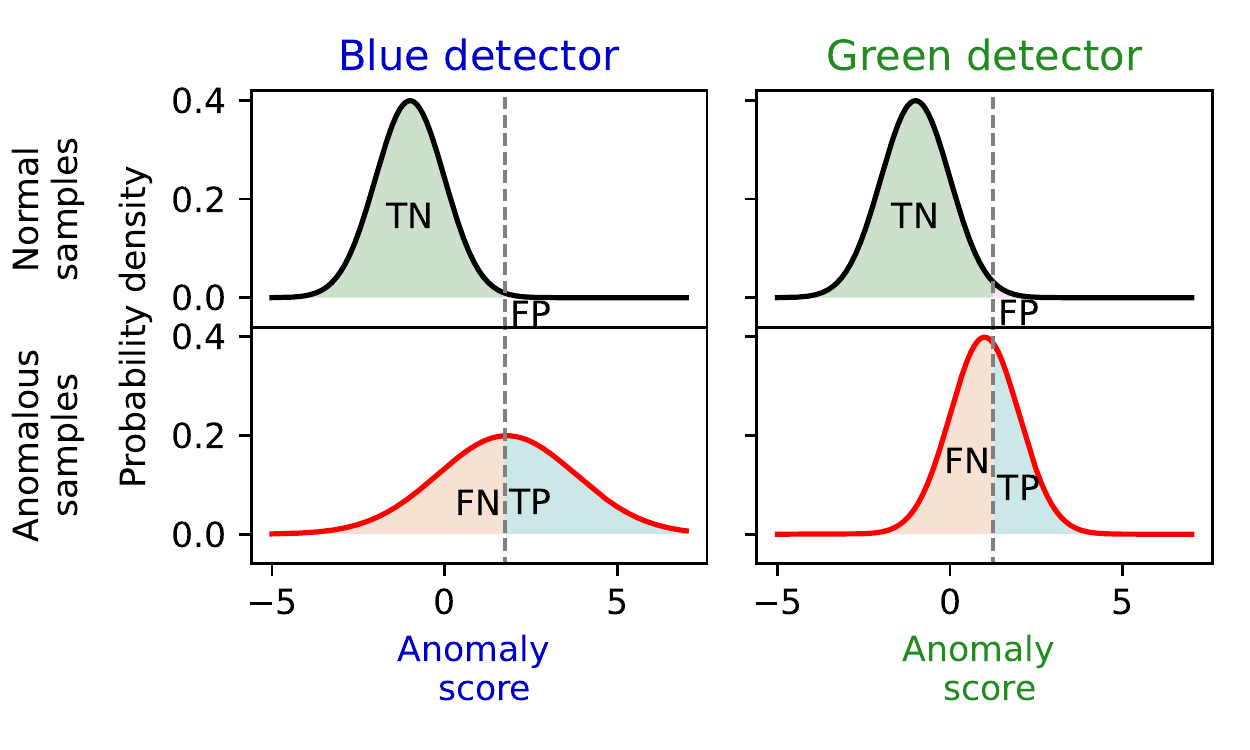}
            \caption{Thresholded optimally for \pwf[1]}\label{fig:auc_distr_b1}
        \end{subfigure}\\
        \begin{subfigure}{0.6\linewidth}
            \centering
            \includegraphics[width=0.9\linewidth]{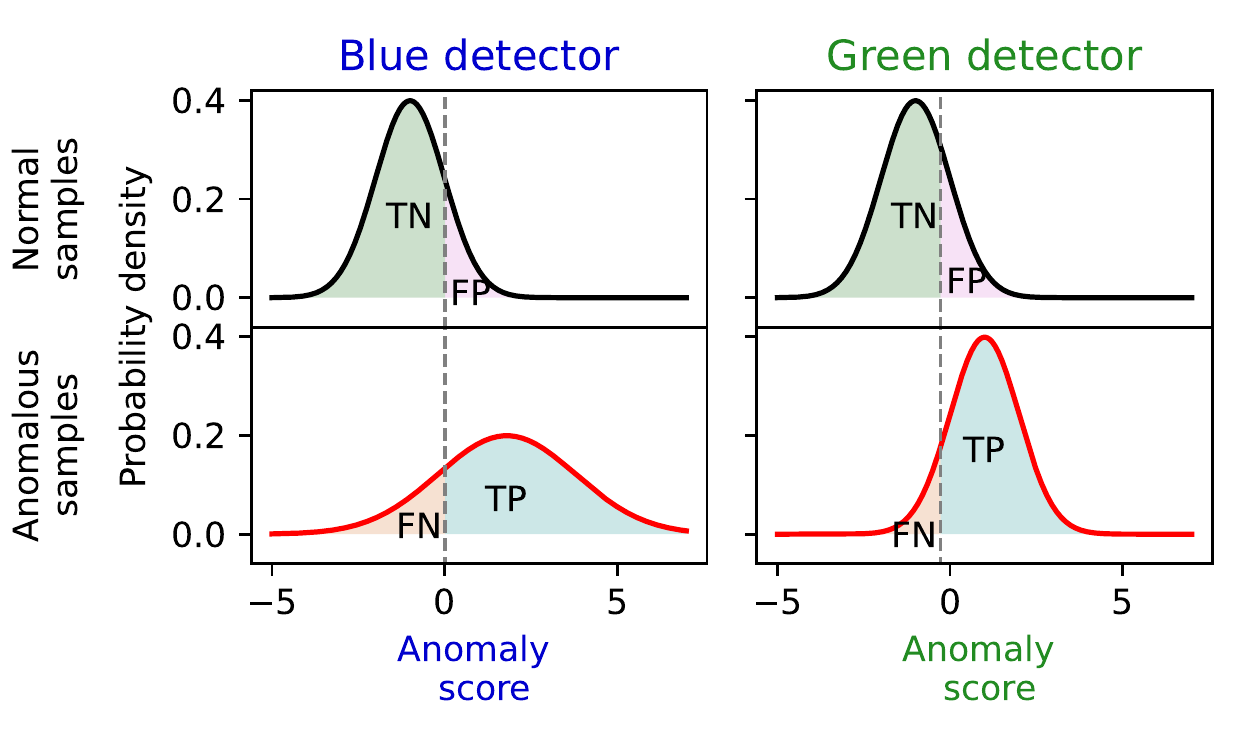}
            \caption{Thresholded optimally for \pwf[8]}\label{fig:auc_distr_b8}
        \end{subfigure}
        \caption{Probability density functions for the anomaly scores of the positive (red) and negative (black) samples, for the blue and green detectors. Anomalous samples generally give higher anomaly scores, although many normal and anomalous points have similar scores, making it hard to set a threshold.
        \ref{fig:auc_distr_b1} and \ref{fig:auc_distr_b8} show the counting metrics (as the shaded areas) for thresholds optimal for \pwf[\beta]~for $\beta = 1$ and $\beta=8$ respectively. Keep in mind that only 2\% of samples are anomalies, i.e. $\text{TN}+\text{FP} = 49(\text{FN}+\text{TP})$, which is not shown in the figures. FP and FN are of comparable sizes in \ref{fig:auc_distr_b1}.}
        \label{fig:auc_distr}
    \end{figure}

\begin{figure}[tb!]\centering
        \begin{subfigure}{0.5\linewidth}
            \centering
            \includegraphics[width=\textwidth]{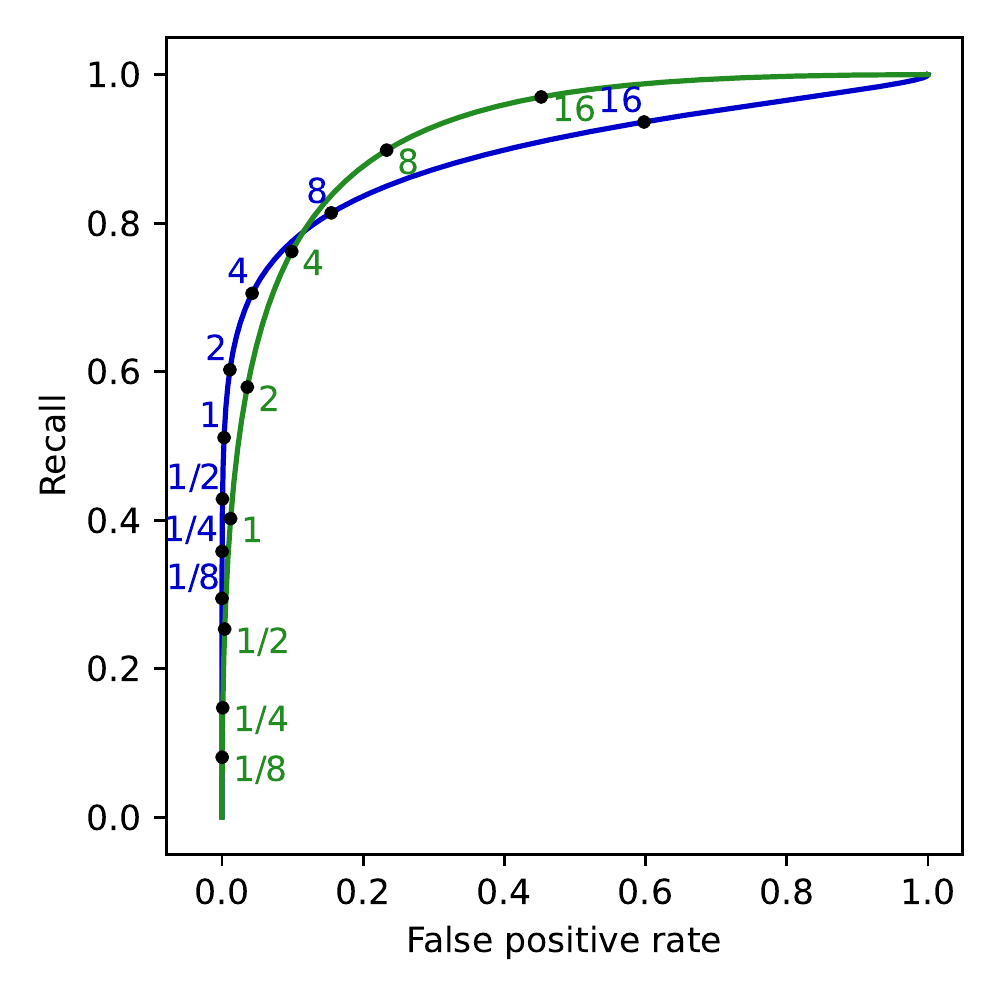}
            \caption{ROC curves}
            \label{fig:auc_roc_f}
        \end{subfigure}
        \begin{subfigure}{0.5\linewidth}
            \centering
            \includegraphics[width=\textwidth]{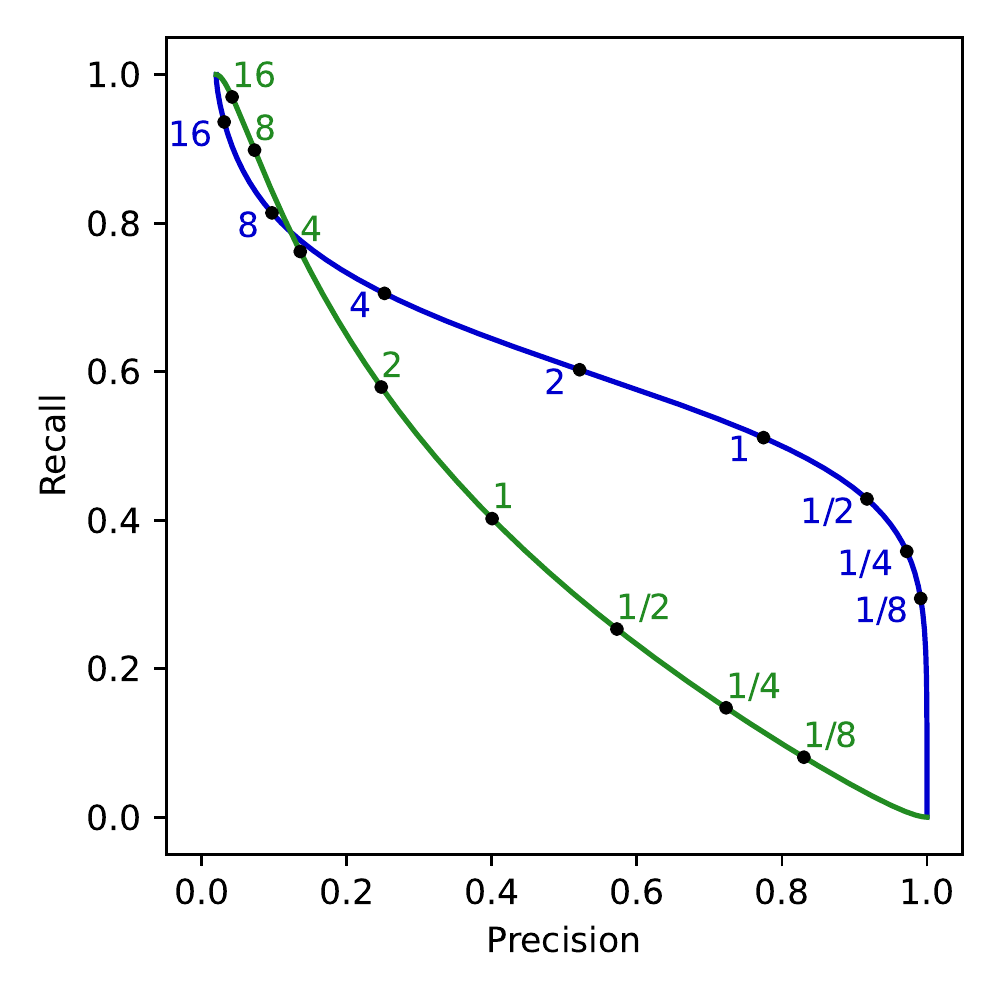}
            \caption{PR curves}
            \label{fig:auc_pr_f}
        \end{subfigure}
        \caption{The ROC and PR curves for the blue and green detectors. The marked dots are the points corresponding to the optimal thresholds for \pwf~with $\beta \in {16,8,4,2,1,1/2, 1/4, 1/8}$.
        We observe that the blue detector is best for higher thresholds, which are optimal for lower values of $\beta$, and vice versa. The green detector has the higher \aucroc, while the blue on have the higher \aucpr.}
        \label{fig:auc_f}
    \end{figure}

We visualize this with an example.
Assume a very large dataset has 2\% anomalies, and that two detectors, named blue and green, produce anomaly scores from the normal distributions visualized in Figure \ref{fig:auc_distr}. That is, the detectors produces anomaly scores from the black distributions in Figure~\ref{fig:auc_distr} for normal points, and from the red one for anomalous points. Note that since \aucroc~and \aucpr~are independent of the time dimension, time is not included in this example.
This results in the ROC-curves in Figure~\ref{fig:auc_roc_f} and PR-curves in Figure~\ref{fig:auc_pr_f}.

From the roc curves in Figure~\ref{fig:auc_roc_f} we see that the green detector outperform the blue detector for most values of the false positive rate. This would result in \aucroc~preferring the green detector. By inspecting the graph, we see that for smaller PFR, the blue detector is better. 
Inspecting the PR curves in Figure~\ref{fig:auc_pr_f}, we see that the blue detector by far would have the best \aucpr, but for low precision the green detector is better.
While the figures really contain the same information \cite{Davis2006TheRB}, it is clear that the difference in x-axis is crucial, not only for AUC-values but for inspection of the curves as well.

Figures~\ref{fig:auc_distr_b1} and \ref{fig:auc_distr_b8} also show the thresholds yielding the optimal f-score at different values of $\beta$. The points on the curves of these values, and more, are shown in Figures \ref{fig:auc_roc_f} and \ref{fig:auc_pr_f}. We see that for $\beta \approx 1$, the recall value has very little impact on the \aucroc, compared to the \aucpr. Indeed, the relevant values of $\beta$ should be quite high for \aucroc~to be more informative than \aucpr. 
But from Figure \ref{fig:auc_distr}, the high $\beta$ might seem more relevant, due to the large increase in TP, and high values of $\beta$ make up a relatively small part of the pr curves in Figure~\ref{fig:auc_pr_f}. As always, what is most suitable comes down to the situation. Since the ROC curve uses the fraction of FP to all normal samples, instead of anomalous predictions, the difference between ROC and PR scales with the imbalance of the data - when the anomalies make up an even smaller fraction of the data, \aucroc~corresponds to even higher values of $\beta$.

\subsubsection{Volume under the surface (\vusroc{l}, \vuspr{l})}
\label{subsec:vus}
The concept of volume under the surface (VUS) was introduced by \cite{Paparrizos2022VolumeUT}, extending \aucroc~and \aucpr. The authors recognized the need for some tolerance for predicted anomalies close to actual anomalies. They addressed this issue by adjusting the labels, and instead of using binary labels of 0 or 1, they use labels with real values in the range $[0,1]$. The original labelled anomalies are still given a value of 1, and normal points that are a certain distance $l$ away from anomalies are given a value of $0$. Labels closer to the original labelled anomalies gradually decrease as the distance from the anomaly increases\footnote{A similar smoothing strategy is done by \cite{Dai2022GraphAugmentedNF} to account for noisy labels, before applying \aucroc.}. The authors refined the point-wise recall by multiplying it with the existence factor used in \rf{bias}{\alpha}. Using the new definitions of recall, precision, and false positive rate, they defined range versions of \aucroc~and \aucpr. However, since this approach depends heavily on the tolerance threshold, $l$, they also introduced the volume under surface metric. Inspired by the way that the AUC metrics integrate away the dependency on the threshold by considering the area under a curve generated from all values of the threshold, the VUS metrics integrates over $l$ to generate the volume under the surface generated by the ROC or PR curve along an axis of values of $l$. This way, the final value takes into account multiple tolerance levels. Nevertheless, the metric still depends on the maximum value for $l$.

\section{Case studies}
\label{sec:experiments}

In this section, we evaluate the presented evaluation metrics on $14$ different case studies, in order to illustrate the different properties of the metrics. It is important to note that the desirability of these properties is highly dependent on the specific domain and use case. Thus, there is no universal "correct" answer for which metrics are best, but for a specific use case there is often one that is most appropriate. By presenting examples and highlighting the properties of the metrics, we aim to provide a clearer understanding of how they can be used effectively in different situations.

To simplify reading the results, the names for each evaluation metric presented, is repeated in Figure~\ref{fig:names}.
\begin{figure*}[]
    \tabcolsep=0.15cm 
    \centering
    \begin{tabular}{cccccc}
        \toprule
        &Short name & Long name & {Point-based} & {Event-based} & Section  \\
        \midrule
        \multirow{14}{*}{\rotatebox{90}{Binary~~~~~~}}
        &\pwf &{Point-wise f-score}&\ding{52}& & \ref{subsec:metrics:pointwise}\\ 
        &\paf &{Point Adjusted f-score}&\ding{52}& &  \ref{subsec:metrics:pointadjusted}\\ 
        &\dtpaf &{Delay thresholded Point Adjusted f-score}&\ding{52}& & \ref{subsec:metrics:pointadjusted}\\ 
        &\pakf{K} &{Point adjusted metrics at K\%}&\ding{52}& & \ref{subsec:metrics:pointadjusted}\\ 
        &\lsf{n} &{Latency and sparsity-aware f-score}&\ding{52}&& \ref{subsec:metrics:pointadjusted}\\ 
        &\segf &{Segment-wise f-score}& &\ding{52}& \ref{subsec:metrics:redefined_counting}\\ 
        &\cf &{Composite f-score}&\ding{52}&\ding{52}& \ref{subsec:metrics:redefined_counting}\\ 
        &\ttolf{\tau} &{Time tolerant f-score}&\ding{52}& & \ref{subsec:metrics:redefined_counting}\\ 
        &\rf{bias}{\alpha}&{Range based f-score}& &\ding{52}& \ref{subsec:metrics:redefined_PR}\\ 
        &\taf{\alpha}{\delta}{\theta}&{Time series aware f-score}& &\ding{52}& \ref{subsec:metrics:redefined_PR} \\ 
        &\etaf{\theta_p}{\theta_r} &{Enhanced time series aware f-score}& &\ding{52}& \ref{subsec:metrics:redefined_PR}\\ 
        &\af &{Affiliation based f-score}& &\ding{52}& \ref{subsec:metrics:redefined_PR}\\ 
        &\nab &{NAB score}&\ding{52}&\ding{52}& \ref{subsec:metrics:other}\\ 
        &\tempdist &{Temporal distance}&\ding{52}& & \ref{subsec:metrics:other}\\ 
        \midrule
        \multirow{6}{*}{\rotatebox{90}{Non-binary~~~~}}
        &\patk[K] &{Precision at K}&\ding{52}& & \ref{subsec:metrics:patk}\\
        &\bestpwf&{Point-wise f-score with optimal threshold}&\ding{52}& & \ref{subsec:metrics:optimal_threshold} \\ 
        &\aucroc&{Area under the reciever operator characteristic curve}&\ding{52}& & \ref{subsec:auc}\\ 
        &\aucpr~&{Area under the precision-recall curve}&\ding{52}& & \ref{subsec:auc}\\ 
        &\vusroc{l} &{Volume under the reciever operator characteristic surface}&\ding{52}& & \ref{subsec:vus}\\
        &\vuspr{l} &{Volume under the precision-recall surface}&\ding{52}& & \ref{subsec:vus} \\
        \bottomrule
    \end{tabular}
    \caption{Overview of all the metrics considered in the case studies.}
    \label{fig:names}
\end{figure*}

Here we outline the decisions made regarding the implementation of the evaluation metrics, and parameter selection. A majority of the metrics have parameters that need to be specified. To maintain consistency in our experiments, we have chosen the same evaluation metric parameters for most of the case studies. However, in some cases, we adjust these parameters to highlight a specific effect.

The $\beta$ in the $f_\beta$ is 1 for all f-score based metrics. For \dtpaf{k}~we use a delay threshold of $k=2$ time points. For \pakf{K}~we require 20\% of the anomaly detected for adjustment. 
The downsampling factor of \lsf{n}~is set to 2, and the temporal tolerance of \ttolf{\tau}~to $\tau=2$ for most experiments, expect for the on in Figure~\ref{tab:discontinuity}, where we use $\tau=10$ to better visualize its effect.

For the range based f-score \rf{bias}{\alpha}, we use $cardinality=1$, and specify $\alpha$ and the positioning bias the metric name in the table for each experiment. See the work of \cite{Tatbul2018PrecisionAR} for the definition of these parameters and functions. We use the same configuration for precision and recall.

For \taf{\alpha}{\delta}{\theta}~we set $\alpha=\theta=0.5$ for all tests. We use $\delta=0$ in most cases since this is more in line with the tests. We use $\delta=10$ for the graph in Figure~\ref{tab:discontinuity} to show the effect of this delta.
For \etaf{\theta_p}{\theta_r}, we use $\theta_p = 0.5$, $\theta_r=0.1$, to show the effect of using different values of these parameter. This will effectively ignore any predicted event with less than 0.5 precision, i.e. if less than half of the predicted event overlaps with anomalies. On the other hand, less than 10\% of an anomalous event must be detected for it to be counted as undetected. Using $\theta_r=\theta_p=0.5$ would yield results similar to that of \taf[1]{0.5}{0}{0.5}~in most cases.

\nab~is implemented using the \textit{standard application profile} \cite{nab_whitepaper}. As \nab~is implemented for use with longer anomalies, it does not run in the cases where there are events of length 1 in the labels. We do not include \nab~in these cases.

\patk~is the precision of the $K$ highest anomaly scores. For \patk[K]~we set $K$ to the number of anomaly points in the labels.
Due to many equal anomaly scores in the test cases, a threshold including $K$ points will often include $L>K$ points. In these cases we report \patk[L]~instead.

For \vusroc{l}~and \vuspr{l}~we use a maximum tolerance of $l=4$.

While we have implemented the simple metrics ourselves, the more complicated ones were taken from open source implementations by the authors of the metrics. \aucroc~and \aucpr~are from sklearn \cite{Pedregosa2011ScikitlearnML}.
Our implementation of the metrics, along with the code for generating the tables and figures in this paper, are available on Github\footnote{\url{https://github.com/sondsorb/TSAD_eval}}.

\subsection{Binary cases}

In order to test the preferences of different metrics, we have made a series of simple experiments with one time series of labels, and two imperfect prediction time series that resemble the labels in different ways. We then test which of the two predictions each of the metrics prefer. For each test we refer to a figure showing the time series and scores, with the optimal one for each metric shown in bold.

\subsubsection{Partial detection vs covering}
In anomaly detection, it may be sufficient to detect only a portion of the anomalous event. However, the correct duration of the event is still useful. Figure~\ref{tab:detection_over_covering} illustrates the different ways in which these aspects are addressed by various metrics. The point-wise f-score considers each point equally, regardless of whether the event has already been partially detected. In contrast, some metrics give the highest score to methods that detect only one point, providing no incentive to detect the entire event.

\subsubsection{Effect of anomaly length}

Most point-based metrics value each time point equally, while most event-based metrics value each event equally. Other options are \etaf{\theta_p}{\theta_r}, which weight events by the square root of their length,
\cf~which counts points and events for precision and recall respectively,
and \nab, counting TP event-wise and FP point-wise. 
These differences may lead to some unwanted prioritizations.
Figure~\ref{tab:length_problem_1} shows a situation with two short anomalies and one longer. For point-based metrics, it is better to predict the long one than both of the short ones.
For datasets with high variance in anomaly length, or a combination of point anomalies and event anomalies, an event-based metric is often more appropriate. 
On the other hand, event-based evaluation metrics can be sensitive to sets of short anomalies close to each other, as seen in Figure~\ref{tab:length_problem_2}, where the event-based metrics prioritize the cluster of three events over the single long one.

\subsubsection{Preference for short predicted anomalies}
\afterpage{
\begin{landscape}
\begin{figure}
\showDetectionOverCovering
    \caption{\textbf{Partial detection vs covering:} Detecting all the anomalies can be more valuable than covering one of them. Some metrics reflect this, but not all. Some do not value covering at all, and give optimal score to the bottom prediction.}
    \label{tab:detection_over_covering}
    \centering
\showLengthProblemI
    \caption{\textbf{Effect of anomaly length:} Point-based metrics value anomalies by their length, thus giving higher score to the top prediction than the bottom prediction, even if the latter one discover more of the anomaly events.}
    \label{tab:length_problem_1}
    \centering
\showLengthProblemII
    \caption{\textbf{Effect of anomaly length:} 
    Depending on the labelling strategy, some datasets might have several non-contiguous anomalies in close proximity. The event-based metrics will value detecting the (whole) cluster of anomalies over detecting the later contiguous anomaly.}
    \label{tab:length_problem_2}
    \centering
\showShortPrecitions
    \caption{\textbf{Preference for short predicted anomalies:} Most metrics value the bottom prediction at least as good as the top prediction, as it has the same precision but detects more of the anomaly. \paf, \tempdist~and \nab, on the other hand, has a strong preference for short predicted events.}
    \label{tab:short_predictions}
    \centering
\showLabellingProblem
    \caption{\textbf{Value proximity:} 
    Possible predictions for the time series in Figure~\ref{fig:changed_mean}. For each predicion, we use the other prediction as the labels. This shows how much metrics are affected if the detector and labeller uses different labelling strategies. A metric not sensitive to this would give good scores in both these situations.
    Point-based metrics like \paf~and \pwf~are heavily affected, event-based metrics a bit less. The metrics rewarding proximity are clearly the most tolerant for differing labelling strategy.}
    \label{tab:labelling_problem}
\end{figure}
\end{landscape}
}

For \paf~and \nab, there is no gain in having more than one TP point within an anomaly, while every FP is punished point-wise. This leads to a considerable preference for short predicted anomalies, as they can give high reward with a comparatively low risk. As seen in Figure~\ref{tab:short_predictions}, if two detection methods find the same anomalous events, but one of them produce longer predicted anomalies, the score may be very different.
This may seem like the precision/recall tradeoff in disguise - these two prediction could come from the same anomaly scores, but using different thresholds. However, some methods indeed predict shorter anomaly events than other methods, independent of the threshold.

\subsubsection{Score as a function of position of the predicted event}

To visualize how the different metrics value predicted events at different positions relative to a labelled event, we made a scenario with a time series of length 100, with one anomalous event from step 40 to 60, and a prediction with one anomalous event of length 5, at variable positions. We calculate the score for each position of the predicted anomaly, and plot this in a graph, as visualized in Figure~\ref{fig:disc_graph_case}.
Figure~\ref{tab:discontinuity} visualizes the score for each metric as a function of the position of the predicted event.
We include \rf{bias}{\alpha} with two positioning bias functions two show the different effects they have on the score.
As we see, the sensitivity to the position of the prediction varies considerably. \segf[1]~only has two values in the score, and \paf[1]~and \etaf[1]{0.5}{0.1}~has almost the same shape, with only slightly reduced score at the edges. Many of the other metrics have more gradually changing scores. As abnormality in reality seldom is a binary concept, gradually changing scores should be more fair in most cases.

\paragraph{Value earliness}
We see that \dtpaf[1]{2}, \lsf[1]{2}, \nab~and \rf[1]{front}{0}~all value earliness, but in different ways, and to varying degree. \nab~only has a slight preference for early detection, while \lsf{2}~and \rf[1]{front}{0}~have about linearly decreasing scores. \dtpaf[1]{2}~changes very abruptly, and only values very early detections.

\paragraph{Value proximity}
In cases where ground truth labels are not precise, methods should be rewarded more/punished less for a false positive close to a true anomaly than farther from them.
Note that the value of earliness might interfere with this, so balancing these concepts can be difficult. We have not found any one metric considering both of these concepts.
\af~and \tempdist~stands out as the only ones valuing relative proximity over the whole time series.
Along with \ttolf{10}, these are the only ones valuing detecting anomalies \textit{before} the labelled anomaly, while also \taf[1]{0.5}{10}{0.5}~and (barely) \nab~value detection \textit{after} the labelled anomaly.

\begin{figure*}
    \centering
    \begin{tikzpicture}[scale=1, baseline=-\the\dimexpr\fontdimen22\textfont2\relax]
        \nomalies[first x = -0.2, second x=-0.1, last x=9.79, y=1.2, radius=0.05]  
        \anomalies[first x=3.8, second x = 3.9, last x =5.79, radius=0.05]
        \foreach \j in {0.8,3.1,5.4}{
            \node at (\j-0.2, 0.97) {...};
            \node at (\j+1, 0.97) {...};
            \draw [black, decorate,decoration={brace,amplitude=.2cm,mirror}] (\j-0.4,0.9) -- +(0:1.6) node [black, midway] {};
        }
        \nomalies[first x = 0.8, second x=0.9, last x=1.69, y=1, radius=0.05]  
        \nomalies[first x = 3.1, second x=3.2, last x=3.99, y=1, radius=0.05]  
        \nomalies[first x = 5.4, second x=5.5, last x=6.29, y=1, radius=0.05]  
        \anomalies[first x = 1, second x=1.1, last x=1.49, y=1, radius=0.05]  
        \anomalies[first x = 3.3, second x=3.4, last x=3.79, y=1, radius=0.05]  
        \anomalies[first x = 5.6, second x=5.7, last x=6.09, y=1, radius=0.05]  
        \draw[->] (1.2,0.7) -- (1.2,0.04);
        \draw[->] (3.5,0.7) -- (3.5,0.04);
        \draw[->] (5.8,0.7) -- (5.8,0.24);
        \draw[->] (2,1) -- (2.4,1);
        \draw[->] (4.3,1) -- (4.7,1);
        \draw[->] (6.6,1) -- (7,1);
        \draw[-, gray] (3.5,0) -- (3.5,0.1);
        \draw[-, gray] (4.0,0) -- (4.0,0.1);
        \draw[-, gray] (5.5,0) -- (5.5,0.1);
        \draw[-, gray] (6.0,0) -- (6.0,0.1);
        \draw[-, gray] (0,0) -- (9.5,0);
        \draw[->, gray] (9.75,1.2) -- (10.4,1.2) node[left, align=right, xshift=0.2cm, yshift=-.2cm] {Time};
        \draw[->, gray] (9.5,0) -- (10.4,0) node[left, align=right, xshift=0.2cm, yshift=-.0cm] {Temporal position of\\predicted anomaly};
        \draw[->, gray] (0,0) -- (0,0.5) node[left, pos=0.6, align=right] {Metric\\score};
        \foreach \i/\a in
        {0.0/ 0.0, 0.1/ 0.0, 0.2/ 0.0, 0.3/ 0.0, 0.4/ 0.0, 0.5/ 0.0, 0.6/ 0.0, 0.7/ 0.0, 0.8/ 0.0, 0.9/ 0.0, 1.0/ 0.0, 1.1/ 0.0, 1.2/ 0.0, 1.3/ 0.0, 1.4/ 0.0, 1.5/ 0.0, 1.6/ 0.0, 1.7/ 0.0, 1.8/ 0.0, 1.9/ 0.0, 2.0/ 0.0, 2.1/ 0.0, 2.2/ 0.0, 2.3/ 0.0, 2.4/ 0.0, 2.5/ 0.0, 2.6/ 0.0, 2.7/ 0.0, 2.8/ 0.0, 2.9/ 0.0, 3.0/ 0.0, 3.1/ 0.0, 3.2/ 0.0, 3.3/ 0.0, 3.4/ 0.0, 3.5/ 0.0, 3.6/ 0.1, 3.7/ 0.2, 3.8/ 0.3, 3.9/ 0.4, 4.0/ 0.5, 4.1/ 0.5, 4.2/ 0.5, 4.3/ 0.5, 4.4/ 0.5, 4.5/ 0.5, 4.6/ 0.5, 4.7/ 0.5, 4.8/ 0.5, 4.9/ 0.5, 5.0/ 0.5, 5.1/ 0.5, 5.2/ 0.5, 5.3/ 0.5, 5.4/ 0.5, 5.5/ 0.5, 5.6/ 0.4, 5.7/ 0.3, 5.8/ 0.2, 5.9/ 0.1, 6.0/ 0.0, 6.1/ 0.0, 6.2/ 0.0, 6.3/ 0.0, 6.4/ 0.0, 6.5/ 0.0, 6.6/ 0.0, 6.7/ 0.0, 6.8/ 0.0, 6.9/ 0.0, 7.0/ 0.0, 7.1/ 0.0, 7.2/ 0.0, 7.3/ 0.0, 7.4/ 0.0, 7.5/ 0.0, 7.6/ 0.0, 7.7/ 0.0, 7.8/ 0.0, 7.9/ 0.0, 8.0/ 0.0, 8.1/ 0.0, 8.2/ 0.0, 8.3/ 0.0, 8.4/ 0.0, 8.5/ 0.0, 8.6/ 0.0, 8.7/ 0.0, 8.8/ 0.0, 8.9/ 0.0, 9.0/ 0.0, 9.1/ 0.0, 9.2/ 0.0, 9.3/ 0.0, 9.4/ 0.0, 9.5/ 0.0}{
        \coordinate (now) at (\i,\a) {};
          \ifthenelse{\equal{\i}{0.0}}{}{
          \draw[-, teal, thick] (prev) -- (now);
          }
          \coordinate (prev) at (\i,\a) {};
        }
    \end{tikzpicture}
    \caption{\textbf{Score as a function of position of the predicted event:} Each point on the graph is the output of the considered evaluation metric (\pwf[1]~in this case) for one full prediction, where the position of the predicted anomaly changes. The vertical marks indicate the start and end of areas where some, but not all, predicted anomaly points are correct. In other words, before the first and after the last mark, the predictions have no TP points, while between the second and third mark, the predictions have no FP points.}
    \label{fig:disc_graph_case}

    \begin{tabular}[t]{cc}
    \toprule
    Metric&Score\\
    \midrule
    \pwf[1]&\begin{tikzpicture}[baseline=-\the\dimexpr\fontdimen22\textfont2\relax]
    \draw[-, gray] (3.5,-0.2) -- (3.5,-0.1);
    \draw[-, gray] (4.0,-0.2) -- (4.0,-0.1);
    \draw[-, gray] (5.5,-0.2) -- (5.5,-0.1);
    \draw[-, gray] (6.0,-0.2) -- (6.0,-0.1);
    \draw[-, gray] (0,-0.2) -- (9.5,-0.2);
    \foreach \i/\a in
    {0.0/ -0.2, 0.1/ -0.2, 0.2/ -0.2, 0.3/ -0.2, 0.4/ -0.2, 0.5/ -0.2, 0.6/ -0.2, 0.7/ -0.2, 0.8/ -0.2, 0.9/ -0.2, 1.0/ -0.2, 1.1/ -0.2, 1.2/ -0.2, 1.3/ -0.2, 1.4/ -0.2, 1.5/ -0.2, 1.6/ -0.2, 1.7/ -0.2, 1.8/ -0.2, 1.9/ -0.2, 2.0/ -0.2, 2.1/ -0.2, 2.2/ -0.2, 2.3/ -0.2, 2.4/ -0.2, 2.5/ -0.2, 2.6/ -0.2, 2.7/ -0.2, 2.8/ -0.2, 2.9/ -0.2, 3.0/ -0.2, 3.1/ -0.2, 3.2/ -0.2, 3.3/ -0.2, 3.4/ -0.2, 3.5/ -0.2, 3.6/ -0.1, 3.7/ 0.0, 3.8/ 0.1, 3.9/ 0.2, 4.0/ 0.3, 4.1/ 0.3, 4.2/ 0.3, 4.3/ 0.3, 4.4/ 0.3, 4.5/ 0.3, 4.6/ 0.3, 4.7/ 0.3, 4.8/ 0.3, 4.9/ 0.3, 5.0/ 0.3, 5.1/ 0.3, 5.2/ 0.3, 5.3/ 0.3, 5.4/ 0.3, 5.5/ 0.3, 5.6/ 0.2, 5.7/ 0.1, 5.8/ 0.0, 5.9/ -0.1, 6.0/ -0.2, 6.1/ -0.2, 6.2/ -0.2, 6.3/ -0.2, 6.4/ -0.2, 6.5/ -0.2, 6.6/ -0.2, 6.7/ -0.2, 6.8/ -0.2, 6.9/ -0.2, 7.0/ -0.2, 7.1/ -0.2, 7.2/ -0.2, 7.3/ -0.2, 7.4/ -0.2, 7.5/ -0.2, 7.6/ -0.2, 7.7/ -0.2, 7.8/ -0.2, 7.9/ -0.2, 8.0/ -0.2, 8.1/ -0.2, 8.2/ -0.2, 8.3/ -0.2, 8.4/ -0.2, 8.5/ -0.2, 8.6/ -0.2, 8.7/ -0.2, 8.8/ -0.2, 8.9/ -0.2, 9.0/ -0.2, 9.1/ -0.2, 9.2/ -0.2, 9.3/ -0.2, 9.4/ -0.2, 9.5/ -0.2}{
    \coordinate (now) at (\i,\a) {};
      \ifthenelse{\equal{\i}{0.0}}{}{
      \draw[-, teal, thick] (prev) -- (now);
      }
      \coordinate (prev) at (\i,\a) {};
    }
    \end{tikzpicture}
    \\
    \paf[1]&\begin{tikzpicture}[baseline=-\the\dimexpr\fontdimen22\textfont2\relax]
    \draw[-, gray] (3.5,-0.2) -- (3.5,-0.1);
    \draw[-, gray] (4.0,-0.2) -- (4.0,-0.1);
    \draw[-, gray] (5.5,-0.2) -- (5.5,-0.1);
    \draw[-, gray] (6.0,-0.2) -- (6.0,-0.1);
    \draw[-, gray] (0,-0.2) -- (9.5,-0.2);
    \foreach \i/\a in
    {0.0/ -0.2, 0.1/ -0.2, 0.2/ -0.2, 0.3/ -0.2, 0.4/ -0.2, 0.5/ -0.2, 0.6/ -0.2, 0.7/ -0.2, 0.8/ -0.2, 0.9/ -0.2, 1.0/ -0.2, 1.1/ -0.2, 1.2/ -0.2, 1.3/ -0.2, 1.4/ -0.2, 1.5/ -0.2, 1.6/ -0.2, 1.7/ -0.2, 1.8/ -0.2, 1.9/ -0.2, 2.0/ -0.2, 2.1/ -0.2, 2.2/ -0.2, 2.3/ -0.2, 2.4/ -0.2, 2.5/ -0.2, 2.6/ -0.2, 2.7/ -0.2, 2.8/ -0.2, 2.9/ -0.2, 3.0/ -0.2, 3.1/ -0.2, 3.2/ -0.2, 3.3/ -0.2, 3.4/ -0.2, 3.5/ -0.2, 3.6/ 0.255, 3.7/ 0.265, 3.8/ 0.276, 3.9/ 0.288, 4.0/ 0.3, 4.1/ 0.3, 4.2/ 0.3, 4.3/ 0.3, 4.4/ 0.3, 4.5/ 0.3, 4.6/ 0.3, 4.7/ 0.3, 4.8/ 0.3, 4.9/ 0.3, 5.0/ 0.3, 5.1/ 0.3, 5.2/ 0.3, 5.3/ 0.3, 5.4/ 0.3, 5.5/ 0.3, 5.6/ 0.288, 5.7/ 0.276, 5.8/ 0.265, 5.9/ 0.255, 6.0/ -0.2, 6.1/ -0.2, 6.2/ -0.2, 6.3/ -0.2, 6.4/ -0.2, 6.5/ -0.2, 6.6/ -0.2, 6.7/ -0.2, 6.8/ -0.2, 6.9/ -0.2, 7.0/ -0.2, 7.1/ -0.2, 7.2/ -0.2, 7.3/ -0.2, 7.4/ -0.2, 7.5/ -0.2, 7.6/ -0.2, 7.7/ -0.2, 7.8/ -0.2, 7.9/ -0.2, 8.0/ -0.2, 8.1/ -0.2, 8.2/ -0.2, 8.3/ -0.2, 8.4/ -0.2, 8.5/ -0.2, 8.6/ -0.2, 8.7/ -0.2, 8.8/ -0.2, 8.9/ -0.2, 9.0/ -0.2, 9.1/ -0.2, 9.2/ -0.2, 9.3/ -0.2, 9.4/ -0.2, 9.5/ -0.2}{
    \coordinate (now) at (\i,\a) {};
      \ifthenelse{\equal{\i}{0.0}}{}{
      \draw[-, teal, thick] (prev) -- (now);
      }
      \coordinate (prev) at (\i,\a) {};
    }
    \end{tikzpicture}
    \\
    \dtpaf[1]{2}&\begin{tikzpicture}[baseline=-\the\dimexpr\fontdimen22\textfont2\relax]
    \draw[-, gray] (3.5,-0.2) -- (3.5,-0.1);
    \draw[-, gray] (4.0,-0.2) -- (4.0,-0.1);
    \draw[-, gray] (5.5,-0.2) -- (5.5,-0.1);
    \draw[-, gray] (6.0,-0.2) -- (6.0,-0.1);
    \draw[-, gray] (0,-0.2) -- (9.5,-0.2);
    \foreach \i/\a in
    {0.0/ -0.2, 0.1/ -0.2, 0.2/ -0.2, 0.3/ -0.2, 0.4/ -0.2, 0.5/ -0.2, 0.6/ -0.2, 0.7/ -0.2, 0.8/ -0.2, 0.9/ -0.2, 1.0/ -0.2, 1.1/ -0.2, 1.2/ -0.2, 1.3/ -0.2, 1.4/ -0.2, 1.5/ -0.2, 1.6/ -0.2, 1.7/ -0.2, 1.8/ -0.2, 1.9/ -0.2, 2.0/ -0.2, 2.1/ -0.2, 2.2/ -0.2, 2.3/ -0.2, 2.4/ -0.2, 2.5/ -0.2, 2.6/ -0.2, 2.7/ -0.2, 2.8/ -0.2, 2.9/ -0.2, 3.0/ -0.2, 3.1/ -0.2, 3.2/ -0.2, 3.3/ -0.2, 3.4/ -0.2, 3.5/ -0.2, 3.6/ 0.255, 3.7/ 0.265, 3.8/ 0.276, 3.9/ 0.288, 4.0/ 0.3, 4.1/ 0.3, 4.2/ 0.3, 4.3/ -0.2, 4.4/ -0.2, 4.5/ -0.2, 4.6/ -0.2, 4.7/ -0.2, 4.8/ -0.2, 4.9/ -0.2, 5.0/ -0.2, 5.1/ -0.2, 5.2/ -0.2, 5.3/ -0.2, 5.4/ -0.2, 5.5/ -0.2, 5.6/ -0.2, 5.7/ -0.2, 5.8/ -0.2, 5.9/ -0.2, 6.0/ -0.2, 6.1/ -0.2, 6.2/ -0.2, 6.3/ -0.2, 6.4/ -0.2, 6.5/ -0.2, 6.6/ -0.2, 6.7/ -0.2, 6.8/ -0.2, 6.9/ -0.2, 7.0/ -0.2, 7.1/ -0.2, 7.2/ -0.2, 7.3/ -0.2, 7.4/ -0.2, 7.5/ -0.2, 7.6/ -0.2, 7.7/ -0.2, 7.8/ -0.2, 7.9/ -0.2, 8.0/ -0.2, 8.1/ -0.2, 8.2/ -0.2, 8.3/ -0.2, 8.4/ -0.2, 8.5/ -0.2, 8.6/ -0.2, 8.7/ -0.2, 8.8/ -0.2, 8.9/ -0.2, 9.0/ -0.2, 9.1/ -0.2, 9.2/ -0.2, 9.3/ -0.2, 9.4/ -0.2, 9.5/ -0.2}{
    \coordinate (now) at (\i,\a) {};
      \ifthenelse{\equal{\i}{0.0}}{}{
      \draw[-, teal, thick] (prev) -- (now);
      }
      \coordinate (prev) at (\i,\a) {};
    }
    \end{tikzpicture}
    \\
    \pakf[1]{20}&\begin{tikzpicture}[baseline=-\the\dimexpr\fontdimen22\textfont2\relax]
    \draw[-, gray] (3.5,-0.2) -- (3.5,-0.1);
    \draw[-, gray] (4.0,-0.2) -- (4.0,-0.1);
    \draw[-, gray] (5.5,-0.2) -- (5.5,-0.1);
    \draw[-, gray] (6.0,-0.2) -- (6.0,-0.1);
    \draw[-, gray] (0,-0.2) -- (9.5,-0.2);
    \foreach \i/\a in
    {0.0/ -0.2, 0.1/ -0.2, 0.2/ -0.2, 0.3/ -0.2, 0.4/ -0.2, 0.5/ -0.2, 0.6/ -0.2, 0.7/ -0.2, 0.8/ -0.2, 0.9/ -0.2, 1.0/ -0.2, 1.1/ -0.2, 1.2/ -0.2, 1.3/ -0.2, 1.4/ -0.2, 1.5/ -0.2, 1.6/ -0.2, 1.7/ -0.2, 1.8/ -0.2, 1.9/ -0.2, 2.0/ -0.2, 2.1/ -0.2, 2.2/ -0.2, 2.3/ -0.2, 2.4/ -0.2, 2.5/ -0.2, 2.6/ -0.2, 2.7/ -0.2, 2.8/ -0.2, 2.9/ -0.2, 3.0/ -0.2, 3.1/ -0.2, 3.2/ -0.2, 3.3/ -0.2, 3.4/ -0.2, 3.5/ -0.2, 3.6/ -0.16, 3.7/ -0.12, 3.8/ -0.08, 3.9/ 0.288, 4.0/ 0.3, 4.1/ 0.3, 4.2/ 0.3, 4.3/ 0.3, 4.4/ 0.3, 4.5/ 0.3, 4.6/ 0.3, 4.7/ 0.3, 4.8/ 0.3, 4.9/ 0.3, 5.0/ 0.3, 5.1/ 0.3, 5.2/ 0.3, 5.3/ 0.3, 5.4/ 0.3, 5.5/ 0.3, 5.6/ 0.288, 5.7/ -0.08, 5.8/ -0.12, 5.9/ -0.16, 6.0/ -0.2, 6.1/ -0.2, 6.2/ -0.2, 6.3/ -0.2, 6.4/ -0.2, 6.5/ -0.2, 6.6/ -0.2, 6.7/ -0.2, 6.8/ -0.2, 6.9/ -0.2, 7.0/ -0.2, 7.1/ -0.2, 7.2/ -0.2, 7.3/ -0.2, 7.4/ -0.2, 7.5/ -0.2, 7.6/ -0.2, 7.7/ -0.2, 7.8/ -0.2, 7.9/ -0.2, 8.0/ -0.2, 8.1/ -0.2, 8.2/ -0.2, 8.3/ -0.2, 8.4/ -0.2, 8.5/ -0.2, 8.6/ -0.2, 8.7/ -0.2, 8.8/ -0.2, 8.9/ -0.2, 9.0/ -0.2, 9.1/ -0.2, 9.2/ -0.2, 9.3/ -0.2, 9.4/ -0.2, 9.5/ -0.2}{
    \coordinate (now) at (\i,\a) {};
      \ifthenelse{\equal{\i}{0.0}}{}{
      \draw[-, teal, thick] (prev) -- (now);
      }
      \coordinate (prev) at (\i,\a) {};
    }
    \end{tikzpicture}
    \\
    \lsf[1]{2}&\begin{tikzpicture}[baseline=-\the\dimexpr\fontdimen22\textfont2\relax]
    \draw[-, gray] (3.5,-0.2) -- (3.5,-0.1);
    \draw[-, gray] (4.0,-0.2) -- (4.0,-0.1);
    \draw[-, gray] (5.5,-0.2) -- (5.5,-0.1);
    \draw[-, gray] (6.0,-0.2) -- (6.0,-0.1);
    \draw[-, gray] (0,-0.2) -- (9.5,-0.2);
    \foreach \i/\a in
    {0.0/ -0.2, 0.1/ -0.2, 0.2/ -0.2, 0.3/ -0.2, 0.4/ -0.2, 0.5/ -0.2, 0.6/ -0.2, 0.7/ -0.2, 0.8/ -0.2, 0.9/ -0.2, 1.0/ -0.2, 1.1/ -0.2, 1.2/ -0.2, 1.3/ -0.2, 1.4/ -0.2, 1.5/ -0.2, 1.6/ -0.2, 1.7/ -0.2, 1.8/ -0.2, 1.9/ -0.2, 2.0/ -0.2, 2.1/ -0.2, 2.2/ -0.2, 2.3/ -0.2, 2.4/ -0.2, 2.5/ -0.2, 2.6/ -0.2, 2.7/ -0.2, 2.8/ -0.2, 2.9/ -0.2, 3.0/ -0.2, 3.1/ -0.2, 3.2/ -0.2, 3.3/ -0.2, 3.4/ -0.2, 3.5/ -0.2, 3.6/ 0.255, 3.7/ 0.255, 3.8/ 0.276, 3.9/ 0.276, 4.0/ 0.3, 4.1/ 0.3, 4.2/ 0.274, 4.3/ 0.274, 4.4/ 0.244, 4.5/ 0.244, 4.6/ 0.212, 4.7/ 0.212, 4.8/ 0.175, 4.9/ 0.175, 5.0/ 0.133, 5.1/ 0.133, 5.2/ 0.086, 5.3/ 0.086, 5.4/ 0.031, 5.5/ 0.031, 5.6/ -0.046, 5.7/ -0.046, 5.8/ -0.123, 5.9/ -0.123, 6.0/ -0.2, 6.1/ -0.2, 6.2/ -0.2, 6.3/ -0.2, 6.4/ -0.2, 6.5/ -0.2, 6.6/ -0.2, 6.7/ -0.2, 6.8/ -0.2, 6.9/ -0.2, 7.0/ -0.2, 7.1/ -0.2, 7.2/ -0.2, 7.3/ -0.2, 7.4/ -0.2, 7.5/ -0.2, 7.6/ -0.2, 7.7/ -0.2, 7.8/ -0.2, 7.9/ -0.2, 8.0/ -0.2, 8.1/ -0.2, 8.2/ -0.2, 8.3/ -0.2, 8.4/ -0.2, 8.5/ -0.2, 8.6/ -0.2, 8.7/ -0.2, 8.8/ -0.2, 8.9/ -0.2, 9.0/ -0.2, 9.1/ -0.2, 9.2/ -0.2, 9.3/ -0.2, 9.4/ -0.2, 9.5/ -0.2}{
    \coordinate (now) at (\i,\a) {};
      \ifthenelse{\equal{\i}{0.0}}{}{
      \draw[-, teal, thick] (prev) -- (now);
      }
      \coordinate (prev) at (\i,\a) {};
    }
    \end{tikzpicture}
    \\
    \segf[1]&\begin{tikzpicture}[baseline=-\the\dimexpr\fontdimen22\textfont2\relax]
    \draw[-, gray] (3.5,-0.2) -- (3.5,-0.1);
    \draw[-, gray] (4.0,-0.2) -- (4.0,-0.1);
    \draw[-, gray] (5.5,-0.2) -- (5.5,-0.1);
    \draw[-, gray] (6.0,-0.2) -- (6.0,-0.1);
    \draw[-, gray] (0,-0.2) -- (9.5,-0.2);
    \foreach \i/\a in
    {0.0/ -0.2, 0.1/ -0.2, 0.2/ -0.2, 0.3/ -0.2, 0.4/ -0.2, 0.5/ -0.2, 0.6/ -0.2, 0.7/ -0.2, 0.8/ -0.2, 0.9/ -0.2, 1.0/ -0.2, 1.1/ -0.2, 1.2/ -0.2, 1.3/ -0.2, 1.4/ -0.2, 1.5/ -0.2, 1.6/ -0.2, 1.7/ -0.2, 1.8/ -0.2, 1.9/ -0.2, 2.0/ -0.2, 2.1/ -0.2, 2.2/ -0.2, 2.3/ -0.2, 2.4/ -0.2, 2.5/ -0.2, 2.6/ -0.2, 2.7/ -0.2, 2.8/ -0.2, 2.9/ -0.2, 3.0/ -0.2, 3.1/ -0.2, 3.2/ -0.2, 3.3/ -0.2, 3.4/ -0.2, 3.5/ -0.2, 3.6/ 0.3, 3.7/ 0.3, 3.8/ 0.3, 3.9/ 0.3, 4.0/ 0.3, 4.1/ 0.3, 4.2/ 0.3, 4.3/ 0.3, 4.4/ 0.3, 4.5/ 0.3, 4.6/ 0.3, 4.7/ 0.3, 4.8/ 0.3, 4.9/ 0.3, 5.0/ 0.3, 5.1/ 0.3, 5.2/ 0.3, 5.3/ 0.3, 5.4/ 0.3, 5.5/ 0.3, 5.6/ 0.3, 5.7/ 0.3, 5.8/ 0.3, 5.9/ 0.3, 6.0/ -0.2, 6.1/ -0.2, 6.2/ -0.2, 6.3/ -0.2, 6.4/ -0.2, 6.5/ -0.2, 6.6/ -0.2, 6.7/ -0.2, 6.8/ -0.2, 6.9/ -0.2, 7.0/ -0.2, 7.1/ -0.2, 7.2/ -0.2, 7.3/ -0.2, 7.4/ -0.2, 7.5/ -0.2, 7.6/ -0.2, 7.7/ -0.2, 7.8/ -0.2, 7.9/ -0.2, 8.0/ -0.2, 8.1/ -0.2, 8.2/ -0.2, 8.3/ -0.2, 8.4/ -0.2, 8.5/ -0.2, 8.6/ -0.2, 8.7/ -0.2, 8.8/ -0.2, 8.9/ -0.2, 9.0/ -0.2, 9.1/ -0.2, 9.2/ -0.2, 9.3/ -0.2, 9.4/ -0.2, 9.5/ -0.2}{
    \coordinate (now) at (\i,\a) {};
      \ifthenelse{\equal{\i}{0.0}}{}{
      \draw[-, teal, thick] (prev) -- (now);
      }
      \coordinate (prev) at (\i,\a) {};
    }
    \end{tikzpicture}
    \\
    \cf[1]&\begin{tikzpicture}[baseline=-\the\dimexpr\fontdimen22\textfont2\relax]
    \draw[-, gray] (3.5,-0.2) -- (3.5,-0.1);
    \draw[-, gray] (4.0,-0.2) -- (4.0,-0.1);
    \draw[-, gray] (5.5,-0.2) -- (5.5,-0.1);
    \draw[-, gray] (6.0,-0.2) -- (6.0,-0.1);
    \draw[-, gray] (0,-0.2) -- (9.5,-0.2);
    \foreach \i/\a in
    {0.0/ -0.2, 0.1/ -0.2, 0.2/ -0.2, 0.3/ -0.2, 0.4/ -0.2, 0.5/ -0.2, 0.6/ -0.2, 0.7/ -0.2, 0.8/ -0.2, 0.9/ -0.2, 1.0/ -0.2, 1.1/ -0.2, 1.2/ -0.2, 1.3/ -0.2, 1.4/ -0.2, 1.5/ -0.2, 1.6/ -0.2, 1.7/ -0.2, 1.8/ -0.2, 1.9/ -0.2, 2.0/ -0.2, 2.1/ -0.2, 2.2/ -0.2, 2.3/ -0.2, 2.4/ -0.2, 2.5/ -0.2, 2.6/ -0.2, 2.7/ -0.2, 2.8/ -0.2, 2.9/ -0.2, 3.0/ -0.2, 3.1/ -0.2, 3.2/ -0.2, 3.3/ -0.2, 3.4/ -0.2, 3.5/ -0.2, 3.6/ -0.033, 3.7/ 0.086, 3.8/ 0.175, 3.9/ 0.244, 4.0/ 0.3, 4.1/ 0.3, 4.2/ 0.3, 4.3/ 0.3, 4.4/ 0.3, 4.5/ 0.3, 4.6/ 0.3, 4.7/ 0.3, 4.8/ 0.3, 4.9/ 0.3, 5.0/ 0.3, 5.1/ 0.3, 5.2/ 0.3, 5.3/ 0.3, 5.4/ 0.3, 5.5/ 0.3, 5.6/ 0.244, 5.7/ 0.175, 5.8/ 0.086, 5.9/ -0.033, 6.0/ -0.2, 6.1/ -0.2, 6.2/ -0.2, 6.3/ -0.2, 6.4/ -0.2, 6.5/ -0.2, 6.6/ -0.2, 6.7/ -0.2, 6.8/ -0.2, 6.9/ -0.2, 7.0/ -0.2, 7.1/ -0.2, 7.2/ -0.2, 7.3/ -0.2, 7.4/ -0.2, 7.5/ -0.2, 7.6/ -0.2, 7.7/ -0.2, 7.8/ -0.2, 7.9/ -0.2, 8.0/ -0.2, 8.1/ -0.2, 8.2/ -0.2, 8.3/ -0.2, 8.4/ -0.2, 8.5/ -0.2, 8.6/ -0.2, 8.7/ -0.2, 8.8/ -0.2, 8.9/ -0.2, 9.0/ -0.2, 9.1/ -0.2, 9.2/ -0.2, 9.3/ -0.2, 9.4/ -0.2, 9.5/ -0.2}{
    \coordinate (now) at (\i,\a) {};
      \ifthenelse{\equal{\i}{0.0}}{}{
      \draw[-, teal, thick] (prev) -- (now);
      }
      \coordinate (prev) at (\i,\a) {};
    }
    \end{tikzpicture}
    \\
    \ttolf[1]{10}&\begin{tikzpicture}[baseline=-\the\dimexpr\fontdimen22\textfont2\relax]
    \draw[-, gray] (3.5,-0.2) -- (3.5,-0.1);
    \draw[-, gray] (4.0,-0.2) -- (4.0,-0.1);
    \draw[-, gray] (5.5,-0.2) -- (5.5,-0.1);
    \draw[-, gray] (6.0,-0.2) -- (6.0,-0.1);
    \draw[-, gray] (0,-0.2) -- (9.5,-0.2);
    \foreach \i/\a in
    {0.0/ -0.2, 0.1/ -0.2, 0.2/ -0.2, 0.3/ -0.2, 0.4/ -0.2, 0.5/ -0.2, 0.6/ -0.2, 0.7/ -0.2, 0.8/ -0.2, 0.9/ -0.2, 1.0/ -0.2, 1.1/ -0.2, 1.2/ -0.2, 1.3/ -0.2, 1.4/ -0.2, 1.5/ -0.2, 1.6/ -0.2, 1.7/ -0.2, 1.8/ -0.2, 1.9/ -0.2, 2.0/ -0.2, 2.1/ -0.2, 2.2/ -0.2, 2.3/ -0.2, 2.4/ -0.2, 2.5/ -0.2, 2.6/ -0.16, 2.7/ -0.12, 2.8/ -0.08, 2.9/ -0.04, 3.0/ 0.0, 3.1/ 0.031, 3.2/ 0.059, 3.3/ 0.086, 3.4/ 0.11, 3.5/ 0.133, 3.6/ 0.155, 3.7/ 0.175, 3.8/ 0.194, 3.9/ 0.212, 4.0/ 0.229, 4.1/ 0.244, 4.2/ 0.259, 4.3/ 0.274, 4.4/ 0.287, 4.5/ 0.3, 4.6/ 0.3, 4.7/ 0.3, 4.8/ 0.3, 4.9/ 0.3, 5.0/ 0.3, 5.1/ 0.287, 5.2/ 0.274, 5.3/ 0.259, 5.4/ 0.244, 5.5/ 0.229, 5.6/ 0.212, 5.7/ 0.194, 5.8/ 0.175, 5.9/ 0.155, 6.0/ 0.133, 6.1/ 0.11, 6.2/ 0.086, 6.3/ 0.059, 6.4/ 0.031, 6.5/ 0.0, 6.6/ -0.04, 6.7/ -0.08, 6.8/ -0.12, 6.9/ -0.16, 7.0/ -0.2, 7.1/ -0.2, 7.2/ -0.2, 7.3/ -0.2, 7.4/ -0.2, 7.5/ -0.2, 7.6/ -0.2, 7.7/ -0.2, 7.8/ -0.2, 7.9/ -0.2, 8.0/ -0.2, 8.1/ -0.2, 8.2/ -0.2, 8.3/ -0.2, 8.4/ -0.2, 8.5/ -0.2, 8.6/ -0.2, 8.7/ -0.2, 8.8/ -0.2, 8.9/ -0.2, 9.0/ -0.2, 9.1/ -0.2, 9.2/ -0.2, 9.3/ -0.2, 9.4/ -0.2, 9.5/ -0.2}{
    \coordinate (now) at (\i,\a) {};
      \ifthenelse{\equal{\i}{0.0}}{}{
      \draw[-, teal, thick] (prev) -- (now);
      }
      \coordinate (prev) at (\i,\a) {};
    }
    \end{tikzpicture}
    \\
    \taf[1]{0.5}{10}{0.5}&\begin{tikzpicture}[baseline=-\the\dimexpr\fontdimen22\textfont2\relax]
    \draw[-, gray] (3.5,-0.2) -- (3.5,-0.1);
    \draw[-, gray] (4.0,-0.2) -- (4.0,-0.1);
    \draw[-, gray] (5.5,-0.2) -- (5.5,-0.1);
    \draw[-, gray] (6.0,-0.2) -- (6.0,-0.1);
    \draw[-, gray] (0,-0.2) -- (9.5,-0.2);
    \foreach \i/\a in
    {0.0/ -0.2, 0.1/ -0.2, 0.2/ -0.2, 0.3/ -0.2, 0.4/ -0.2, 0.5/ -0.2, 0.6/ -0.2, 0.7/ -0.2, 0.8/ -0.2, 0.9/ -0.2, 1.0/ -0.2, 1.1/ -0.2, 1.2/ -0.2, 1.3/ -0.2, 1.4/ -0.2, 1.5/ -0.2, 1.6/ -0.2, 1.7/ -0.2, 1.8/ -0.2, 1.9/ -0.2, 2.0/ -0.2, 2.1/ -0.2, 2.2/ -0.2, 2.3/ -0.2, 2.4/ -0.2, 2.5/ -0.2, 2.6/ -0.2, 2.7/ -0.2, 2.8/ -0.2, 2.9/ -0.2, 3.0/ -0.2, 3.1/ -0.2, 3.2/ -0.2, 3.3/ -0.2, 3.4/ -0.2, 3.5/ -0.2, 3.6/ -0.11, 3.7/ -0.02, 3.8/ 0.109, 3.9/ 0.205, 4.0/ 0.3, 4.1/ 0.3, 4.2/ 0.3, 4.3/ 0.3, 4.4/ 0.3, 4.5/ 0.3, 4.6/ 0.3, 4.7/ 0.3, 4.8/ 0.3, 4.9/ 0.3, 5.0/ 0.3, 5.1/ 0.3, 5.2/ 0.3, 5.3/ 0.3, 5.4/ 0.3, 5.5/ 0.3, 5.6/ 0.3, 5.7/ 0.299, 5.8/ 0.296, 5.9/ 0.289, 6.0/ 0.267, 6.1/ 0.219, 6.2/ 0.146, 6.3/ 0.06, 6.4/ -0.055, 6.5/ -0.124, 6.6/ -0.168, 6.7/ -0.189, 6.8/ -0.197, 6.9/ -0.199, 7.0/ -0.2, 7.1/ -0.2, 7.2/ -0.2, 7.3/ -0.2, 7.4/ -0.2, 7.5/ -0.2, 7.6/ -0.2, 7.7/ -0.2, 7.8/ -0.2, 7.9/ -0.2, 8.0/ -0.2, 8.1/ -0.2, 8.2/ -0.2, 8.3/ -0.2, 8.4/ -0.2, 8.5/ -0.2, 8.6/ -0.2, 8.7/ -0.2, 8.8/ -0.2, 8.9/ -0.2, 9.0/ -0.2, 9.1/ -0.2, 9.2/ -0.2, 9.3/ -0.2, 9.4/ -0.2, 9.5/ -0.2}{
    \coordinate (now) at (\i,\a) {};
      \ifthenelse{\equal{\i}{0.0}}{}{
      \draw[-, teal, thick] (prev) -- (now);
      }
      \coordinate (prev) at (\i,\a) {};
    }
    \end{tikzpicture}
    \\
    \etaf[1]{0.5}{0.1}&\begin{tikzpicture}[baseline=-\the\dimexpr\fontdimen22\textfont2\relax]
    \draw[-, gray] (3.5,-0.2) -- (3.5,-0.1);
    \draw[-, gray] (4.0,-0.2) -- (4.0,-0.1);
    \draw[-, gray] (5.5,-0.2) -- (5.5,-0.1);
    \draw[-, gray] (6.0,-0.2) -- (6.0,-0.1);
    \draw[-, gray] (0,-0.2) -- (9.5,-0.2);
    \foreach \i/\a in
    {0.0/ -0.2, 0.1/ -0.2, 0.2/ -0.2, 0.3/ -0.2, 0.4/ -0.2, 0.5/ -0.2, 0.6/ -0.2, 0.7/ -0.2, 0.8/ -0.2, 0.9/ -0.2, 1.0/ -0.2, 1.1/ -0.2, 1.2/ -0.2, 1.3/ -0.2, 1.4/ -0.2, 1.5/ -0.2, 1.6/ -0.2, 1.7/ -0.2, 1.8/ -0.2, 1.9/ -0.2, 2.0/ -0.2, 2.1/ -0.2, 2.2/ -0.2, 2.3/ -0.2, 2.4/ -0.2, 2.5/ -0.2, 2.6/ -0.2, 2.7/ -0.2, 2.8/ -0.2, 2.9/ -0.2, 3.0/ -0.2, 3.1/ -0.2, 3.2/ -0.2, 3.3/ -0.2, 3.4/ -0.2, 3.5/ -0.2, 3.6/ -0.2, 3.7/ -0.2, 3.8/ 0.235, 3.9/ 0.268, 4.0/ 0.3, 4.1/ 0.3, 4.2/ 0.3, 4.3/ 0.3, 4.4/ 0.3, 4.5/ 0.3, 4.6/ 0.3, 4.7/ 0.3, 4.8/ 0.3, 4.9/ 0.3, 5.0/ 0.3, 5.1/ 0.3, 5.2/ 0.3, 5.3/ 0.3, 5.4/ 0.3, 5.5/ 0.3, 5.6/ 0.268, 5.7/ 0.235, 5.8/ -0.2, 5.9/ -0.2, 6.0/ -0.2, 6.1/ -0.2, 6.2/ -0.2, 6.3/ -0.2, 6.4/ -0.2, 6.5/ -0.2, 6.6/ -0.2, 6.7/ -0.2, 6.8/ -0.2, 6.9/ -0.2, 7.0/ -0.2, 7.1/ -0.2, 7.2/ -0.2, 7.3/ -0.2, 7.4/ -0.2, 7.5/ -0.2, 7.6/ -0.2, 7.7/ -0.2, 7.8/ -0.2, 7.9/ -0.2, 8.0/ -0.2, 8.1/ -0.2, 8.2/ -0.2, 8.3/ -0.2, 8.4/ -0.2, 8.5/ -0.2, 8.6/ -0.2, 8.7/ -0.2, 8.8/ -0.2, 8.9/ -0.2, 9.0/ -0.2, 9.1/ -0.2, 9.2/ -0.2, 9.3/ -0.2, 9.4/ -0.2, 9.5/ -0.2}{
    \coordinate (now) at (\i,\a) {};
      \ifthenelse{\equal{\i}{0.0}}{}{
      \draw[-, teal, thick] (prev) -- (now);
      }
      \coordinate (prev) at (\i,\a) {};
    }
    \end{tikzpicture}
    \\
    \af[1]&\begin{tikzpicture}[baseline=-\the\dimexpr\fontdimen22\textfont2\relax]
    \draw[-, gray] (3.5,-0.2) -- (3.5,-0.1);
    \draw[-, gray] (4.0,-0.2) -- (4.0,-0.1);
    \draw[-, gray] (5.5,-0.2) -- (5.5,-0.1);
    \draw[-, gray] (6.0,-0.2) -- (6.0,-0.1);
    \draw[-, gray] (0,-0.2) -- (9.5,-0.2);
    \foreach \i/\a in
    {0.0/ -0.2, 0.1/ -0.188, 0.2/ -0.176, 0.3/ -0.164, 0.4/ -0.153, 0.5/ -0.142, 0.6/ -0.131, 0.7/ -0.12, 0.8/ -0.109, 0.9/ -0.099, 1.0/ -0.088, 1.1/ -0.077, 1.2/ -0.066, 1.3/ -0.055, 1.4/ -0.044, 1.5/ -0.032, 1.6/ -0.021, 1.7/ -0.01, 1.8/ 0.001, 1.9/ 0.012, 2.0/ 0.023, 2.1/ 0.035, 2.2/ 0.046, 2.3/ 0.057, 2.4/ 0.068, 2.5/ 0.079, 2.6/ 0.09, 2.7/ 0.101, 2.8/ 0.113, 2.9/ 0.124, 3.0/ 0.135, 3.1/ 0.146, 3.2/ 0.157, 3.3/ 0.168, 3.4/ 0.179, 3.5/ 0.191, 3.6/ 0.213, 3.7/ 0.233, 3.8/ 0.251, 3.9/ 0.268, 4.0/ 0.283, 4.1/ 0.287, 4.2/ 0.291, 4.3/ 0.294, 4.4/ 0.296, 4.5/ 0.298, 4.6/ 0.299, 4.7/ 0.3, 4.8/ 0.3, 4.9/ 0.299, 5.0/ 0.298, 5.1/ 0.296, 5.2/ 0.294, 5.3/ 0.291, 5.4/ 0.287, 5.5/ 0.283, 5.6/ 0.268, 5.7/ 0.251, 5.8/ 0.233, 5.9/ 0.213, 6.0/ 0.191, 6.1/ 0.179, 6.2/ 0.168, 6.3/ 0.157, 6.4/ 0.146, 6.5/ 0.135, 6.6/ 0.124, 6.7/ 0.113, 6.8/ 0.101, 6.9/ 0.09, 7.0/ 0.079, 7.1/ 0.068, 7.2/ 0.057, 7.3/ 0.046, 7.4/ 0.035, 7.5/ 0.023, 7.6/ 0.012, 7.7/ 0.001, 7.8/ -0.01, 7.9/ -0.021, 8.0/ -0.032, 8.1/ -0.044, 8.2/ -0.055, 8.3/ -0.066, 8.4/ -0.077, 8.5/ -0.088, 8.6/ -0.099, 8.7/ -0.109, 8.8/ -0.12, 8.9/ -0.131, 9.0/ -0.142, 9.1/ -0.153, 9.2/ -0.164, 9.3/ -0.176, 9.4/ -0.188, 9.5/ -0.2}{
    \coordinate (now) at (\i,\a) {};
      \ifthenelse{\equal{\i}{0.0}}{}{
      \draw[-, teal, thick] (prev) -- (now);
      }
      \coordinate (prev) at (\i,\a) {};
    }
    \end{tikzpicture}
    \\
    \nab&\begin{tikzpicture}[baseline=-\the\dimexpr\fontdimen22\textfont2\relax]
    \draw[-, gray] (3.5,-0.2) -- (3.5,-0.1);
    \draw[-, gray] (4.0,-0.2) -- (4.0,-0.1);
    \draw[-, gray] (5.5,-0.2) -- (5.5,-0.1);
    \draw[-, gray] (6.0,-0.2) -- (6.0,-0.1);
    \draw[-, gray] (0,-0.2) -- (9.5,-0.2);
    \foreach \i/\a in
    {0.0/ -0.2, 0.1/ -0.2, 0.2/ -0.2, 0.3/ -0.2, 0.4/ -0.2, 0.5/ -0.2, 0.6/ -0.2, 0.7/ -0.2, 0.8/ -0.2, 0.9/ -0.2, 1.0/ -0.2, 1.1/ -0.2, 1.2/ -0.2, 1.3/ -0.2, 1.4/ -0.2, 1.5/ -0.2, 1.6/ -0.2, 1.7/ -0.2, 1.8/ -0.2, 1.9/ -0.2, 2.0/ -0.2, 2.1/ -0.2, 2.2/ -0.2, 2.3/ -0.2, 2.4/ -0.2, 2.5/ -0.2, 2.6/ -0.2, 2.7/ -0.2, 2.8/ -0.2, 2.9/ -0.2, 3.0/ -0.2, 3.1/ -0.2, 3.2/ -0.2, 3.3/ -0.2, 3.4/ -0.2, 3.5/ -0.2, 3.6/ 0.214, 3.7/ 0.235, 3.8/ 0.257, 3.9/ 0.278, 4.0/ 0.3, 4.1/ 0.299, 4.2/ 0.298, 4.3/ 0.297, 4.4/ 0.296, 4.5/ 0.294, 4.6/ 0.291, 4.7/ 0.288, 4.8/ 0.284, 4.9/ 0.279, 5.0/ 0.273, 5.1/ 0.265, 5.2/ 0.255, 5.3/ 0.244, 5.4/ 0.23, 5.5/ 0.214, 5.6/ 0.193, 5.7/ 0.167, 5.8/ 0.136, 5.9/ 0.102, 6.0/ -0.131, 6.1/ -0.143, 6.2/ -0.153, 6.3/ -0.162, 6.4/ -0.169, 6.5/ -0.175, 6.6/ -0.181, 6.7/ -0.185, 6.8/ -0.188, 6.9/ -0.191, 7.0/ -0.193, 7.1/ -0.194, 7.2/ -0.196, 7.3/ -0.197, 7.4/ -0.197, 7.5/ -0.198, 7.6/ -0.198, 7.7/ -0.199, 7.8/ -0.199, 7.9/ -0.199, 8.0/ -0.199, 8.1/ -0.2, 8.2/ -0.2, 8.3/ -0.2, 8.4/ -0.2, 8.5/ -0.2, 8.6/ -0.2, 8.7/ -0.2, 8.8/ -0.2, 8.9/ -0.2, 9.0/ -0.2, 9.1/ -0.2, 9.2/ -0.2, 9.3/ -0.2, 9.4/ -0.2, 9.5/ -0.2}{
    \coordinate (now) at (\i,\a) {};
      \ifthenelse{\equal{\i}{0.0}}{}{
      \draw[-, teal, thick] (prev) -- (now);
      }
      \coordinate (prev) at (\i,\a) {};
    }
    \end{tikzpicture}
    \\
    \tempdist&\begin{tikzpicture}[baseline=-\the\dimexpr\fontdimen22\textfont2\relax]
    \draw[-, gray] (3.5,-0.2) -- (3.5,-0.1);
    \draw[-, gray] (4.0,-0.2) -- (4.0,-0.1);
    \draw[-, gray] (5.5,-0.2) -- (5.5,-0.1);
    \draw[-, gray] (6.0,-0.2) -- (6.0,-0.1);
    \draw[-, gray] (0,-0.2) -- (9.5,-0.2);
    \foreach \i/\a in
    {0.0/ 0.3, 0.1/ 0.288, 0.2/ 0.276, 0.3/ 0.264, 0.4/ 0.252, 0.5/ 0.24, 0.6/ 0.228, 0.7/ 0.216, 0.8/ 0.203, 0.9/ 0.191, 1.0/ 0.179, 1.1/ 0.167, 1.2/ 0.155, 1.3/ 0.143, 1.4/ 0.131, 1.5/ 0.119, 1.6/ 0.107, 1.7/ 0.095, 1.8/ 0.083, 1.9/ 0.071, 2.0/ 0.059, 2.1/ 0.047, 2.2/ 0.035, 2.3/ 0.022, 2.4/ 0.01, 2.5/ -0.002, 2.6/ -0.014, 2.7/ -0.026, 2.8/ -0.038, 2.9/ -0.05, 3.0/ -0.062, 3.1/ -0.074, 3.2/ -0.086, 3.3/ -0.098, 3.4/ -0.11, 3.5/ -0.122, 3.6/ -0.134, 3.7/ -0.145, 3.8/ -0.156, 3.9/ -0.165, 4.0/ -0.173, 4.1/ -0.18, 4.2/ -0.186, 4.3/ -0.19, 4.4/ -0.194, 4.5/ -0.197, 4.6/ -0.199, 4.7/ -0.2, 4.8/ -0.2, 4.9/ -0.199, 5.0/ -0.197, 5.1/ -0.194, 5.2/ -0.19, 5.3/ -0.186, 5.4/ -0.18, 5.5/ -0.173, 5.6/ -0.165, 5.7/ -0.156, 5.8/ -0.145, 5.9/ -0.134, 6.0/ -0.122, 6.1/ -0.11, 6.2/ -0.098, 6.3/ -0.086, 6.4/ -0.074, 6.5/ -0.062, 6.6/ -0.05, 6.7/ -0.038, 6.8/ -0.026, 6.9/ -0.014, 7.0/ -0.002, 7.1/ 0.01, 7.2/ 0.022, 7.3/ 0.035, 7.4/ 0.047, 7.5/ 0.059, 7.6/ 0.071, 7.7/ 0.083, 7.8/ 0.095, 7.9/ 0.107, 8.0/ 0.119, 8.1/ 0.131, 8.2/ 0.143, 8.3/ 0.155, 8.4/ 0.167, 8.5/ 0.179, 8.6/ 0.191, 8.7/ 0.203, 8.8/ 0.216, 8.9/ 0.228, 9.0/ 0.24, 9.1/ 0.252, 9.2/ 0.264, 9.3/ 0.276, 9.4/ 0.288, 9.5/ 0.3}{
    \coordinate (now) at (\i,\a) {};
      \ifthenelse{\equal{\i}{0.0}}{}{
      \draw[-, teal, thick] (prev) -- (now);
      }
      \coordinate (prev) at (\i,\a) {};
    }
    \end{tikzpicture}
    \\
    \rf[1]{flat}{0.2}&\begin{tikzpicture}[baseline=-\the\dimexpr\fontdimen22\textfont2\relax]
    \draw[-, gray] (3.5,-0.2) -- (3.5,-0.1);
    \draw[-, gray] (4.0,-0.2) -- (4.0,-0.1);
    \draw[-, gray] (5.5,-0.2) -- (5.5,-0.1);
    \draw[-, gray] (6.0,-0.2) -- (6.0,-0.1);
    \draw[-, gray] (0,-0.2) -- (9.5,-0.2);
    \foreach \i/\a in
    {0.0/ -0.2, 0.1/ -0.2, 0.2/ -0.2, 0.3/ -0.2, 0.4/ -0.2, 0.5/ -0.2, 0.6/ -0.2, 0.7/ -0.2, 0.8/ -0.2, 0.9/ -0.2, 1.0/ -0.2, 1.1/ -0.2, 1.2/ -0.2, 1.3/ -0.2, 1.4/ -0.2, 1.5/ -0.2, 1.6/ -0.2, 1.7/ -0.2, 1.8/ -0.2, 1.9/ -0.2, 2.0/ -0.2, 2.1/ -0.2, 2.2/ -0.2, 2.3/ -0.2, 2.4/ -0.2, 2.5/ -0.2, 2.6/ -0.2, 2.7/ -0.2, 2.8/ -0.2, 2.9/ -0.2, 3.0/ -0.2, 3.1/ -0.2, 3.2/ -0.2, 3.3/ -0.2, 3.4/ -0.2, 3.5/ -0.2, 3.6/ 0.052, 3.7/ 0.118, 3.8/ 0.181, 3.9/ 0.241, 4.0/ 0.3, 4.1/ 0.3, 4.2/ 0.3, 4.3/ 0.3, 4.4/ 0.3, 4.5/ 0.3, 4.6/ 0.3, 4.7/ 0.3, 4.8/ 0.3, 4.9/ 0.3, 5.0/ 0.3, 5.1/ 0.3, 5.2/ 0.3, 5.3/ 0.3, 5.4/ 0.3, 5.5/ 0.3, 5.6/ 0.241, 5.7/ 0.181, 5.8/ 0.118, 5.9/ 0.052, 6.0/ -0.2, 6.1/ -0.2, 6.2/ -0.2, 6.3/ -0.2, 6.4/ -0.2, 6.5/ -0.2, 6.6/ -0.2, 6.7/ -0.2, 6.8/ -0.2, 6.9/ -0.2, 7.0/ -0.2, 7.1/ -0.2, 7.2/ -0.2, 7.3/ -0.2, 7.4/ -0.2, 7.5/ -0.2, 7.6/ -0.2, 7.7/ -0.2, 7.8/ -0.2, 7.9/ -0.2, 8.0/ -0.2, 8.1/ -0.2, 8.2/ -0.2, 8.3/ -0.2, 8.4/ -0.2, 8.5/ -0.2, 8.6/ -0.2, 8.7/ -0.2, 8.8/ -0.2, 8.9/ -0.2, 9.0/ -0.2, 9.1/ -0.2, 9.2/ -0.2, 9.3/ -0.2, 9.4/ -0.2, 9.5/ -0.2}{
    \coordinate (now) at (\i,\a) {};
      \ifthenelse{\equal{\i}{0.0}}{}{
      \draw[-, teal, thick] (prev) -- (now);
      }
      \coordinate (prev) at (\i,\a) {};
    }
    \end{tikzpicture}
    \\
    \rf[1]{front}{0}&\begin{tikzpicture}[baseline=-\the\dimexpr\fontdimen22\textfont2\relax]
    \draw[-, gray] (3.5,-0.2) -- (3.5,-0.1);
    \draw[-, gray] (4.0,-0.2) -- (4.0,-0.1);
    \draw[-, gray] (5.5,-0.2) -- (5.5,-0.1);
    \draw[-, gray] (6.0,-0.2) -- (6.0,-0.1);
    \draw[-, gray] (0,-0.2) -- (9.5,-0.2);
    \foreach \i/\a in
    {0.0/ -0.2, 0.1/ -0.2, 0.2/ -0.2, 0.3/ -0.2, 0.4/ -0.2, 0.5/ -0.2, 0.6/ -0.2, 0.7/ -0.2, 0.8/ -0.2, 0.9/ -0.2, 1.0/ -0.2, 1.1/ -0.2, 1.2/ -0.2, 1.3/ -0.2, 1.4/ -0.2, 1.5/ -0.2, 1.6/ -0.2, 1.7/ -0.2, 1.8/ -0.2, 1.9/ -0.2, 2.0/ -0.2, 2.1/ -0.2, 2.2/ -0.2, 2.3/ -0.2, 2.4/ -0.2, 2.5/ -0.2, 2.6/ -0.2, 2.7/ -0.2, 2.8/ -0.2, 2.9/ -0.2, 3.0/ -0.2, 3.1/ -0.2, 3.2/ -0.2, 3.3/ -0.2, 3.4/ -0.2, 3.5/ -0.2, 3.6/ -0.135, 3.7/ -0.04, 3.8/ 0.07, 3.9/ 0.184, 4.0/ 0.3, 4.1/ 0.28, 4.2/ 0.26, 4.3/ 0.239, 4.4/ 0.217, 4.5/ 0.194, 4.6/ 0.17, 4.7/ 0.146, 4.8/ 0.121, 4.9/ 0.094, 5.0/ 0.067, 5.1/ 0.038, 5.2/ 0.008, 5.3/ -0.023, 5.4/ -0.055, 5.5/ -0.089, 5.6/ -0.124, 5.7/ -0.154, 5.8/ -0.177, 5.9/ -0.192, 6.0/ -0.2, 6.1/ -0.2, 6.2/ -0.2, 6.3/ -0.2, 6.4/ -0.2, 6.5/ -0.2, 6.6/ -0.2, 6.7/ -0.2, 6.8/ -0.2, 6.9/ -0.2, 7.0/ -0.2, 7.1/ -0.2, 7.2/ -0.2, 7.3/ -0.2, 7.4/ -0.2, 7.5/ -0.2, 7.6/ -0.2, 7.7/ -0.2, 7.8/ -0.2, 7.9/ -0.2, 8.0/ -0.2, 8.1/ -0.2, 8.2/ -0.2, 8.3/ -0.2, 8.4/ -0.2, 8.5/ -0.2, 8.6/ -0.2, 8.7/ -0.2, 8.8/ -0.2, 8.9/ -0.2, 9.0/ -0.2, 9.1/ -0.2, 9.2/ -0.2, 9.3/ -0.2, 9.4/ -0.2, 9.5/ -0.2}{
    \coordinate (now) at (\i,\a) {};
      \ifthenelse{\equal{\i}{0.0}}{}{
      \draw[-, teal, thick] (prev) -- (now);
      }
      \coordinate (prev) at (\i,\a) {};
    }
    \end{tikzpicture}
    \\
    \bottomrule
    \end{tabular}
    \caption{\textbf{Score as a function of position of the predicted event:} For each metric, the graph shows the scores of the detection scenario shown in Figure~\ref{fig:disc_graph_case}, as a function of the position in the prediction. All graphs are scaled to the same interval for easy comparizon.}
    \label{tab:discontinuity}
\end{figure*}

\begin{figure}
    \centering
\begin{tikzpicture}
\begin{axis}[
    height=0.4\textwidth,
    width=0.7\textwidth,
    xlabel=$t$,
    xmin=0, xmax=5,
    ymin=0, ymax=12,
    samples=30,
    smooth
]
\addplot[blue, mark=*, domain=0:5, draw=none] {sin(800*x)/2+2+7*(x<4)-7*(x<3)};
\end{axis}
\end{tikzpicture}
    \caption{Is there one anomaly at $t\approx 3$ and another at $t\approx4$, or just one anomalous event from 3 to 4? Without any information about domain and time scale, we may only guess what is an anomaly here, or if there even exist any.}
    \label{fig:changed_mean}
\end{figure}
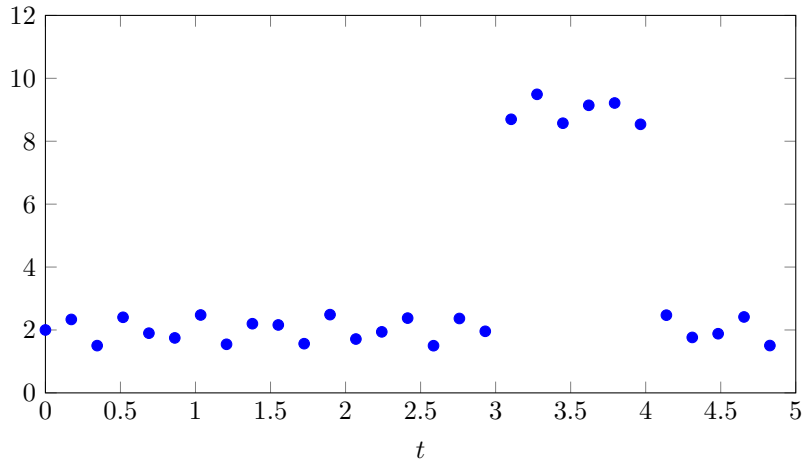

An effect of valuing proximity is that the score is less dependent on the labelling strategy. We show this with an example.
The labels of a dataset are usually not perfect, and often it is not clear what is an anomaly, and where an anomaly starts or ends. While the score of an anomaly detector always will depend heavily on what is considered a GT anomaly and not, the sensitivity to the exact length and location to an anomaly varies. Figure~\ref{fig:changed_mean} shows a situation where it is not clear where to put the anomaly labels. 
One possibility is to mark all the high valued points as anomalous. Another strategy is to label only the points around the discontinuities, e.g. as done by \cite{Lai2021RevisitingTS}. Indeed, there may be nothing anomalous about the points in between these jumps. Yet, if the distance between the jumps is small enough, it makes more sense to view it as a single contiguous anomaly - as noted by \cite{Wu2022CurrentTS}, a single normal point between two anomalies is an anomaly in its own right. Thus at some time scale in between these situations, it should be unclear how to label this event.
Two possible labels corresponding to this time series are shown in Figure~\ref{tab:labelling_problem}, along with scores for predicting the labels from the opposite strategy. The metrics valuing proximity are more tolerant to the labelling strategy, and give good scores in both cases, as opposed to the other metrics.

\subsection{Non-binary cases}

As non-binary metric use the raw anomaly score as input, the space of possible inputs is much larger, making it more difficult to do extensive examinations of how these metrics reacts to a representative variation of realistic inputs. Nevertheless, we attempt to visialize some properties of these metrics as well.
Before presenting these tests, we emphasize that the results of these metrics are dependent only on the relative anomaly score at each point, and not their actual value. This is shown in Figure~\ref{tab:ordering}, where the anomaly scores are both symmetric, and decreasing in the distance from the middle. This gives the same scores for all the metrics, independent of the labels.
For most experiments in this section, we have only a very few possible values of the anomaly scores, and the points that are not visually different, have the same score. The exceptions of this are specified in the captions.

\begin{figure*}
    \centering
\showScoreValueProblem

    \caption{Ordering the time stamps by value of the anomaly scores yield the same order. Only the order matter for the non-binary metrics, not their value, hence these predictions have the same scores. This would be true for any labels.
    }
    \label{tab:ordering}

\end{figure*}

\subsubsection{Effect of anomaly length}

\begin{figure*}[]
    \centering
\showNonbinaryLengthI
    \caption{\textbf{Effect of anomaly length:} Importance of anomaly length for non-binary metrics.}
    \label{tab:nonbinary_length_1}
\end{figure*}

Figure~\ref{tab:nonbinary_length_1} shows that the non-binary metrics mostly favour detecting the long anomalies, as these have more points. However, the VUS metrics can favour detecting the short ones if there are more of them, as the anomaly events are effectively widened by the metric.

\subsubsection{Preference for short predicted anomalies}

\begin{figure*}
    \centering
\showNonbinaryShortPredictions
    \caption{\textbf{Preference for short predicted anomalies:} None of the non-binary metrics prefer the short predicted anomalies.}
    \label{tab:nonbinary_short_preference}
\end{figure*}

Figure~\ref{tab:nonbinary_short_preference} shows predictions with short and wide anomalies, similar to the binary case shown in Figure~\ref{tab:short_predictions}. We see that none of these metrics have the short predicted anomaly preference like \paf~and \nab.

\subsubsection{Partial detection vs covering}
Similar to for the binary metrics, we test the value of detection compared to covering in Figure~\ref{tab:nonbinary_doc}.
Since all the non-binary metrics considered are point-based, none of them value the detection of the second anomaly over covering the first one. \patk~however, value them equally in this case, since $K$ is larger than the number of points with positive anomaly score.

\begin{figure*}[]
    \centering
\tabcolsep=0.08cm
\showNonbinaryDetectionOverCovering
\caption{\textbf{Partial detection vs covering:} As these metrics are point-based, they do not value detection of new anomalies over fully covering existing ones.}
    \label{tab:nonbinary_doc}
\end{figure*}

\subsubsection{Proximity}

\begin{figure*}
    \centering
\showNonbinaryNonsmoothCloseFp
\caption{\textbf{Proximity:} Only VUS metrics value proximity of predicted and labelled anomalies. The other metrics still give positive score, since a low enough threshold marks every point as anomalous.}
\label{tab:nonbinary_nonsmooth_close_fp}
\end{figure*}

\begin{figure*}
    \centering
\showNonbinaryCloseFp
\caption{\textbf{Proximity:} Non-binary metrics value FPs close to true anomalies indirectly, due to the anomaly score often being somewhat smooth. The scores are gaussian function with different shifts, meaning the anomaly scores at the anomalous points are increase as the centre moves towards the anomaly.}
\label{tab:nonbinary_close_fp}
\end{figure*}

By smoothing out the labels, the VUS metrics value proximity of predicted and labelled anomalies, as seen in Figure~\ref{tab:nonbinary_nonsmooth_close_fp}. The other non-binary metrics do not value high anomaly scores close to an anomaly. However, since anomaly scores often are somewhat smooth, high anomaly scores close to the anomaly can indicate that the anomaly score is also relatively high at the anomaly. This is shown in Figure~\ref{tab:nonbinary_close_fp}, where the anomaly scores are bell curves at different locations. This way of valuing proximity does however fully rely on the form of the anomaly score, which may not necessarily be fair.

\subsubsection{Effect of class imbalance}

\begin{figure*}
\centering
\showAucRocProblemIII
\caption{\textbf{Effect of class imbalance:} \aucroc~and \vusroc{l}~scores are heavily affected by the amount of TNs. In the shorter predictions, only the predicted part of the time series is evaluated. The anomaly score is strictly decreasing, ensuring that none of the added points are more anomalous than the previous ones. As only \aucroc~and \vusroc{l}~are affected by this, the other metrics are not included.}
\label{tab:auc_roc_problem_3}
\end{figure*}

Unlike all the other metrics we have considered, \aucroc~and \vusroc{l}~include the number TNs in their formulas. This means including extra points that would not affect other metrics, will affect these. This is shown in Figure~\ref{tab:auc_roc_problem_3}. We see that the scores of \aucroc~and \vusroc{l}~increase from about 0.05 to about 0.95 as the anomaly ratio is decreased from 4/8 to 4/64. This means that for low anomaly ratios, precise detections are less important. This can yield high scores that seem impressive, but are difficult to interpret correctly.
An example of this changing the required precision is shown in Figure~\ref{tab:auc_roc_problem_2}. While \aucroc~and \vusroc{l}~prefer the short predicted event in the short time series, they prefer the less precise one in the long time series.

\begin{figure*}
    \centering
\showAucRocProblemII
\caption{\textbf{Effect of class imbalance:} The bottom two anomaly scores are extensions of the top two. The extra TNs change which anomaly score \aucroc~and \vusroc{l}~prefer.}
\label{tab:auc_roc_problem_2}
\end{figure*}

\section{Categorization}
\label{sec:categorization}

In this section we present how each metric relates to the properties presented in Section~\ref{sec:properties}.
Figure~\ref{tab:categories} show the properties of each metric. We will in the following paragraphs explain how these results were determined.
If the result of any test depends on a parameter\footnote{For \rf{bias}{\alpha}, where the function parameters could in principle be anything, we have only used the ones suggested in the original paper}, we mark it by an asteriks. We also use an asteriks for partially obtained properties, by what we mean is explained for the relevant properties below.

For \textit{Early detection}, we consider the results shown in Figure~\ref{tab:discontinuity} for the binary metrics. If the score at the second mark is higher than the third mark, we consider the metric to value earliness. Since none of the non-binary metrics use the direction of the time series in their calculation, they can not have this property due to symmetry in the time dimension.

The \textit{Long anomalies} preference property is based on the result in Figure~\ref{tab:length_problem_1} and Figure~\ref{tab:nonbinary_length_1} for binary and non-binary metrics respectively - metrics not preferring the bottom model are considered to have this property.
Similarly, the \textit{Short predicted anomaly} preference property is based on Figures \ref{tab:short_predictions} and \ref{tab:nonbinary_short_preference}. Metrics giving \textit{better} score to the top model have this property.
Metrics with the \textit{Partial detection} preference are those not preferring the top prediction in Figures~\ref{tab:detection_over_covering} and \ref{tab:nonbinary_doc}.

The \textit{Proximity} property for binary metrics is based on Figure~\ref{tab:discontinuity}. Metrics that have non-zero score on the first \textit{and} last mark are considered to fully have this property, while metrics with non-zero score only the last mark partially have the property and are marked with an asterisk. For non-binary metrics, metrics distinguishing the anomaly scores Figure~\ref{tab:nonbinary_nonsmooth_close_fp} are considered to have the property, while metrics only distinguishing the smoother anomaly scores of Figure~\ref{tab:nonbinary_close_fp} are marked with an asterisk.
We do not consider this property do depend on parameters for \ttolf{\tau}, \vusroc{l}~and \vuspr{l}, 
since $\tau=0$ or $l=0$, the only values where the property is not obtained, 
would make the metrics identical to \pwf, \aucroc~and \aucpr.

The \textit{Requires threshold} parameter simply indicates the binary vs non-binary metrics. \textit{\# parameters} indicate the number of parameters for each metric, including $\beta$ for f-scores, all specifiable functions for \rf{bias}{\alpha}, the distance exponent in \tempdist, and all the TP, FP and FN weigths in \nab.

\textit{Time aware} indicates the metrics that consider time dimension adjacency in any way, \textit{Indifferent to imbalance} are the metrics ignoring the amount of true negatives, and \textit{General} are the metrics suitable for all TSAD application. It should be clear by now that the latter property is quite unachievable without a impractically large amount of parameters. Nevertheless, the category is included for anyone skimming the article in search of simple answers.

\begin{figure*}[!ht] 
\newcommand{\cmark}{\textcolor{brown}{\ding{52}}}
\newcommand{\xmark}{\textcolor{blue}{\ding{56}}}
\newcolumntype{R}[2]{
    >{\adjustbox{angle=#1,lap=\width-(#2)}\bgroup}
    l
    <{\egroup}
}
\newcommand*\rot{\multicolumn{1}{R{70}{1em}}}
    \centering
    \tabcolsep=0.2cm
    \begin{tabular}{l lllll c ll c lll}
    \toprule
        Metric & \multicolumn{5}{c}{Preferences} &\ \ & \multicolumn{2}{c}{Requirements} &\ \ & \multicolumn{3}{c}{Suitability} \\
        \cline{2-6} \cline{8-9} \cline{11-13}
        ~ & \rot{Early detection} & \rot{Long anomalies} & \rot{Short predicted anomalies} & \rot{Partial detection} & \rot{Proximity} && \rot{Requires threshold} & \rot{\# parameters} &&\rot{Time aware}&\rot{Indifferent to imbalance}&\rot{General}\\ \hline
        \pwf & \xmark & \cmark & \xmark & \xmark & \xmark && \cmark &  1 && \xmark & \cmark & \xmark\\ 
        \paf & \xmark & \cmark & \cmark & \cmark & \xmark && \cmark & 1 && \cmark & \cmark & \xmark\\ 
        \dtpaf & * & \cmark & * & \cmark & \xmark && \cmark & 2 && \cmark & \cmark & \xmark\\ 
        \pakf{K} & \xmark & \cmark & * & * & \xmark && \cmark & 2 && \cmark & \cmark & \xmark\\ 
        \lsf{n} & * & * & * & \cmark & * && \cmark & 2 && \cmark & \cmark & \xmark\\ 
        \segf & \xmark & \xmark & \xmark & \cmark & \xmark && \cmark & 1 && \cmark & \cmark & \xmark\\ 
        \cf & \xmark & \xmark & \xmark & \cmark & \xmark && \cmark & 1 && \cmark & \cmark & \xmark\\ 
        \ttolf{\tau} & \xmark & * & \xmark & * & \cmark && \cmark & 2 && \cmark & \cmark & \xmark\\ 
        \rf{bias}{\alpha} & * & \xmark & \xmark & * & \xmark && \cmark & 8 && \cmark & \cmark & \xmark\\  
        \taf{\alpha}{\delta}{\theta} & \xmark & \xmark & \xmark & * & * && \cmark & 4 && \cmark & \cmark & \xmark\\ 
        \etaf{\theta_p}{\theta_r} & \xmark & \xmark & \xmark & * & \xmark && \cmark & 3 && \cmark & \cmark & \xmark\\ 
        \af & \xmark & \xmark & \xmark & \cmark & \cmark && \cmark & 1 && \cmark & \cmark & \xmark\\ 
        \nab & \cmark & \xmark & \cmark & \cmark & * && \cmark & 3 && \cmark & \cmark & \xmark\\ 
        \tempdist & \xmark & * & \cmark & * & \cmark && \cmark & 1 && \cmark & \cmark & \xmark\\ 
        \hline
        \patk[K] & \xmark & \cmark & \xmark & \xmark & * && \xmark & 1 && \xmark & \cmark & \xmark\\
        \bestpwf & \xmark & \cmark & \xmark & \xmark & * && \xmark &  1 && \xmark & \cmark & \xmark\\ 
        \aucroc & \xmark & \cmark & \xmark & \xmark & * && \xmark & 0 && \xmark & \xmark & \xmark\\ 
        \aucpr~& \xmark & \cmark & \xmark & \xmark & * && \xmark & 0 && \xmark & \cmark & \xmark\\ 
        \vusroc{l} & \xmark & * & \xmark & \xmark & \cmark && \xmark & 1 && \cmark & \xmark & \xmark\\
        \vuspr{l} & \xmark & \cmark & \xmark & \xmark & \cmark && \xmark & 1 && \cmark & \cmark & \xmark\\ 
    \bottomrule

    \end{tabular}
    \caption{
    The properties of all the metrics.\\\cmark = has property, \\\xmark = does not have property, \\ * = partially / parameter dependent.
    }
    \label{tab:categories}
\end{figure*}

\section{Conclusion}
\label{sec:conclusion}
Through an extensive literature review on time series anomaly detection (TSAD), we found several different ways to evaluate algorithms. While a rigorous discussion on several of the available metrics can be found in a few papers, some of which strongly disagree with each other on what are important properties of an evaluation metrics, most papers choose metrics that have been repeatedly faulted in the literature.
We have tested 20 TSAD evaluation metrics in several case studies, and categorized them based on 10 different properties. As TSAD is a diverse field, no evaluation metrics is appropriate in all cases, and it should be chosen with care in each case.
For the same reason, it is difficult to provide detailed guidelines for how to do this. However, we summarize some of the main takeaways from our study:
\begin{itemize}
    \item The choice of evaluation metric has a large impact on the rankings of TSAD methods, underscoring the need for careful alignment of evaluation metrics with specific problem requirements. 
    \item 
    Some metrics give high scores to certain prediction strategies that are not necessarily good strategies. For example, predicting only very long or very short anomalies can result in unreasonably high scores, leading to the selection of inappropriate methods and an overestimation of expected performance.
    \item Some metrics may result in very bad scores for certain types of predictions, even though the predictions are valuable, such as predicting long anomalous events, or predicting anomalies too early or late.  This can lead to selecting ineffective methods and underestimating the expected performance.
    \item Due to the way the labels are compared to the prediction, some metrics are less appropriate for certain kinds of labelling strategies.
\end{itemize}
Therefore, it is crucial to carefully select the appropriate evaluation metric for a given problem, taking into consideration its preferences for specific types of predictions. Simple case studies such as the ones presented in this work can be helpful for gaining such understanding.

There are several directions of future research based on this study.
First of all, there is room for defining novel evaluation metrics. For example, valuing earliness and proximity are two very useful traits, but none of the metrics we could find include both.
Furthermore, much more investigation can be done of existing metrics that did not meet the limitations of this work, e.g. supplementary performance analysis metrics, or combinations of techniques of the included metrics.
Finally, when publishing results in TSAD research in general, we suggest including results from multiple metrics, as well as making both the code and the anomaly scores available, to enable easy comparison with any evaluation metric.

\section*{Acknowledgments}
This research was carried out with the support of the ML4ITS project (312062), funded by the Norwegian Research Council (NFR).

\bibliographystyle{unsrt}  
\bibliography{references}

\end{document}